\documentclass[9pt,twocolumn,twoside]{article}           
\usepackage[dvips]{graphicx}
\usepackage{amsfonts}
\usepackage{blindtext, rotating}
\usepackage{amssymb}
\usepackage{amsbsy} 
\usepackage{amsmath}
\usepackage[ansinew]{inputenc}
\usepackage{subfigure}
\usepackage{cite}
\usepackage{color}
\usepackage{hyperref}
\usepackage[margin=1.2cm]{geometry}
\usepackage{times}
\usepackage[all]{xy}
\usepackage{lscape}
\usepackage{flushend,pmat}
\usepackage{url}

\usepackage{multirow}

\usepackage{color}
\usepackage{soul}
\usepackage[normalem]{ulem} 

\usepackage{times}
\usepackage[all]{xy}
\usepackage{lscape}
\usepackage{flushend,pmat}
\usepackage{url}

\usepackage{multicol}

\newtheorem{property}{\it Property}

\newcommand{\red}[1]{\textcolor[rgb]{0.5,0,0}{#1}}           
\newcommand{\green}[1]{\textcolor[rgb]{0,0.5,0}{#1}}           

\newcommand{\x}{{\bf x}}
\newcommand{\R}{{\bf R}}
\newcommand{\rr}{{\bf r}}
\newcommand{\e}{{\bf e}}

\newcommand{\m}{{\bf m}}

\newcommand{\I}{{\bf I}}
\newcommand{\M}{{\bf M}}
\newcommand{\W}{{\bf W}}
\newcommand{\V}{{\bf V}}
\newcommand{\X}{{\bf X}}
\newcommand{\E}{{\bf E}}
\newcommand{\Q}{{\bf Q}}

\begin{document}

\title{Principal Polynomial Analysis}
\date{}
\author{Valero Laparra, Sandra Jim\'enez, \\
Devis Tuia, Gustau Camps-Valls and Jes\'us Malo
\thanks{Electronic version of an article published as Int. J. Neural Syst. 24(7) (2014) [DOI: 10.1142/S0129065714400073] [copyright World Scientific Publishing Company].}
\thanks{Image Processing Laboratory (IPL), Universitat de Val\`encia, Catedr\'atico A. Escardino - 46980 Paterna, Val\`encia (Spain). E-mail: \{valero.laparra, jesus.malo, gustau.camps\}@uv.es}
\thanks{This work was partially supported by the Spanish Ministry of Economy and Competitiveness (MINECO) under project TIN2012-38102-C03-01, and under a  EUMETSAT contract.}
}
\pagestyle{myheadings}
\markboth{2015. Published in IJNS - World Scientific. DOI: 10.1142/S0129065714400073}{Laparra et al., 2015}

\maketitle

\begin{abstract}
This paper presents a new framework for manifold learning based on a sequence of principal polynomials that capture the possibly nonlinear nature of the data. The proposed Principal Polynomial Analysis (PPA) generalizes PCA by modeling the directions of maximal variance by means of curves, instead of straight lines. Contrarily to previous approaches, PPA reduces to performing simple univariate regressions, which makes it computationally feasible and robust. Moreover, PPA shows a number of interesting analytical properties. \emph{First}, PPA is a volume-preserving map, which in turn guarantees the existence of the inverse. \emph{Second}, such an inverse can be obtained in closed form. Invertibility is an important advantage over other learning methods, because it permits to understand the identified features in the input domain where the data has physical meaning. Moreover, it allows to evaluate the performance of dimensionality reduction in sensible (input-domain) units. Volume preservation also allows an easy computation of information theoretic quantities, such as the reduction in multi-information after the transform. \emph{Third}, the analytical nature of PPA leads to a clear geometrical interpretation of the manifold: it allows the computation of Frenet-Serret frames (local features) and of generalized curvatures at any point of the space. And \emph{fourth}, the analytical Jacobian allows the computation of the metric induced by the data, thus generalizing the Mahalanobis distance.
These properties are demonstrated theoretically and illustrated experimentally. The performance of PPA is evaluated in dimensionality and redundancy reduction, in both synthetic and real datasets from the UCI repository.
\end{abstract}

\section{Introduction} \label{introduction}

Principal Component Analysis (PCA), also known as the Karhunen-Lo\`eve transform or the Hotelling transform, is a well-known method in machine learning, signal processing and statistics~\cite{Jolliffe02}.
PCA essentially builds an orthogonal transform to convert a set of observations of possibly correlated variables into a set of linearly uncorrelated variables.
PCA has been used for manifold description and dimensionality reduction in a wide range of applications because of its simplicity, energy compaction, intuitive interpretation, and invertibility. Nevertheless, PCA is hampered by data exhibiting nonlinear relations. {In this paper, we present a nonlinear generalization of PCA that, unlike other alternatives, keeps all the above mentioned appealing properties of PCA.}

\subsection{\em Desirable properties in manifold learning}

In recent years, several dimensionality reduction methods have been proposed to deal with manifolds that can not be linearly described
{(}see~\cite{Lee07} for a comprehensive review): the approaches proposed range from local methods~\cite{Tenenbaum2000,Roweis02,Verbeek02,Teh03,Brand03}, to kernel-based and spectral decompositions~\cite{Roweis00,Scholkopf98,Weinberger04},
neural networks~\cite{Kramer91,Hinton06,Scholz07}, and projection pursuit {methods}~\cite{Huber85,Laparra11}. However, despite the advantages of nonlinear methods, classical PCA still remains the most widely used dimensionality reduction technique in real applications. This is because PCA: 1) is easy to apply, 2) involves solving a convex problem, for which efficient solvers exist, 3) identifies features which are easily interpretable in terms of original variables, and 4) has a straightforward inverse and out-of-sample extension.

The above properties, which are the base of the success of PCA, are not always present in the new nonlinear dimensionality reduction
methods due either to complex formulations, to the introduction of a number of non-intuitive free parameters to be tuned, {to} their high computational cost, {to} their non-invertibility or, in some cases, to strong assumptions about the manifold.
More plausibly, the limited adoption of nonlinear methods in daily practice has to do with the lack of feature and model interpretability.
In this regard, the usefulness of data description methods is tied to the following properties:
\vspace{-0.1cm}
\begin{enumerate}
\item {\em Invertibility of the transform.} It allows both characterizing the transformed domain and evaluating the quality of the transform. On the one hand, inverting the data back to the input domain is important to understand the features in physically meaningful units, while analyzing the results in the transformed domain is typically more complicated (if not impossible). On the other hand, invertible transforms like PCA allow the assessment of the dimensionality reduction errors as simple reconstruction distortion.
    \vspace{-0.1cm}
\item {\em Geometrical interpretation of the manifold.} Understanding the system that generated the data is the ultimate goal of manifold learning. Inverting the transform is just one step towards knowledge extraction. Geometrical interpretation and analytical characterization of the manifolds give us further insight into the problem. Ideally, one would like to compute geometric properties from the learned model, such as the curvature and torsion of the manifold, or the metric induced by the data. This geometrical characterization allows to understand {the} latent parameters governing the system.
\end{enumerate}
\vspace{-0.1cm}
It is worth noting that both properties are scarcely achieved in the manifold learning literature. For instance, {\em spectral} methods do not generally yield intuitive mappings between the original and the intrinsic curvilinear coordinates of the low dimensional manifold. Even though a metric can be derived from particular kernel functions~\cite{Burges99}, the interpretation of the transformation is hidden behind an implicit mapping function, and solving the pre-image problem is generally not straightforward ~\cite{Honeine11}. In such cases, the application of (indirect) evaluation techniques has become a relevant issue for methods leading to non-invertible transforms~\cite{Venna10}. One could argue that direct and inverse transforms can be alternatively derived from mixtures of local models~\cite{Brand03}. However, the effect of these local alignment operations in the metric is not trivial. In the same way, explicit geometric descriptions of the manifold, such as the computation of curvatures, is not obvious from other invertible transforms, as autoencoders or deep networks~\cite{Kramer91,Hinton06,Scholz07,Laparra11}.

{In this paper, we introduce the Principal Polynomial Analysis (PPA), which is a {\em nonlinear generalization of PCA} that still shares all its important properties. PPA is computationally easy as it only relies on matrix inversion and multiplication, and it is robust since it reduces to a series of marginal (univariate) regressions.} PPA implements a volume-preserving and invertible map. Not only the features are easy to interpret in the input space but, additionally, the analytical nature of PPA allows to compute classical geometrical descriptors such as curvature, torsion and the induced metric at any point of the manifold. Applying the learned transform to new samples is also {as} straightforward as in PCA. {Preliminary versions of PPA were presented in~\cite{Laparra12mlsp}, and applied to remote sensing in~\cite{Laparra11igarss}. However, those conference papers did not study the analytical properties of PPA (volume preservation, invertibility, and model geometry),
nor compared with approaches that follow similar logic like NL-PCA.}

\subsection{\em Illustration of Principal Polynomial Analysis}
\label{motivation}

\begin{figure*}[t!]
\centerline{
\includegraphics[width=4.8cm]{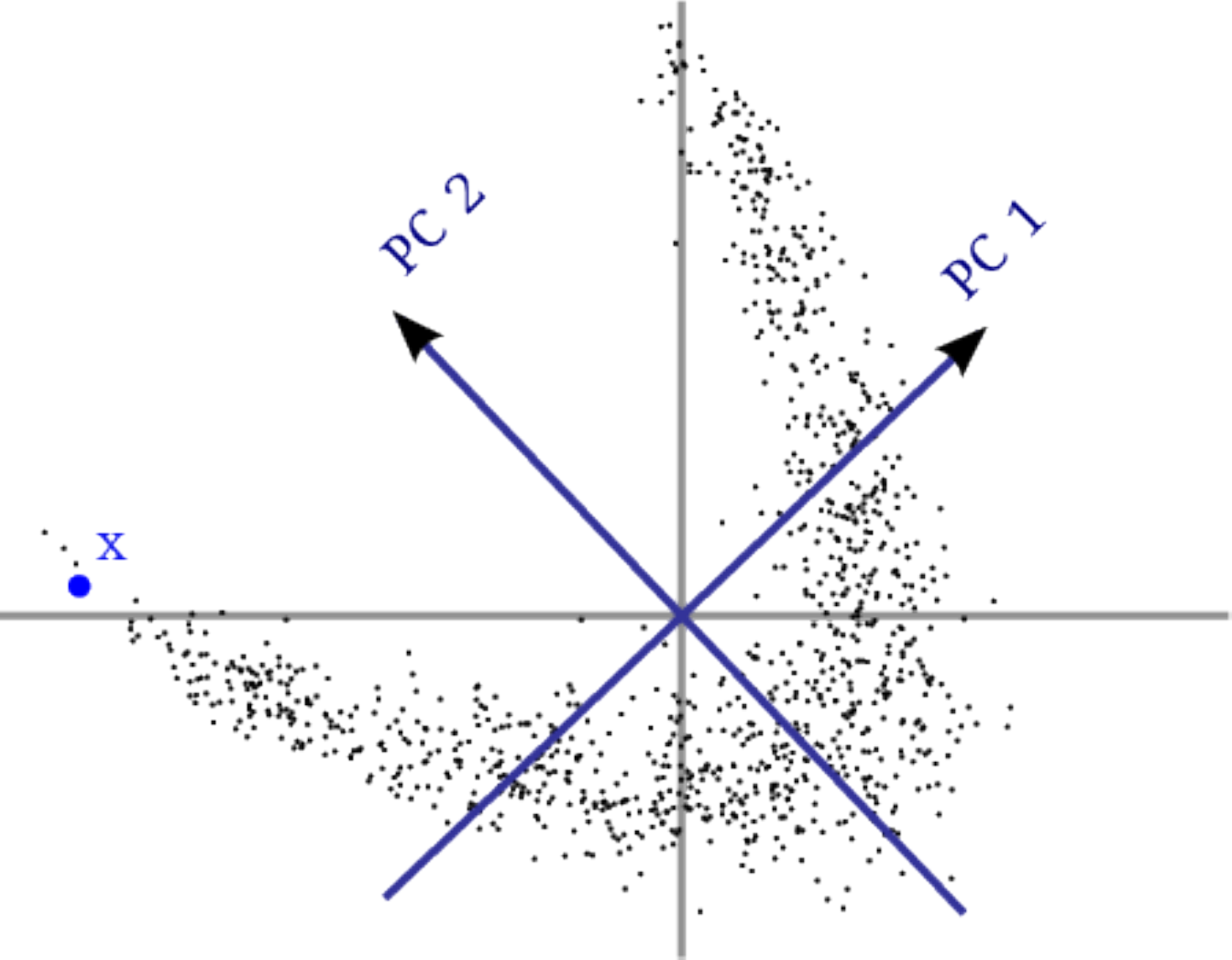} \hspace{+0.0cm}
\includegraphics[width=4.8cm]{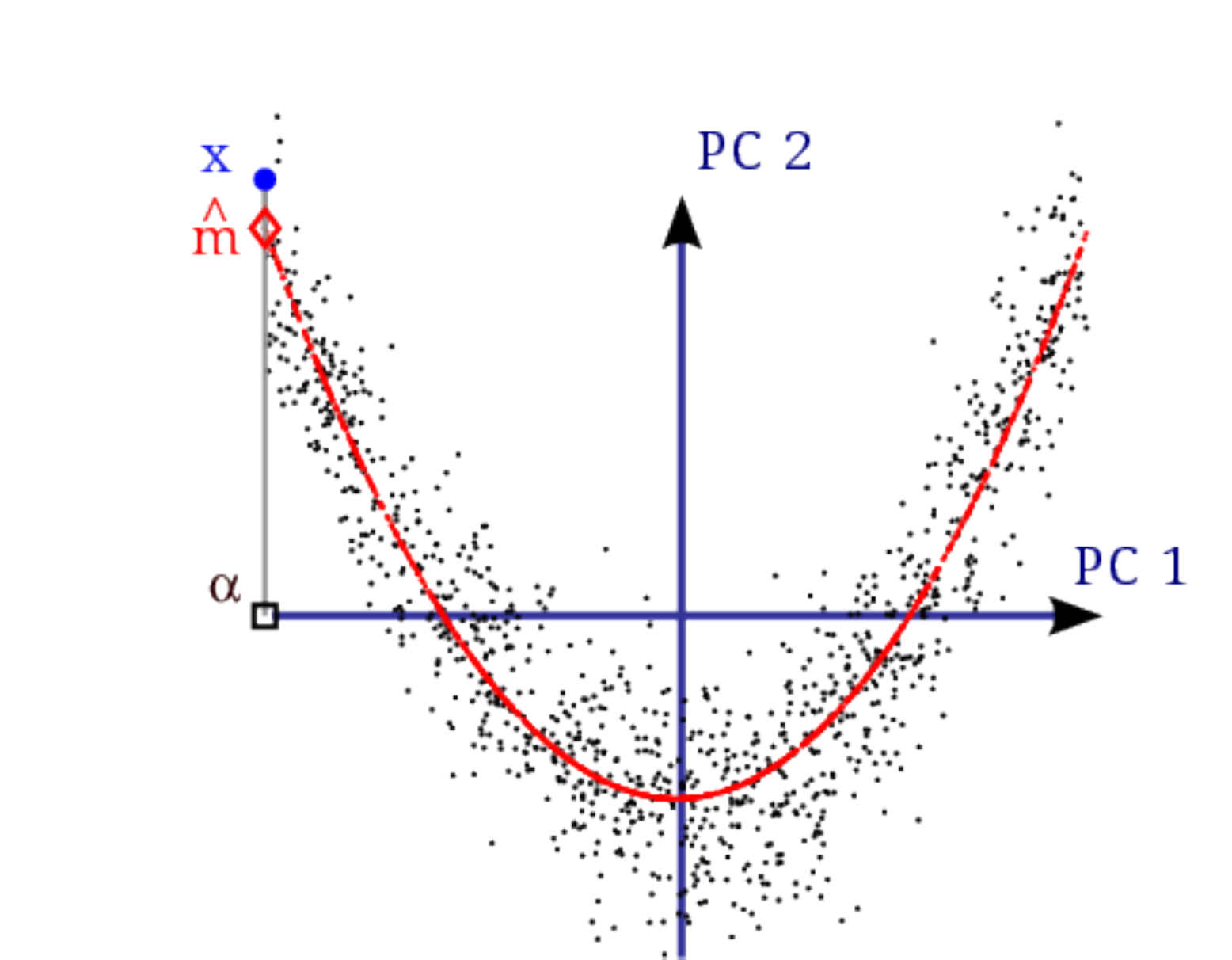} \hspace{+0.0cm}
\includegraphics[width=4.8cm]{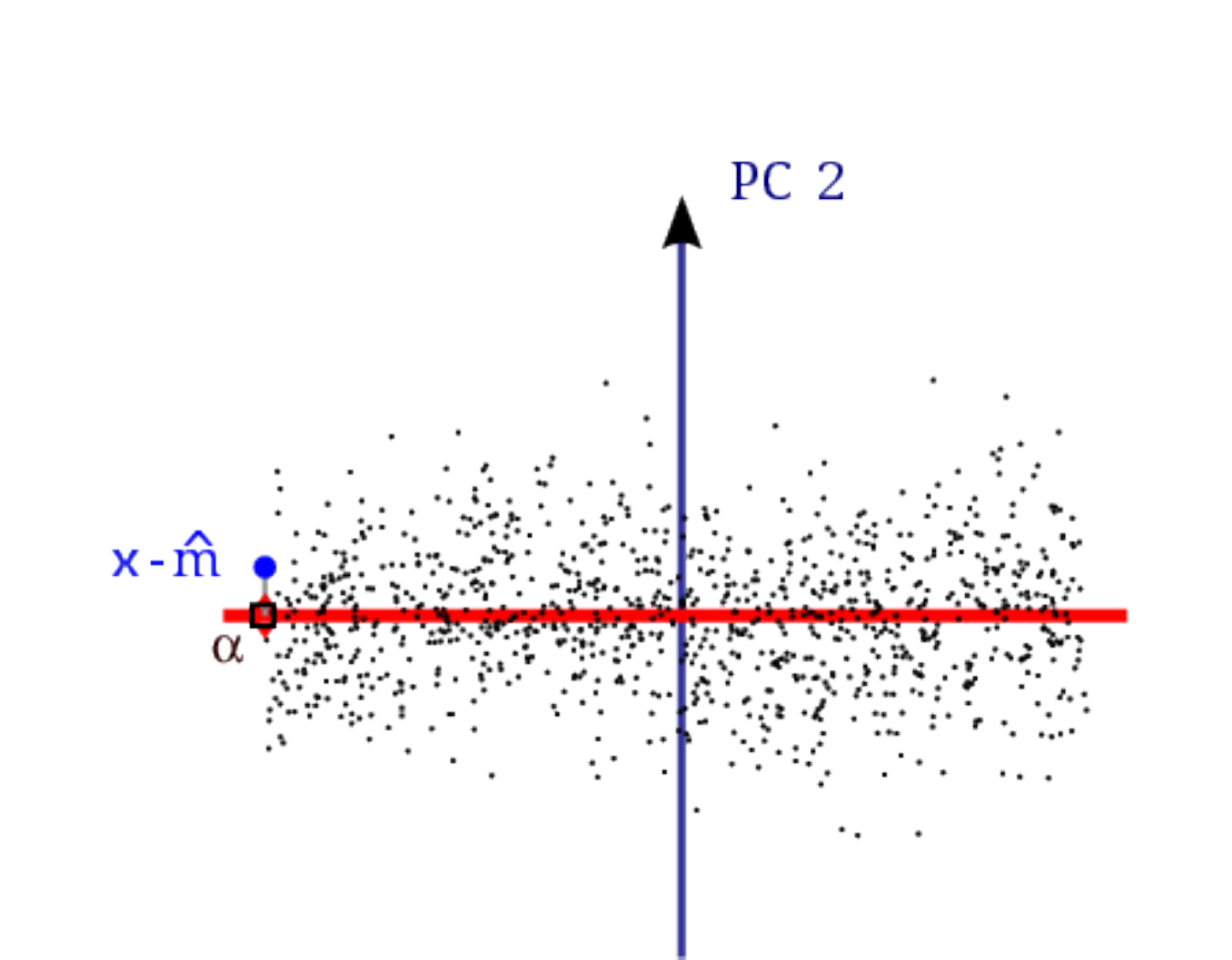}
}
\vspace{-0.0cm}
\caption{The \emph{two operations} in each stage of PPA: projection and subtraction of the polynomial prediction.
Left: input mean-centered data. {An illustrative sample, $\bf{x}$, is highlighted.
This set is not suitable for PCA because it does not fulfil the \emph{conditional mean independence} assumption: the location
of the conditional mean in the subspace orthogonal to PC1 strongly depends on PC1.}
Center: {PCA projection (rotation) and estimation of the conditional mean by a polynomial of degree $2$ (red curve) fitted to minimize
the residual $|\bf{x}-\hat{\bf{m}}| \, \, \, \forall \bf{x}$}.
The black square ($\alpha$) is the projection of $\bf{x}$ onto PC1.
The diamond (in red), $\hat{\bf{m}}$, in the curve represents the estimated conditional mean of $\bf{x}$ predicted from the projection $\alpha$.
The advantage of the polynomial with regard to the straight line is that it accounts for what can
be nonlinearly predicted. Right: the data after removing the estimated conditional mean (PPA solution).
See the on-line paper for color figures.}
\label{fig2}
\vspace{-0cm}
\end{figure*}

The proposed PPA method can be motivated {by considering the conditional mean of the data.}
In essence, PCA is optimal for dimensionality reduction in {a} mean square error (MSE) sense \emph{if and only if} the conditional mean in each principal component is constant along the considered dimension. Hereafter, we will refer to this as the \emph{conditional mean independence} assumption. Unfortunately, this symmetry requirement does not apply in general, as many datasets live in non-Gaussian and/or curved manifolds. See for instance the data in Fig.~\ref{fig2} (left): the dimensions have a nonlinear relation even after PCA rotation (center). In this situation, the mean of the second principal component given the first principal component can be easily expressed with a parabolic function (red line). For data manifolds lacking the required symmetry, nonlinear modifications of PCA should remove the residual nonlinear dependence.

Following the previous intuition, PPA aims to remove the condition mean.
Left panel in Fig.~\ref{fig2} shows the input $2d$ data distribution, where we highlight
a point of interest, $\x$. PPA is a sequential algorithm (as PCA) that transforms one dimension at each step in the sequence.
The procedure in each step consists of two operations.
The \emph{first operation} looks for the best vector for data projection.
Even though different possibilities will be considered later (Section 2.3), a convenient choice for this operation
is the leading eigenvector of PCA. Figure~\ref{fig2}[middle] shows the data after this projection: although the linear
dependencies have been removed, there are still relations between the first and the second data dimensions.
The \emph{second operation} consists in subtracting the conditional mean to every sample. The conditional mean is estimated by fitting
a curve predicting the residual using the projections estimated by the first operation.

{This step, composed of the two operations above, describes the} $d$-dimensional data along \emph{one} curvilinear dimension
through (1) a projection score onto certain leading vector, and (2) a curve depending on the projection score.
PPA differs from PCA in this second operation because it bends the straight line into a curve{, thus} capturing part of the nonlinear relations between the leading direction and the orthogonal subspace.
Since this example is two-dimensional, PPA ends after one step. However, when there are more dimensions, the two-operations are repeated for the remaining dimensions.
At the first step, the $(d-1)$-dimensional information still to be described is the departure from the curve in the subspace orthogonal to the leading vector.
This {data of} reduced dimension is the input for the next step in the sequence.
The last PPA dimension will be the $1d$ residual which, in this example, corresponds to the {residuals in the} second dimension.

\subsection{\em Outline of the paper}

The paper is organized as follows.
Section 2 {formalizes the forward PPA transform and analytically proves that PPA
generalizes PCA and improves its performance in dimensionality reduction.
The objective function of PPA, its restrictions, and its computational cost are
then analyzed.
Section 3 studies the main properties of PPA: Jacobian, volume preservation, invertibility, and metric.
In Section 4 we discuss the differences between PPA and related work.}
{In Section 5, we check the generalization of Mahalanobis distance using the PPA metric,
and its ability to characterize the manifold geometry (curvature and torsion).}
Finally, we report results on standard databases for dimensionality and redundancy reduction.
Section 6 concludes the paper.
{Additionally, the appendix details a step-by-step example of the forward transform.}



\section{Principal Polynomial Analysis} \label{ppa}

{In this section, we start by reviewing the PCA formulation as a deflationary (or sequential) method
that addresses one dimension at a time. This is convenient since it allows to introduce PPA as
the generalization that uses polynomials instead of straight lines in the sequence.}

\subsection{\em The baseline: Principal Component Analysis}

Given a $d$-dimensional centered random variable $\x$, the PCA transform, ${\bf R}$, maps data from the input domain, $\mathcal{X} \subseteq \mathbb{R}^{d \times 1}$, to a response domain, $\mathcal{R} \subseteq \mathbb{R}^{d \times 1}$.
PCA can be actually seen as a sequential mapping (or a set of concatenated $d-1$ transforms). Each transform in the sequence explains a single dimension of the input data by computing a single component of the response:
{\small
\begin{equation}
       \left(
         \begin{array}{c}
            \\
            \\
           \x_0 \\
            \\
            \\
         \end{array}
       \right)
        \buildrel {\bf R}_1 \over \longrightarrow
       \left(
         \begin{array}{c}
            \\
           \alpha_1 \\
            \\
           \x_1 \\
            \\
         \end{array}
       \right)
        \buildrel {\bf R}_{2} \over \longrightarrow
       \left(
         \begin{array}{c}
            \alpha_1\\
            \alpha_2\\
                    \\
                \x_2 \\
                    \\
         \end{array}
       \right)
        \cdots
        \buildrel {\bf R}_{d-1} \over \longrightarrow
              \left(
         \begin{array}{c}
            \alpha_1\\
            \alpha_2\\
                 \vdots   \\
            \alpha_{d-1} \\
                 \x_{d-1} \\
         \end{array}
       \right),
       \label{seq_squeme}
\end{equation}
}
and hence the PCA transformation can be expressed as:
${\bf R} = {\bf R}_{d-1} \circ {\bf R}_{d-2} \circ \cdots \circ {\bf R}_{2} \circ {\bf R}_{1}$.
Here vectors, $\x_p$, and transforms, ${\bf R}_p$, refer to the $p$-th step of the sequence.
Each of these elementary transforms, ${\bf R}_{p}$, acts only on part of the dimensions of the output of the
previous transform: the residual, $\x_{p-1}$. Subscript $p=0$ refers to the input data so $\x_0 = \x$.
This sequential (deflationary) interpretation, which is also
applicable to PPA as we will see later, is convenient to derive most of the properties of PPA in Section 3.

In PCA, each transform ${\bf R}_{p}$:
{(1) $\alpha_p$, which is the projection of the data coming from the previous step, $\x_{p-1}$, onto the unit norm vector $\e_p$; and (2) $\x_p^{\mathrm{PCA}}$, which are the residual data for the next step, obtained by projecting $\x_{p-1}$ {in} the complement space:
\vspace{-0.3cm}
\begin{eqnarray}
        \alpha_p  &=& \e_p^\top \x_{p-1} \nonumber \\
       \x_p^{\mathrm{PCA}} &=& \E_p^\top \x_{p-1},
\label{PCAapprox}
\end{eqnarray}
where $\E_{p}^\top$ is a $(d-p)\times(d-p+1)$ matrix containing the remaining set of vectors. }
In PCA, $\e_p$ is the vector that maximizes the variance of the projected data:
\begin{equation}
      \e_p = \arg\max_{\e}{\{ \mathbb{E} [(\e^{\top} \x_{p-1})^2] \}},
\label{eq:orthoerror}
\end{equation}
{where $\e \in \mathbb{R}^{(d-p+1) \times 1}$ represents the set of possible unit norm vectors. $\E_p^\top$ can be any matrix that spans the subspace orthonormal to $\e_p$, and its rows contain $d-p$ orthonormal vectors. Accordingly, $\e_p$ and $\E_p$ fulfil:
\begin{eqnarray}
       \E_p^\top \e_p &=& \boldsymbol{\O} \nonumber \\
       \E_p^\top \E_p &=& \I_{(d-p) \times (d-p)},
\label{OrthogRelat}
\end{eqnarray}
which will be referred to as the {\em orthonormality relations} of $\e_p$ and $\E_p$ in the discussion below.}

In the $p$-th step of the sequence, the data yet to be explained is $\x_{p}$.
Therefore, truncating the PCA expansion at dimension $p$ implies ignoring the information contained in $\x_p$ so that the dimensionality reduction error is:
\begin{equation}
   \mathrm{MSE}^{\mathrm{PCA}}_p = \mathbb{E} [ \| \E_p^\top \x_{p-1} \|_2^2] = \mathbb{E} [ \|\x_{p} \|_2^2].
   \label{PCAerror}
\end{equation}
PCA is the optimal linear solution for dimensionality reduction in MSE terms since Eq.~\eqref{eq:orthoerror} implies minimizing the dimensionality reduction error in Eq.~\eqref{PCAerror} due to the orthonormal nature of the projection vectors $\e_p$ and $\E_p$.

\subsection{\em The extension: Principal Polynomial Analysis}
\label{the_extension}

{PPA removes the conditional mean in order to reduce the reconstruction error of PCA in Eq.~\eqref{PCAerror}.
When the data fulfill the \emph{conditional mean independence} requirement, the conditional mean at every point in the $\e_p$ direction
is zero. In this case, the data vector goes through the means in the subspace spanned by $\E_p$, resulting in a small PCA truncation error.}
However, this is not true in general (cf.~Fig.~\ref{fig2}) and then the conditional mean $\m_p = \mathbb{E} [\x_p | \alpha_p] \neq 0$.
In order to remove the conditional mean $\m_p$ from $\x_p$, PPA modifies the elementary PCA transforms in Eq.~\eqref{PCAapprox} by subtracting an estimation of the conditional mean, $\hat \m_p$:
\begin{eqnarray}
	\alpha_p &=& \e_p^\top \x_{p-1} \nonumber \\
    \x_{p}^{\mathrm{PPA}} &=& \E_p^\top \x_{p-1} - \hat\m_p
    \label{PPAapprox}
\end{eqnarray}
{Assuming for now that the leading vector, $\e_p$, is computed in the same way as in PCA, PPA only differs from PCA in
the second operation of each transform ${\bf R}_{p}$ (cf.~Eq.~\eqref{PCAapprox}). However, this suffices to ensure the superiority of PPA over PCA.} We will refer to this particular choice of $\e_p$ as the \emph{PCA-based solution} of PPA. In Section 2.3, we consider more general solutions to optimize the objective function at the cost of facing a non-convex problem. In any case, and independently of the method used to choose $\e_p$, the truncation error in PPA is:
\begin{equation}
   \mathrm{MSE}^{\mathrm{PPA}}_p = \mathbb{E} [\| \E_p^\top \x_{p-1} - \hat\m_p \|_2^2].
   \label{PPAerror}
\end{equation}

\paragraph{Estimation of the conditional mean at step $p$.} The conditional mean can be estimated with any regression method $\hat{\m}_p = g(\alpha_p)$. In this work, we propose to estimate the conditional mean at each step of the sequence using a polynomial function with coefficients $w_{p_{ij}}$ and degree $\gamma_p$. Hence, the estimation problem becomes:
\vspace{0.0cm}

{\footnotesize
\begin{equation}
  \hat{\m}_p\\
=
 \begin{pmatrix}
  w_{p_{11}} & w_{p_{12}} & \cdots & w_{p_{1(\gamma_p+1)}} \\
  w_{p_{21}} & w_{p_{22}} & \cdots & w_{p_{2(\gamma_p+1)}} \\
  w_{p_{31}} & w_{p_{32}} & \cdots & w_{p_{3(\gamma_p+1)}} \\
  \vdots & \vdots & \ddots & \vdots \\
  w_{p_{(d-p)1}} & w_{p_{(d-p)2}} & \cdots & w_{p_{(d-p)(\gamma_p+1)}}
 \end{pmatrix}
	\begin{pmatrix}
   1 \\
   \alpha_p \\
  \alpha_p^2 \\
  \vdots \\
\alpha_p^{\gamma_p} \\
 \end{pmatrix},
 \label{mean_estimation}
\end{equation}
}

\vspace{0.0cm}
which, in matrix notation is
$\hat{\m}_p = \W_p \bold{v}_p,$
where $\W_p \in \mathbb{R}^{(d-p) \times (\gamma_p+1)}$, and $\bold{v}_p = [1, \alpha_p,\alpha_p^2 ,\ldots,\alpha_p^{\gamma_p}]^\top$.

{Note} that when considering $n$ input examples, we may stack them column-wise in a matrix $\X_0 \in \mathbb{R}^{d \times n}$.
{In the above mentioned \emph{PCA-based solution}, the $p$-th step of the PPA sequence starts by computing PCA on $\X_{p-1}$. Then,} we use the first eigenvector of the sample covariance
as leading vector $\e_p$, and the remaining eigenvectors as $\E_p$. These eigenvectors are orthonormal; if a different {strategy} is used to find $\e_p$, then $\E_p$ can be chosen to be any orthonormal complement of $\e_p$ (see Section 2.3).
From the projections of the $n$ samples onto the leading vector (i.e. from the $n$ coefficients $\alpha_{p, k}$ with $k=1,\ldots,n$),
we build the Vandermonde matrix $\V_p \in \mathbb{R}^{(\gamma_p +1) \times n}$, by stacking the $n$ column vectors $\bold{v}_{p \,k}$,
with $k=1,\ldots,n$.

Then, {the least squares solution for the matrix $\W_p$ of coefficients of the polynomial is}:
\begin{equation}
	\W_p = (\E_p^\top \X_{p-1}) \V_p^{\dag},
\label{matrix_W_ls}
\end{equation}
where ${\dag}$ stands for the pseudoinverse operation.
Hence, the estimation of the conditional mean for all the samples, column-wise stacked in matrix $\hat{\M}_p$, is:
\begin{equation}
	\hat{\M}_p = \W_p \V_p,
    \label{matrix_estimate}
\end{equation}
and the residuals for the next step are, $\X_p^{\mathrm{PPA}} = \E_p^\top \X_{p-1} - \hat{\M}_p$.

{Summarizing, the extra elements with respect to PCA are a single matrix inversion in Eq.~\eqref{matrix_W_ls}
and the matrix product in Eq.~\eqref{matrix_estimate}. Also note that the estimation of the proposed polynomial
is much simpler than fitting a polynomial depending on a natural parameter such as the orthogonal projection on the curve,
as one would do according to the classical Principal Curve definition~\cite{Has89}.
Since the proposed {objective} function in Eq.~\eqref{PPAerror} does not estimate distortions orthogonal to the curve but {rather} those orthogonal to the leading vector $\e_p$, the computation of the projections is straightforward and decoupled from the computation of $\W_p$.}
The proposed estimation in Eq.~\eqref{matrix_W_ls} consists of  $d-p$ separate univariate problems only: this means that PPA needs to fit $d-p$ one-dimensional polynomials depending solely on the (easy-to-compute) projection parameter $\alpha_p$. Since lots of samples $n \gg \gamma_p+1$ are typically available, the estimation of such polynomials is usually robust.
{The convenience of this decoupling is illustrated in the step-by-step example presented in the appendix.}
Since we compute $\W_p$ using least squares, we obtain three important properties:
\vspace{0.1cm}

{
\begin{property}
The PPA error does not depend on the particular selection of the basis $\E_p$ if it satisfies the orthonormality relations in Eq.~\eqref{OrthogRelat}.\\
\end{property}
{\bf Proof:}
Using different basis $\E'_p$ in the subspace orthogonal to $\e_p$
is equivalent to applying an arbitrary $(d-p)\times(d-p)$ rotation matrix, ${\bf G}$, to the difference vectors
expressed in this subspace in Eq.~\eqref{PPAerror}:
$\mathrm{MSE}^{\mathrm{PPA}}_p({\bf G}) = \mathbb{E} [ \left( {\bf G} ( \E_p^\top \x_{p-1} - \hat\m_p ) \right)^\top {\bf G} ( \E_p^\top \x_{p-1} - \hat\m_p )]$. Since ${\bf G}^\top {\bf G} = \I$, the error is independent of this rotation, and hence independent of the basis.
}
\vspace{0.1cm}

{
\begin{property}
The PPA error is equal to or smaller than the PCA error.\\
\end{property}
{\bf Proof:} The PPA Eqs.~\eqref{PPAerror} and~\eqref{matrix_W_ls} reduce to PCA Eq.~\eqref{PCAerror} in the restricted case of $\W_p=\boldsymbol{\O}$. Since, in general, PPA allows for $\W_p \neq \boldsymbol{\O}$, this implies that $\mathrm{MSE}^{\mathrm{PPA}}_p  \leq \mathrm{MSE}^{\mathrm{PCA}}_p$. Even though the superiority of PPA over PCA in MSE terms is clearer when taking $\e_p$ as in PCA, this property holds in general. If a better choice for $\e_p$ is available, it would reduce the error while having no negative impact in the cost function,
since it is independent from the basis $\E_p$ chosen (see \emph{Property 1} above).
}
\vspace{0.1cm}

{
\begin{property}
PPA reduces to PCA when using first degree polynomials (i.e. straight lines).\\
\end{property}
{\bf Proof:} In this particular situation ($\gamma_p = 1,~~\forall p$), the first eigenvector of $\X_{p-1}$ is the best direction to project onto~\cite{Jolliffe02}. Additionally, when using first degree polynomials, $\V_p$ is very simple and $\V_p^\dag$ can be computed analytically. Plugging this particular $\V_p^\dag$ into Eq.~\eqref{matrix_W_ls}, it is easy to see that
$\W_p=\boldsymbol{\O}$ since the data is centered and $\alpha_p$ is decorrelated of $\E_p^\top \x_{p-1}$. Therefore, when using straight lines $\W_p$ vanishes and PPA reduces to PCA.
}

\vspace{0.1cm}
{Finally, also note that, as in any nonlinear method,
in PPA there is a trade-off between the flexibility to fit the training data
and the generalization ability to cope with new data.
In PPA, this can be easily controlled selecting the polynomial degree $\gamma_p$.
This can be done through standard cross-validation (as in our experiments),
or by using any other model selection procedure such as leave-one-out or (nested) $v$-fold cross-validation.
Note that this parameter is also interpretable and easy to tune, since it controls the flexibility of
the curves or the reduction of PPA to PCA in the $\gamma=1$ case.}

\subsection{\em PPA cost function: alternative solutions and optimization problems}\label{leading_vector}

\begin{figure*}[t!]
\begin{center}
\vspace{-0.5cm}
\begin{tabular}{ccc}
a & b & c\\
\includegraphics[width=1.3cm]{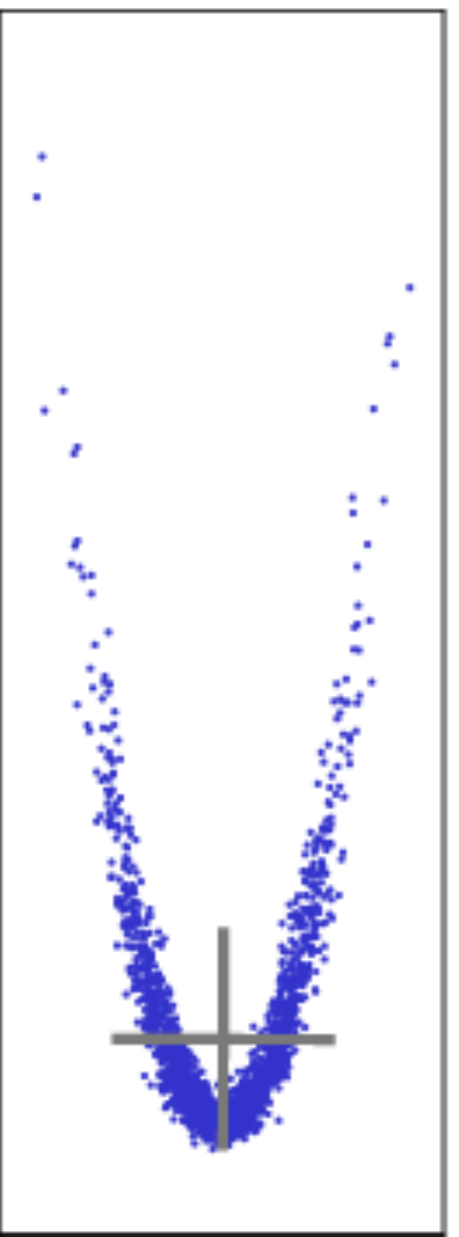} \hspace{-0.0cm}
\includegraphics[width=1.3cm]{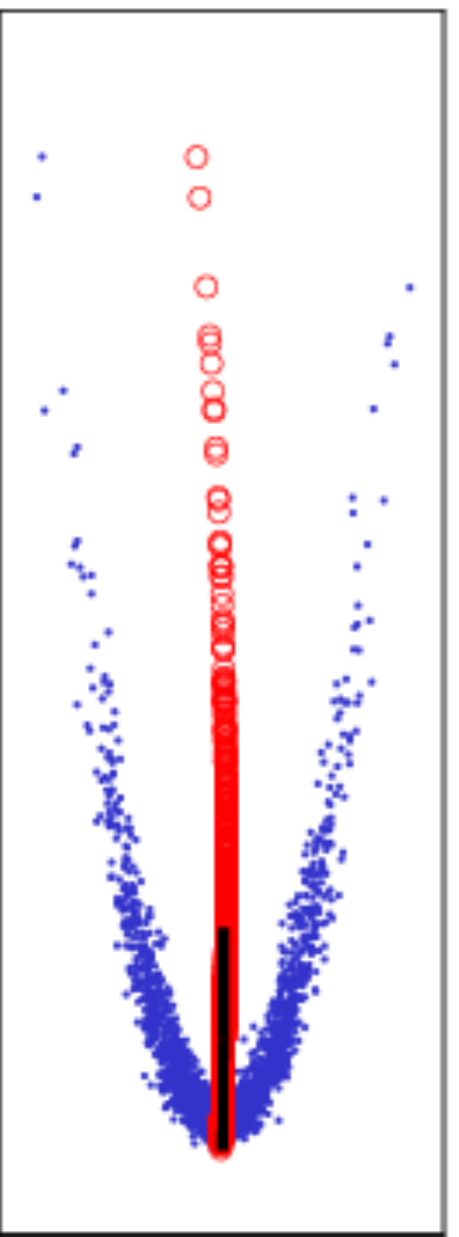} \hspace{-0.0cm}
\includegraphics[width=1.3cm]{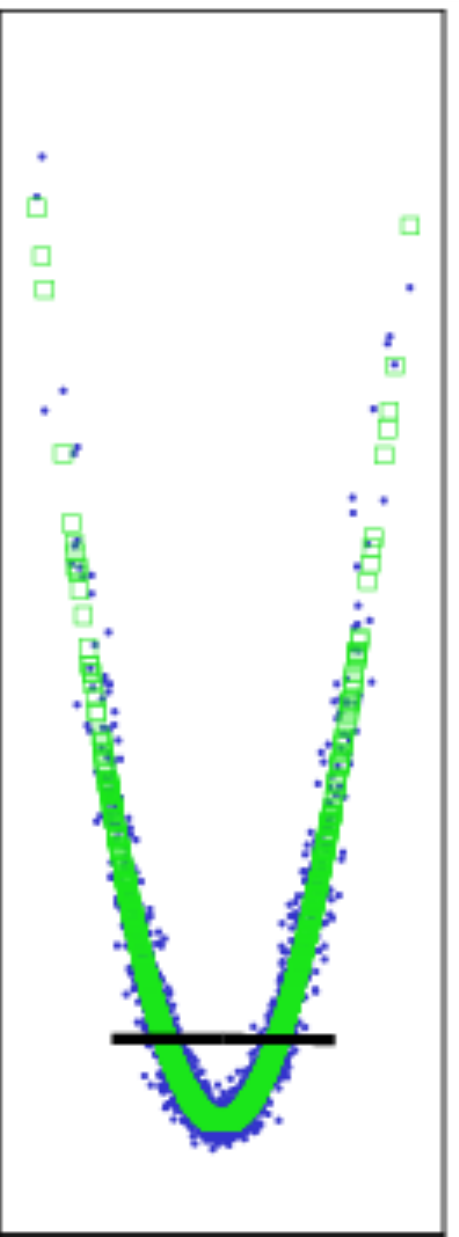} \hspace{0.0cm}
&
\includegraphics[width=4.5cm]{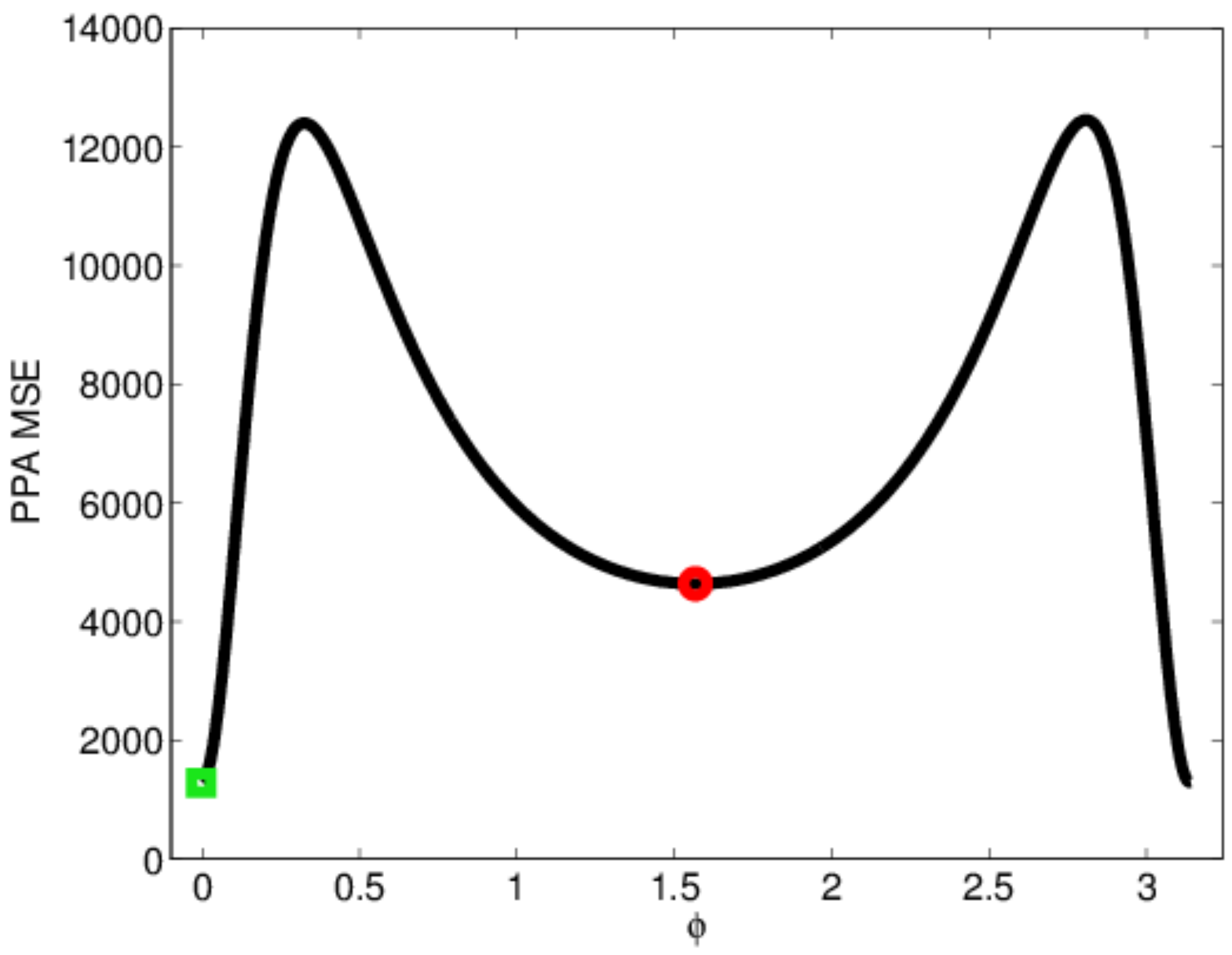}
&
\includegraphics[width=1.3cm]{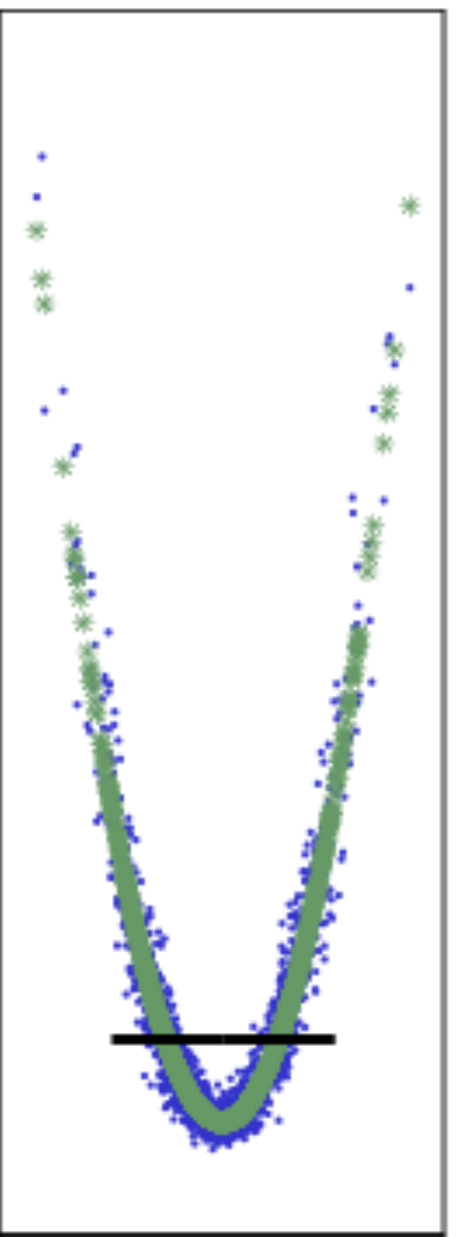} \hspace{-0cm}
\includegraphics[width=1.3cm]{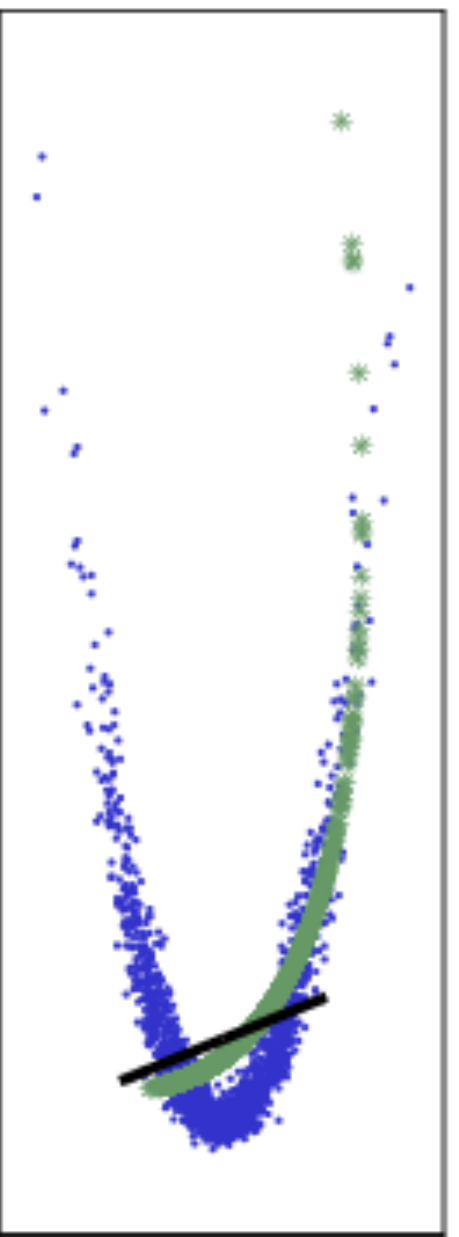} \hspace{-0cm}
\includegraphics[width=1.3cm]{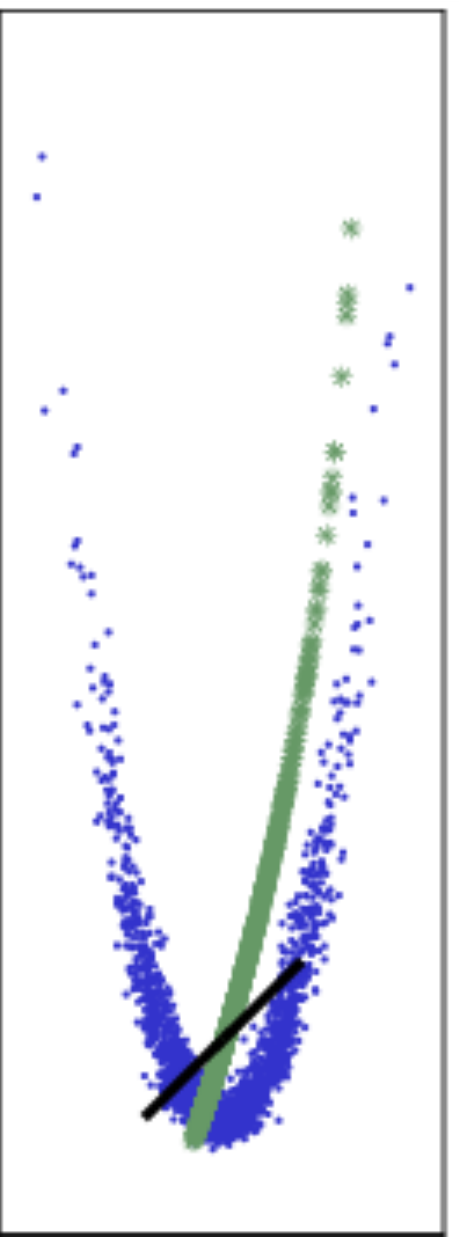} \hspace{-0cm}
\includegraphics[width=1.3cm]{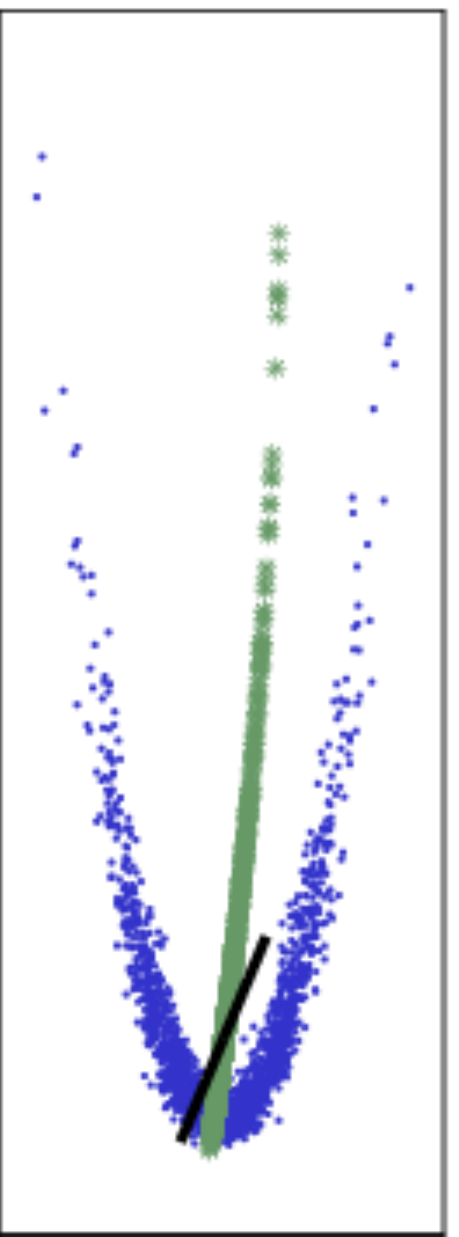}\hspace{-0cm}
\includegraphics[width=1.3cm]{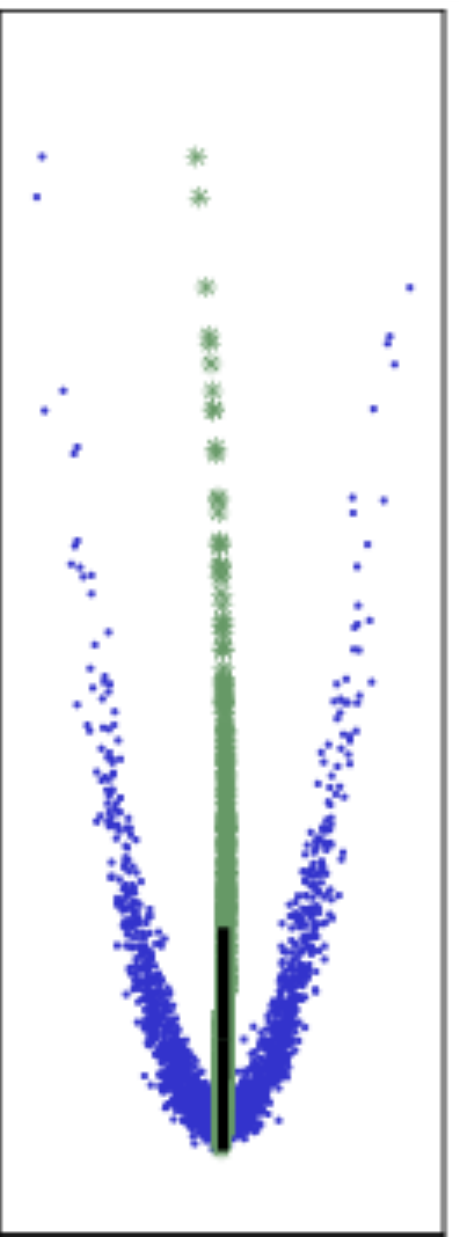}
\end{tabular}
\end{center}
\caption{\small PPA objective is non-convex. (a) Samples drawn from a noisy parabola (blue) and the eigenvectors of the covariance matrix, PC1 and PC2 (in gray). The PPA parabolas obtained from projections onto PC1 (PCA-based solution) and onto the PC2 are plot in \red{$\circ$} and \green{$\Box$} respectively. (b) Dimensionality reduction error, $f(\e)$, for $\e_p$ vectors with different orientation $\phi$, where $\phi=0$ corresponds to PC2 (\green{$\Box$}) and $\phi=\frac{\pi}{2}$ corresponds to PC1 (\red{$\circ$}). (c) Fitted PPA parabolas (\green{$\ast$}) for a range of orientations of the corresponding $\e_p$ (in black).}
\label{non-convex}
\end{figure*}

By construction PPA improves the dimensionality reduction performance of PCA when using the
restricted PCA-based solution.
Here we show that better solutions for the PPA cost function may exist, but unfortunately are not easy to obtain.
Possible improvements would involve (1) alternative functions to estimate
the conditional mean, and (2) more adequate projection vectors $\e_p$.

Better estimations of the conditional mean can be obtained with prior knowledge about the system that generated the data. For instance, if one knows that samples should follow an helical distribution, a linear combination of sinusoids could be a better choice. Even for these cases, least squares would obtain the weights of the linear combination. Nevertheless, in this work, we restrict ourselves to polynomials since they provide flexible enough solutions by using the appropriate degree. Below we show that one can fit complicated manifolds, e.g. helices, with generic polynomials. More interestingly, geometric descriptions of manifold, such as curvature or torsion, can be computed from the PPA model despite being functionally different from the actual generative model.

The selection of appropriate $\e_p$ is more critical, since
\emph{Property 1} implies that MSE does not depend on $\E_p$, but only on $\e_p$.
The cost function for $\e_p$ measuring the dimensionality reduction error is $f(\e)$:
\begin{eqnarray}
\label{PPA_cost}
	\nonumber \e_p & = & \arg\min_{\e}{f(\e)} = \arg\min_{\e}{ \mathbb{E} [ \| \E_p^\top \x_{p-1} - \W_p \bold{v}_p \|_2^2]}, \\
               \text{s.t.} &   & \E_p^\top \E_p = \I \nonumber\\
                           &   & \E_p^\top \e_p = \boldsymbol{\O} \nonumber \\
                           &   & \W_p = (\E_p^\top \X_{p-1}) \V_p^{\dag}     \nonumber.
\end{eqnarray}
This constrained optimization does not have a closed-form solution, and one has to resort to gradient-descent alternatives.
The gradient of the cost function $f(\e)$ is:
\begin{equation}
	\frac{\partial f}{\partial \e_{p_j}}  = \mathbb{E} \bigg[
\sum_{i = 1}^{d-p} 2 (\E_{p_i}^\top \x_{p-1} - \hat{\m}_{p_{i}})
\, \, \, \W_{p_i} \Q \, {\bf v}_{p} \, \x_{(p-1)_{j}} \bigg],
\label{eq:PPA_GD_2}
\end{equation}
where {$\E_{p_i}^\top$ and $\W_{p_i}$ refer to the $i$-th rows of the corresponding matrices,
$\hat{\m}_{p_{i}}$ and $\x_{(p-1)_{j}}$ are the $i$-th and $j$-th components of the corresponding vectors,} and $\Q \in \mathbb{R}^{p \times p}$ is:
\begin{equation}
	\Q =
 \begin{pmatrix}
  0 & 1 & 0 & 0 \cdots & 0 \\
  0 & 0 & 2 & \cdots & 0 \\
  \vdots  & \vdots  & \vdots  & \ddots & \vdots  \\
  0 & 0 & \cdots & 0 & p-1 \\
  0 & 0 & \cdots & 0 & 0
 \end{pmatrix},
\end{equation}

In general, the PPA cost function is non-convex. The properties of $f(\e)$ for the particular dataset at hand will determine the complexity of the problem and the accuracy of the restricted PCA-based solution. Actually, the example in Fig.~\ref{non-convex} shows that, in general, the PCA-based solution for $\e_p$ is suboptimal, and better solutions may be difficult to find given the non-convexity of the cost function. In this $2d$ illustration, the only free parameter is the orientation of $\e_p$. Fig.~\ref{non-convex}(b) shows the values of the error, $f(\e)$, as a function of the orientation of $\e$.
{Since PCA ranks the projection by increasing variance (Eq.~\eqref{eq:orthoerror}), the PCA solution is suboptimal with respect to the one obtained by PPA with gradient descent.}
The first PCA eigenvector does not optimize Eqs.~\eqref{PPAerror} or~\eqref{PPA_cost}. Even worse, the risk of getting stuck into a suboptimal solution is high when using random initialization and simple gradient descent search.

The results in this section suggest that the simple PCA-based solution for $\e_p$ may be improved at the expense of solving a non-convex problem. According to this, in Section 4 we will present results for PPA optimized by using both {the} gradient descent and the PCA-based solution{s}. But in {all cases}, and thanks to \emph{Property 2}, PPA obtains better results than PCA.

\subsection{{\em PPA computational cost}}\label{cputime}

{PPA is computationally more costly than PCA,
which in a na\"ive implementation roughly scales cubically with the problem dimensionality ${\mathcal O}(d^3)$.
In the case of PCA-based PPA, this cost is increased because, in each of the $d-1$ deflationary steps, the
pseudoinverse of the matrix ${\bf V}_p$ has to be computed.
These pseudoinverses involve $d-1$ operations of cost ${\mathcal O}((\gamma+1)^3)$.
Therefore, in total, the cost of PCA-based PPA is ${\mathcal O}(d^3 + (d-1)(\gamma+1)^3)$.}

{If the gradient-descent optimization, Eq.~\eqref{PPA_cost}, is used, the
cost increases substantially since the same problem is solved for a number of
iterations $k$ until convergence, ${\mathcal O}(k(d^3+(d-1)(\gamma+1)^3))$.
The cost associated to this search may be prohibitive in many applications,
but it is still lower than the cost of other generalizations of PCA: kernel-PCA scales with the number of samples,
${\mathcal O}(n^3)$, which is typically larger than the dimensionality $n\gg d$,
and non-analytic Principal Curves are slow to apply since they require
computing $d$ curves per sample.}

\subsection{\em PPA Restrictions}
PPA has two main restrictions that limit the class of manifolds for which PPA is well suited.
First, PPA needs to fit uni-valued functions in each regression in order to ensure the transform is a bijection. This may not be a good solution when the manifold exhibits bifurcations, self-intersections, or holes. {While other (non-analytical) principal curves methods can deal with such complexities~\cite{Kegl02,Ozertem11}, their resulting representations could be ambiguous, since a single coordinate value would map close points, which are far in the input space. This can be in turn problematic to define an inverse function.}\\
Secondly, PPA assumes stationarity along the principal directions as done in PCA. This is not a problem if the data follow the same kind of conditional probability density function along each principal curve. However, such condition does not hold in general. More flexible frameworks such as the Sequential Principal Curves Analysis~\cite{Laparra12} are good alternatives to circumvent this shortcoming, but at the price of a higher computational cost.


\section{Jacobian, invertibility and induced metric} \label{properties}

The most appealing characteristics of PPA (invertibility of the nonlinear transform, its geometric properties and the identified features)
are closely related to the Jacobian of the transform.
This section presents the analytical expression of the Jacobian of PPA as well as the induced properties of volume preservation and invertibility. Then we introduce the analytical expression for the inverse and the metric induced by PPA.

\subsection{\em PPA Jacobian}
\label{sect_Jacobian}

Since PPA is a composition of transforms, cf. Eq.~\eqref{seq_squeme}, its Jacobian is the product of the Jacobians at each step:
\begin{equation}
      \nabla \R(\x) = \prod_{p=d-1}^{1} \nabla \R_p = \nabla \R_{d-1} \nabla \R_{d-2} \cdots \nabla \R_2 \nabla \R_1.
      \label{det_sequence}
\end{equation}
Therefore, the question reduces to compute the Jacobian $\nabla \R_p$ for each elementary transform in the sequence.
Taking into account the expression for each elementary transform in Eq.~\eqref{PPAapprox}, and the way $\m_p$ is estimated in
Eq.~\eqref{matrix_estimate}, simple derivatives lead to:
\begin{equation}\label{individual_jacobian}
{\footnotesize
\nabla \R_p =
\begin{pmat}({|})
     {\bf I}_{(p-1)\times(p-1)}   & {\bf0}_{(p-1)\times(d-p+1)}   \cr\-
     {\bf 0}_{(d-p+1)\times(p-1)} &  \left(
                                 \begin{array}{c}
                                   \e^\top_p \\[2mm]
                                   \E^\top_p \\
                                 \end{array}
                               \right)        -
                               \left(
                                 \begin{array}{c}
                                   {\bf 0}_{1\times(d-p+1)} \\[2mm]
                                   {\bf u}_p \e^\top_p \\
                                 \end{array}
                               \right)
                               \cr
\end{pmat},}
\end{equation}
where ${\bf u}_p = \W_p \dot{\bold{v}}_p$ and $\dot{\bold{v}}_p = [0, 1,2\alpha_p,\ldots,\gamma_p \alpha_p^{\gamma_p-1}]^\top$.
Note that the block structure of the Jacobian of each elementary transform and the identity in the top left block are justified by the fact that
each $\R_p$ only acts on the residual $\x_{p-1}$ of the previous transform, i.e. $\R_p$ does not modify the first $p-1$ components of the previous output.

\begin{figure*}[t!]
\begin{center}
\small
\setlength{\tabcolsep}{7pt}
\begin{tabular}{cccc}
(a) Original data  & (b) Input domain & (c) PPA domain & (d) Whitened PPA domain \\[-0.1cm]
\includegraphics[width=3.51cm]{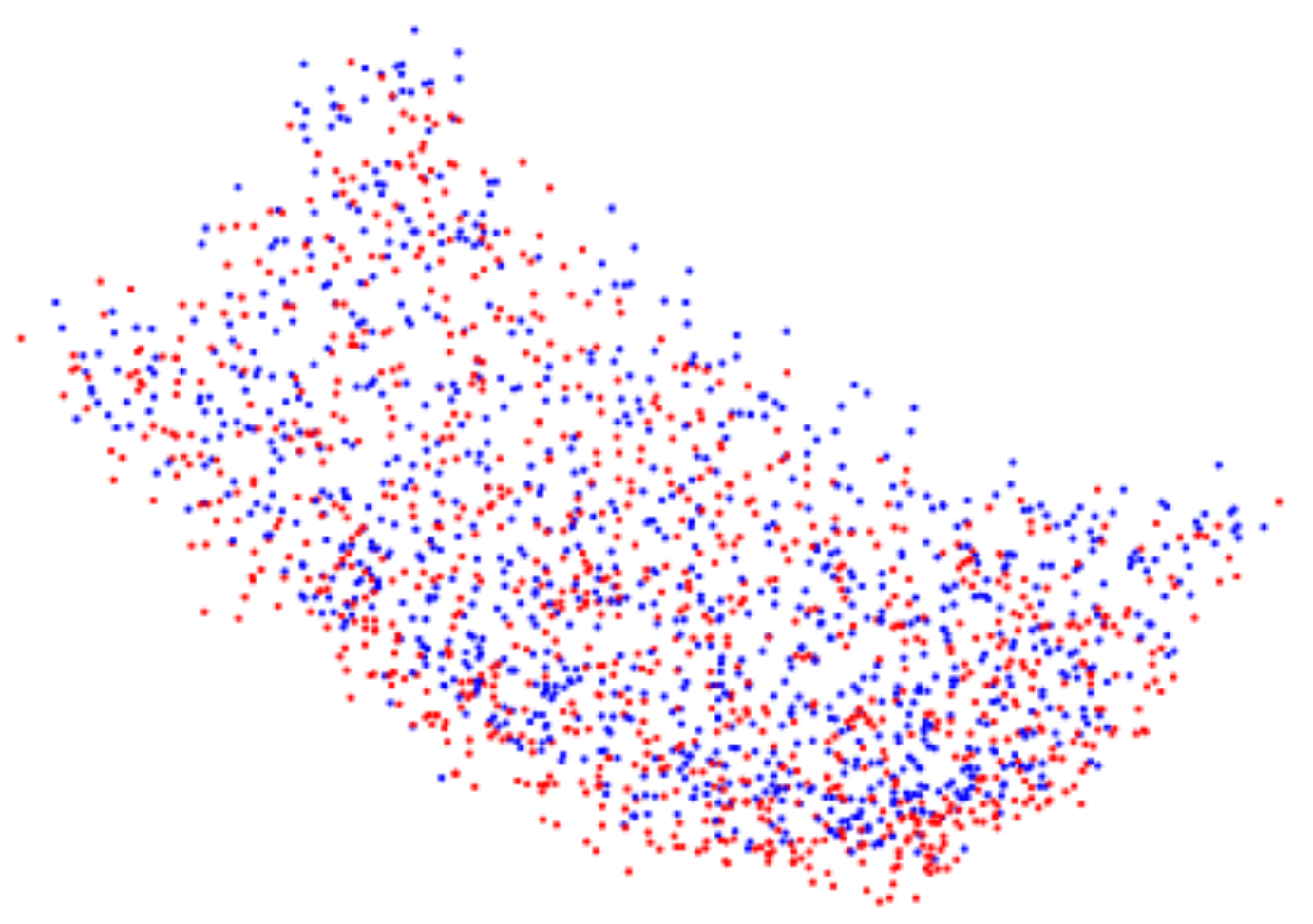} &
\includegraphics[width=3.51cm]{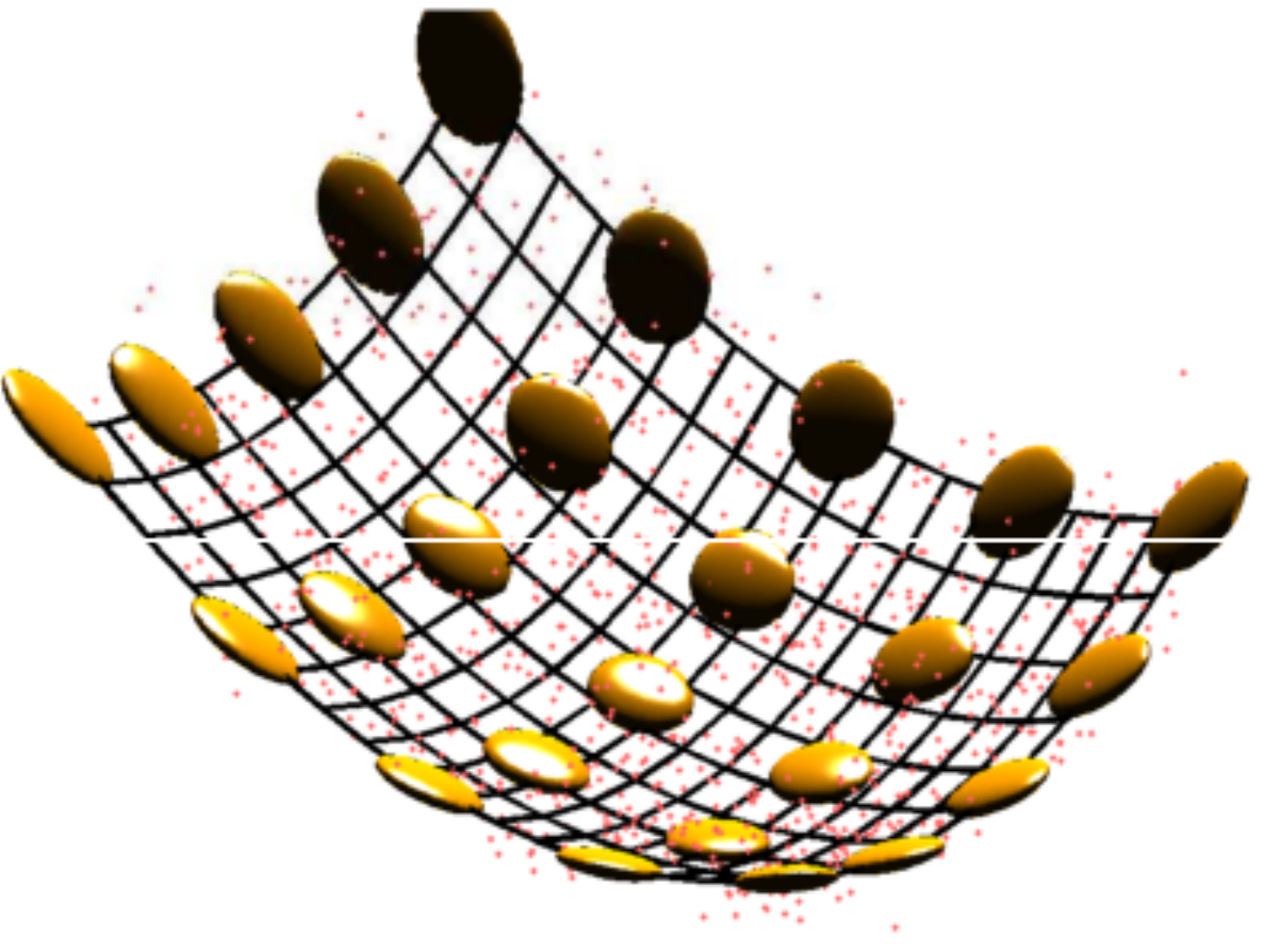} &
\includegraphics[width=4.16cm]{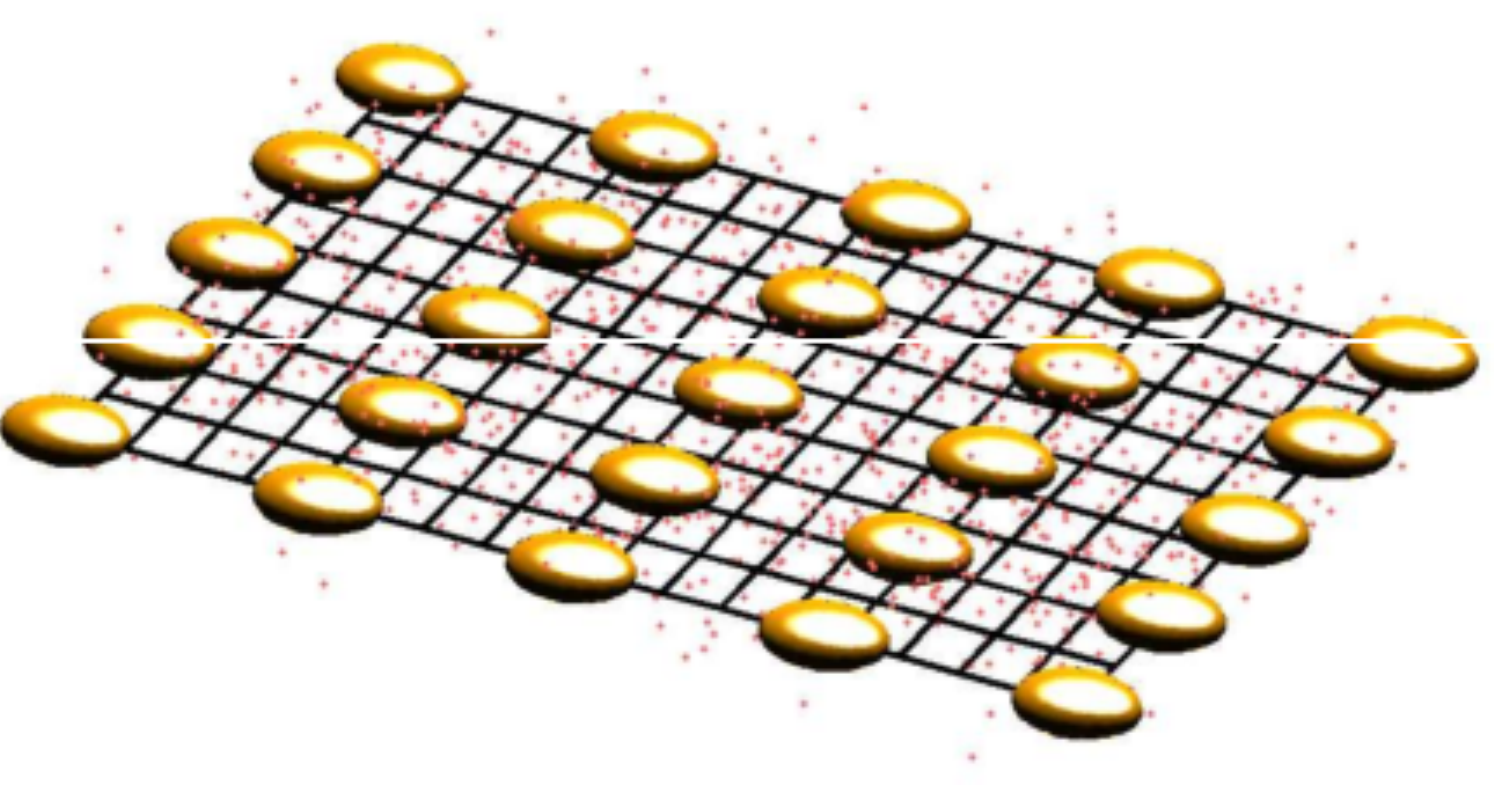} &
\includegraphics[width=2.86cm]{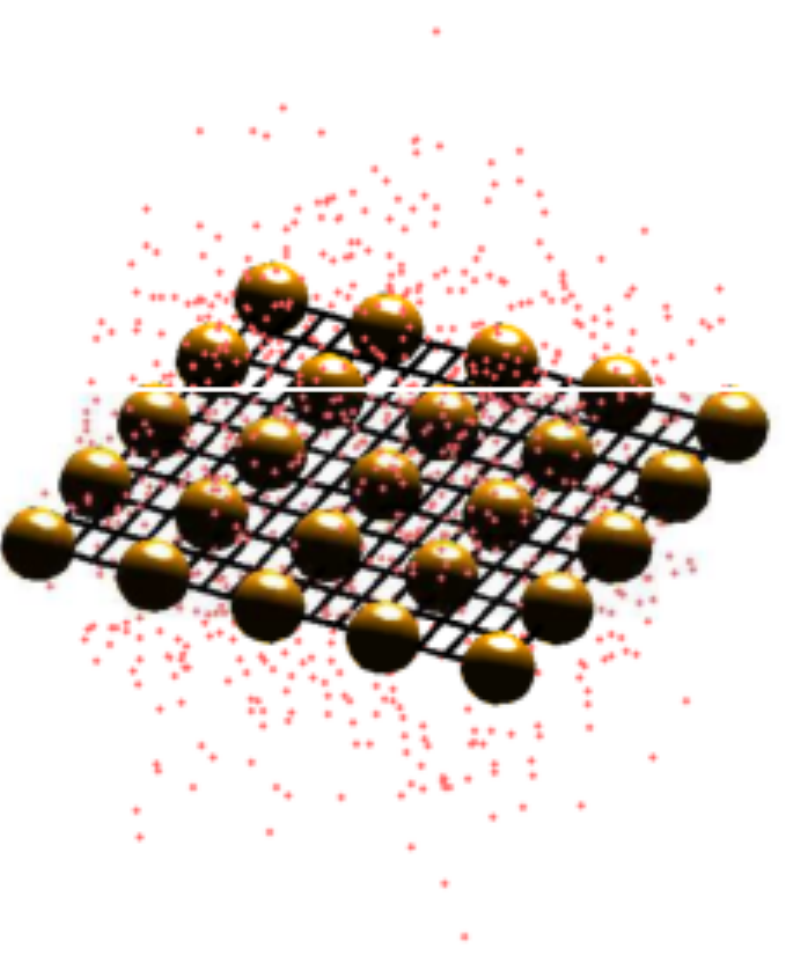} \\
\end{tabular}
\end{center}
\vspace{-0.5cm}
\caption{\small PPA curvilinear features and discrimination ellipsoids based on the PPA metric. (a) Non-linearly separable data. PPA results for the first class data: (b) in the input domain, (c) in the PPA domain, and (d) in the whitened PPA domain, which is included here for the sake of comparison with the Mahalanobis metric. The curvilinear features (black grid) are computed from the polynomials found by PPA, while the unit radius spheres represent the metric  induced by the whitened PPA domain in each domain.} \label{ellipsoids}
\end{figure*}

\subsection{{\em PPA is a volume-preserving mapping}}\label{sec:vol}

{{\bf Proof:}
The volume of any $d$-cube is invariant under a nonlinear mapping $\R$ if $|\nabla \R(\x)|=1$, $\forall \x \in \mathcal{X}$~\cite{Dubrovin82}.
In the case of PPA,
the above is true if $|\nabla \R_p|=1$ for every elementary transform $\R_p$ in Eq.~\eqref{det_sequence}.
To prove this, we need to focus on the determinant of the bottom-right submatrix of $\nabla \R_p$, since $\scriptsize\begin{pmat}|{|}|
     \bf A   & 0   \cr\-
     0 &  \bf B \cr
\end{pmat} = |\bf A| |\bf B|$, where in our case $\bf A$ is the identity matrix.
 Since the determinant of a matrix is the volume of the parallelogram defined by the row vectors in the matrix, the parallelogram defined by the vector $\e_p^\top$ and the vectors in $\E_p^\top$ is a unit volume $(d-p+1)$-cube due to the orthonormal nature of these vectors. The right-hand matrix subtracts a scaled version of the leading vector, ${\bf u}_{p i} \e_p^\top$, to the vector in the $i$-th row of $\E_p^\top$, with $i=[1,\ldots,d-p]$. Independently of weights ${\bf u}_{p i}$, this is a shear mapping of the $(d-p+1)$-cube defined by $\e_p^\top$ and $\E_p^\top$. Therefore, after the subtraction, the determinant of this submatrix is still $1$. As a result $|\nabla \R_p|=1$, and hence $|\nabla \R(\x)|=1,~\forall \x \in \mathcal{X}$.}

\vspace{0.1cm}
Volume preservation is an appealing property when dealing with distributions in different domains. Note that probability densities under transforms depend only on the determinant of the Jacobian: $p_x({\bf x}) = p_y({\bf y}) |\nabla \R({\bf x})|$, for PPA $p_x({\bf x}) = p_y({\bf y})$. A possible use of this property will be shown in sec.~\ref{MI} to compute the multi-information reduction achieved by the transform.

\subsection{\em PPA is invertible}
{{\bf Proof:} A nonlinear transform is invertible if its derivative (Jacobian) exists and it is non-singular $\forall {\bf x}$. This is because,
in general, the inverse can be thought as the integration of a differential equation defined by the inverse of the Jacobian~\cite{Logan94,Epifanio04}. Therefore, the volume preservation property, which ensures that the Jacobian
is non-singular, also guarantees the existence of the inverse.}\\
Here we present a straightforward way to compute the inverse by undoing each of the elementary transforms in the PPA sequence. Given that there is no loss of information in each PPA step, the inverse has perfect reconstruction, i.e. if there is no dimensionality reduction the inverted data is equal to the original one.
Given a transformed point, $\rr = [\alpha_1,\alpha_2,\ldots,\alpha_{d-1},\x_{d-1}]^\top$, and the parameters of the learned transform (i.e. the variables $\e_p$, $\E_p$, and $\W_p$, for $p=1,\ldots,d-1$), the inverse is obtained by recursively applying th{e following} transform:
\begin{eqnarray}
\x_{p-1} =
\begin{pmat}({|})
 & & \cr
\e_p & & \E_p \cr
 & & \cr
\end{pmat}
\begin{pmatrix}
  \alpha_p \\
   \\
  \x_p + \W_p {\bf v}_p \\
\end{pmatrix}
\label{inversa}
\end{eqnarray}

\subsection{\em PPA generalizes Mahalanobis distance} \label{Metric}

When dealing with non-linear transformations, it is useful to have a connection between the metrics (distances) in the input and transformed domains. For instance, if one applies a classification method in the transformed domain, it is critical to understand which are the classification boundaries in the original domain.

Consistently with results reported for other nonlinear mappings~\cite{Epifanio03,Malo06a,Laparra10a,Laparra12},
the PPA-induced distance in the input space follows a standard change of metric under change of coordinates~\cite{Dubrovin82} and can be computed as:
\begin{equation}
       \mathrm{d}^2_{\mathrm{PPA}}(\x,\x+\Delta\x) = \Delta\x^\top {\bf M}(\x) \Delta\x,
       \label{gral_Mah_PPA_distance}
\end{equation}
and the PPA-induced metric ${\bf M}(\x)$ is tied to the Jacobian,
\begin{equation}
       {\bf M}(\x) = \nabla \R(\x)^\top \boldsymbol{\Lambda}_{\mathrm{PPA}}^{-1} \nabla \R(\x)
       \label{gral_Mah_PPA_metric}
\end{equation}
and $\boldsymbol{\Lambda}_{\mathrm{PPA}}$ defines the metric in the PPA domain.
In principle, one can choose $\boldsymbol{\Lambda}_{\mathrm{PPA}}$ depending on the prior knowledge about the problem.
For instance, a classical choice in classification problems is the Mahalanobis metric ~\cite{Mahalanobis36,Duda07}.
{Mahalanobis metric is equivalent to using Euclidean metric after whitening, i.e. after dividing each PCA dimension by
its standard deviation.
One can generalize Mahalanobis metric using PPA by selecting a $\boldsymbol{\Lambda}_{\mathrm{PPA}}$ as a matrix whose diagonal is composed by the variance of each dimension in the PPA domain.
Or analogously, employing the Euclidean metric after whitening the PPA transform.}
Figure~\ref{ellipsoids} shows an example of the unit distance loci induced by the generalized Mahalanobis PPA metric in different domains.
The benefits of this metric for classification will be illustrated in Section~\ref{exp_metric}.


\section{Related Methods} \label{related}

The qualitative idea of generalizing principal components from straight lines to curves is not new. Related work includes approaches based on (1) non-analytical principal curves~\cite{Einbeck05,Zhang10,Einbeck10b,OzertemTesis,Ozertem11,Laparra12}, (2) fitting analytic curves~\cite{Jolliffe02,Donnell94,Besse95}, and (3) implicit methods based on neural networks and autoencoders~\cite{Kramer91,Hinton06,Scholz07} as well as reproducing kernels as in the kernel-PCA~\cite{Scholkopf98}. Here we review the differences between PPA and these approaches.
\paragraph{Non-analytic Principal Curves.}
In the Principal Curves literature, interpretation of the principal subspaces as $d$-dimensional nonlinear representations is only marginally treated in~\cite{Ozertem11,Einbeck10b,OzertemTesis}. This is due to the fact that such subspaces are not explicitly formulated as data transforms. Actually, in~\cite{Ozertem11} the authors acknowledge that, even though their algorithm could be used as a representation if applied sequentially, such an interpretation was not possible at that point since the projections lacked the required accuracy.
{The proposed PPA is closer to the recently proposed Sequential Principal Curves Analysis (SPCA)~\cite{Laparra12} where standard and secondary principal curves~\cite{Has89,Delicado01} are used as curvilinear axes to remove the nonlinear dependence among the input dimensions.
While flexible and interpretable, defining a transformation based on non-parametric Principal Curves (as in SPCA) has two main drawbacks: (1) it is computationally demanding since, in $d$-dimensional scenarios, the framework requires drawing $d$ individual Principal Curves {\em per} test sample, and (2) the lack of analytical form in the principal curves implies non-trivial parameter tuning to obtain the appropriate flexibility of the curvilinear coordinates. To resolve these issues and ensure minimal parameter tuning, we propose here to fit polynomials that estimate the conditional mean along each linear direction.}
We acknowledge that the higher flexibility of methods based on non-parametric Principal Curves suggests possibly better performances than PPA. However, it is difficult to prove such intuition, since, contrarily to PPA, these methods do not provide an analytic solution.

\paragraph{Methods fitting analytic curves.}
\emph{Additive Principal Components} (APC) proposed in~\cite{Donnell94} explicitly fits a sequence of nonlinear functions as {PPA}. {However, the philosophy of their approach differs from Principal Curves since they focus on the low variance features.} In the linear case, sequential or deflationary approaches may equivalently start by looking for features that explain most or least of the variance. However, in the nonlinear APC case, the interpretation of low variance features is very different from the high variance features~\cite{Donnell94}. The high variance features identified by APC do not represent a summary of the data, as Principal Curves do. In the nonlinear case, minimizing the variance is not the same as minimizing the representation error, which is our goal. Therefore, our approach is closer to Principal Curves approaches of the previous paragraph than to APC.

    {Our method also presents a model and minimization of the representation error substantially different to the \emph{Fixed Effect Curvilinear Model} in~\cite{Besse95}.} This difference in the formulation is not trivial since it makes their formulation fully $d$-dimensional, while we restrict ourselves to a sequential framework where $d-1$ polynomials are fitted, one at a time.
    Moreover, the PPA projections onto the polynomial are extracted using the subspace orthogonal to the leading vector, which makes the
    estimation even simpler.
   Additionally, their goal (minimizing the representation error in a nonlinearly transformed domain) is not equivalent to minimizing the dimensionality reduction error in the input space (as it is the case for PPA).

\paragraph{Neural networks and autoencoders.}

{Neural network approaches, namely \emph{nonlinear PCA}~\cite{Kramer91,Jolliffe02,Scholz07} and \emph{autoencoders}~\cite{Hinton06}, share many properties of PPA: they can be enforced to specifically reduce the MSE, are non-linear, invertible, and can be easily applied to new samples~\cite{Scholz05}. However, the nonlinear features are not explicit in the formulation and one is forced to use the inverse of the transformation to visualize the curvilinear coordinates of the identified low dimensional subspace.
Another inconvenience is selecting the network architecture and fitting the model parameters (see~\cite{Scholz12} for a recent review), upon which  the regularization ability of the network depends. The number of hidden units is typically {assumed to be} higher than the dimensionality of the input space, but there is still no clear way to set the network beforehand. As opposed to more explicit methods (PPA or SPCA), the curvature of the $d$ dimensional dataset is not encoded using $d$ nonlinear functions with different relevance, which makes the geometrical analysis difficult.}

\begin{figure*}[t]
\begin{center}
\small
\setlength{\tabcolsep}{1pt}
\vspace{-0cm}
\begin{tabular}{cccc}
\includegraphics[width=3.95cm]{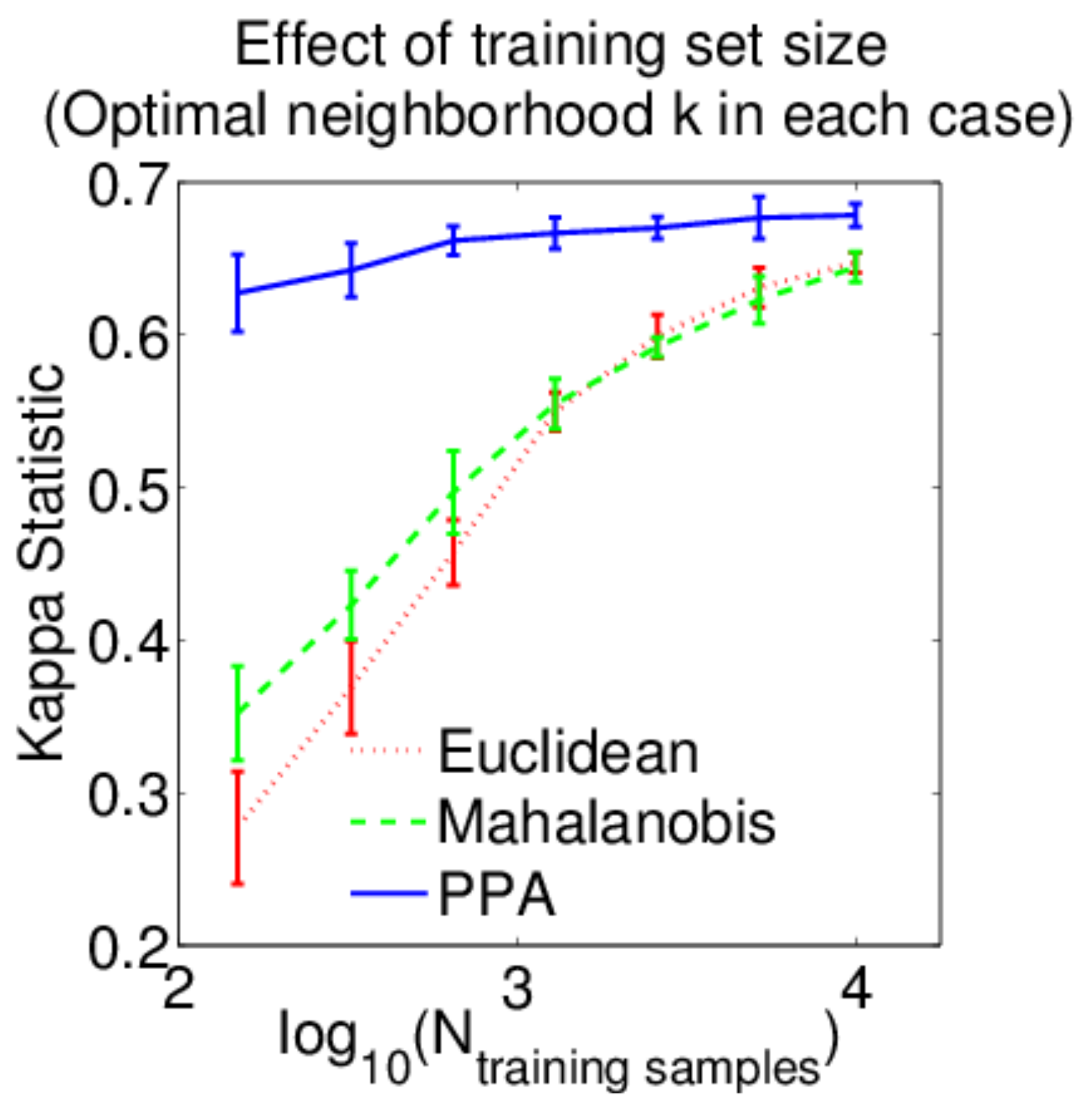} &
\includegraphics[width=3.7cm]{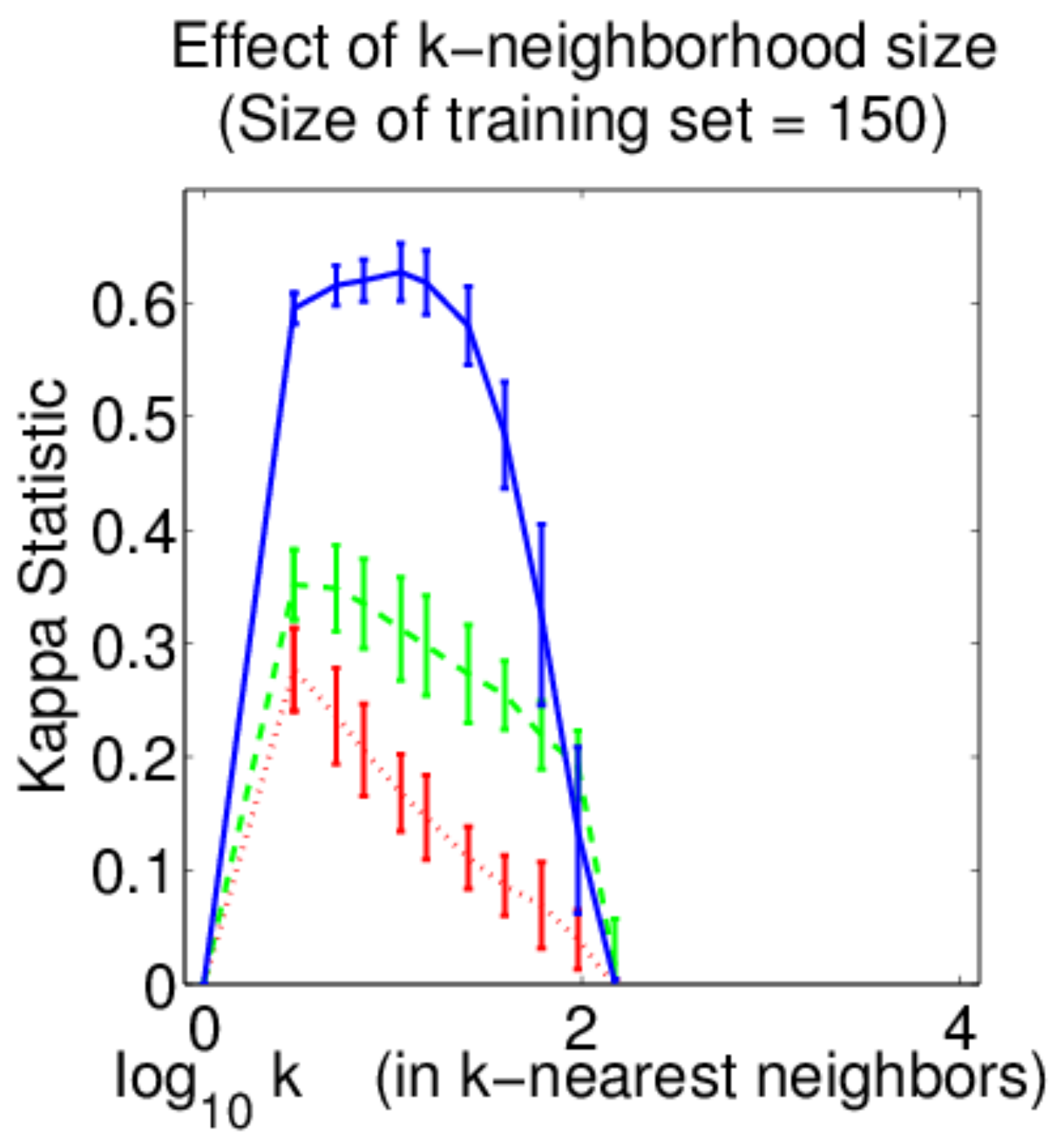} &
\includegraphics[width=3.7cm]{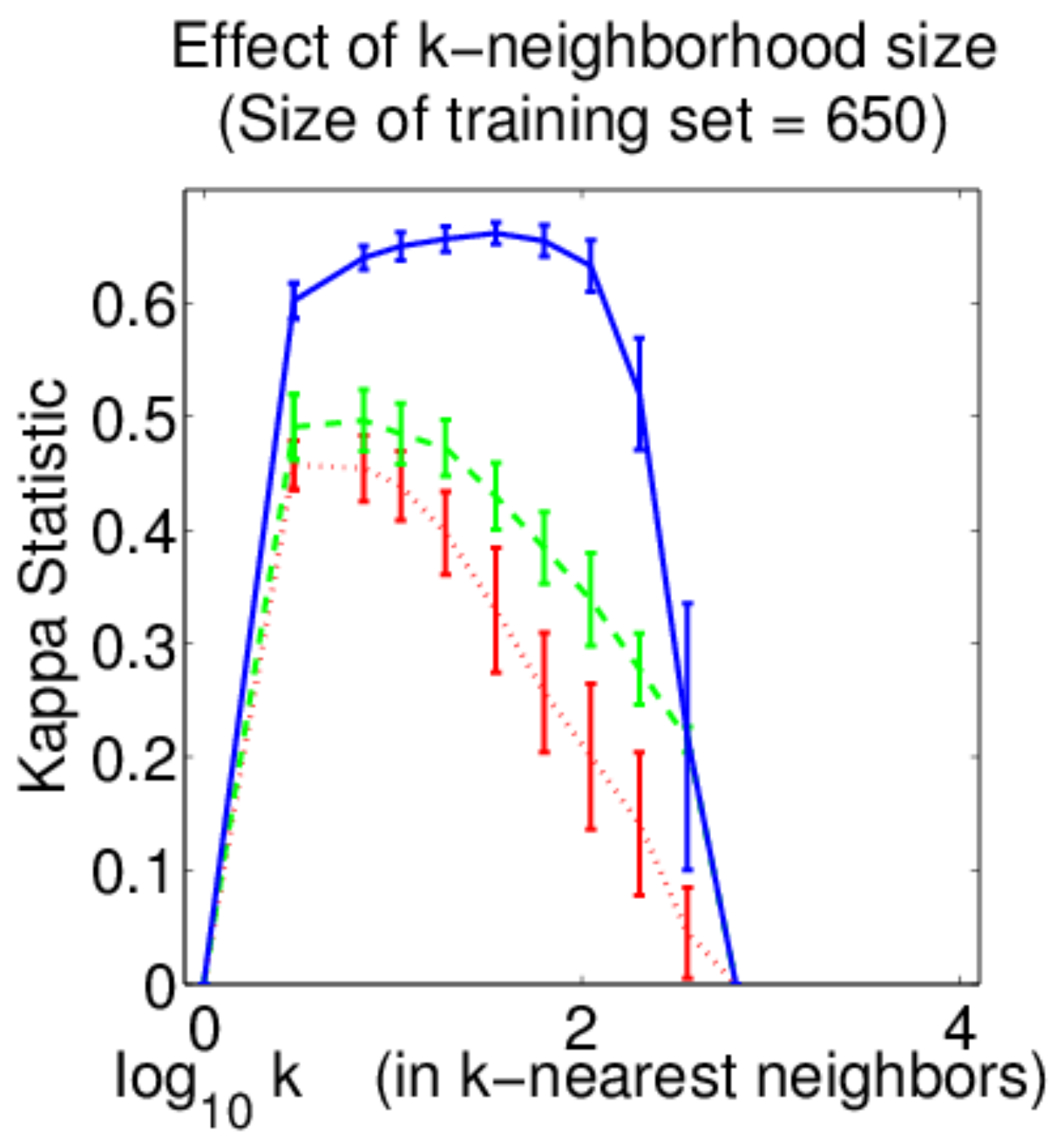} &
\includegraphics[width=3.7cm]{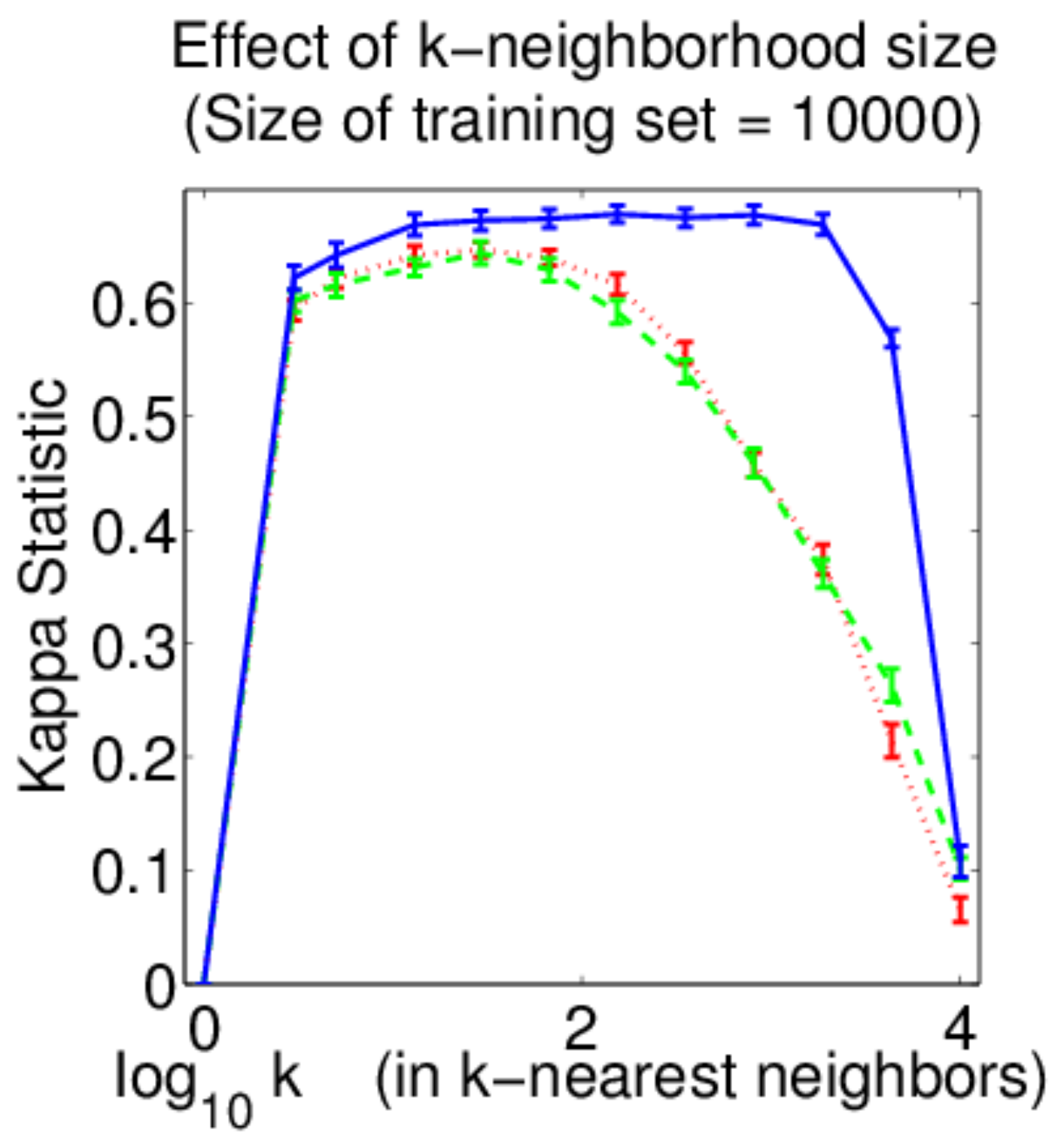} \\
\includegraphics[width=3.95cm]{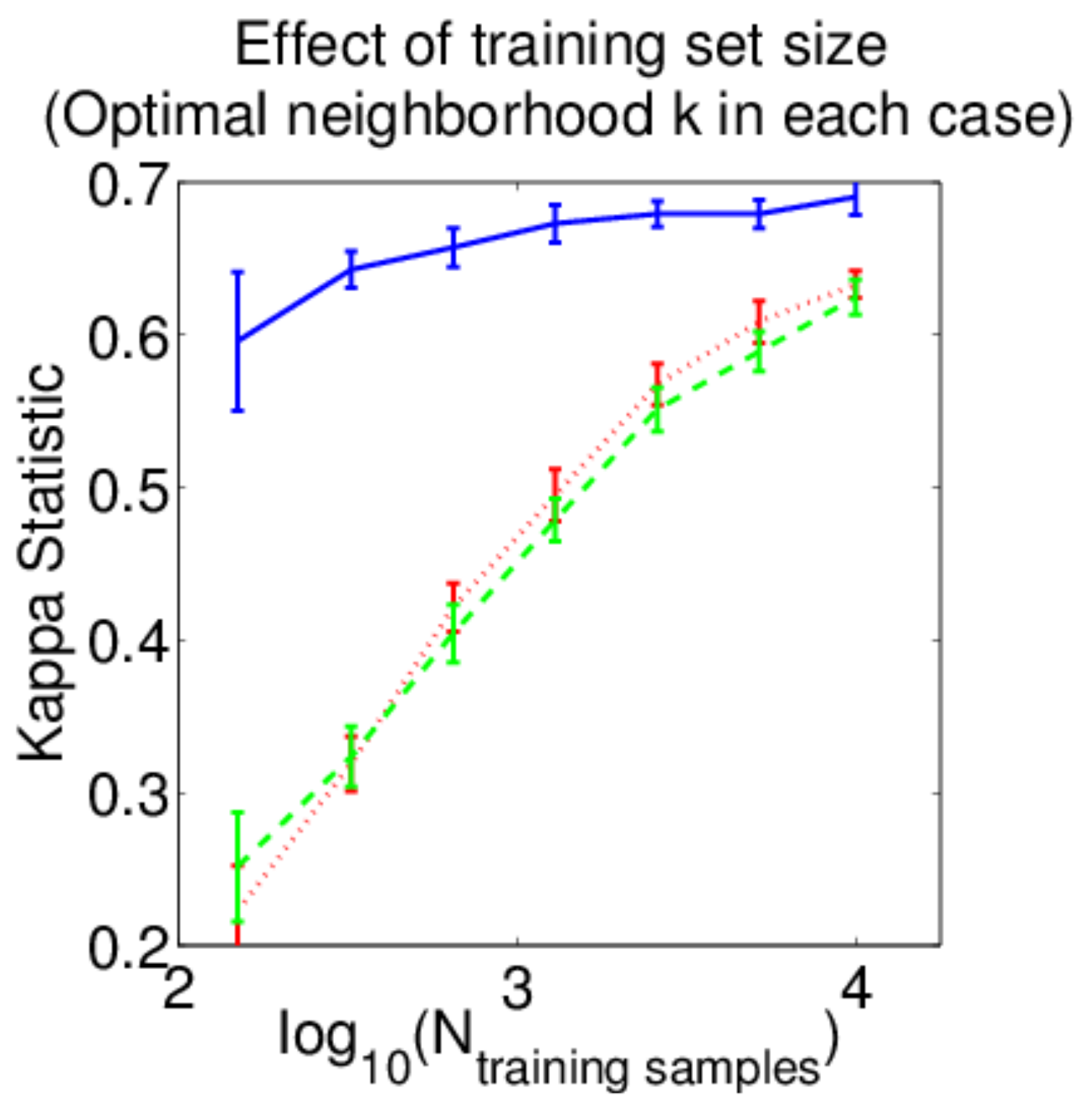} &
\includegraphics[width=3.7cm]{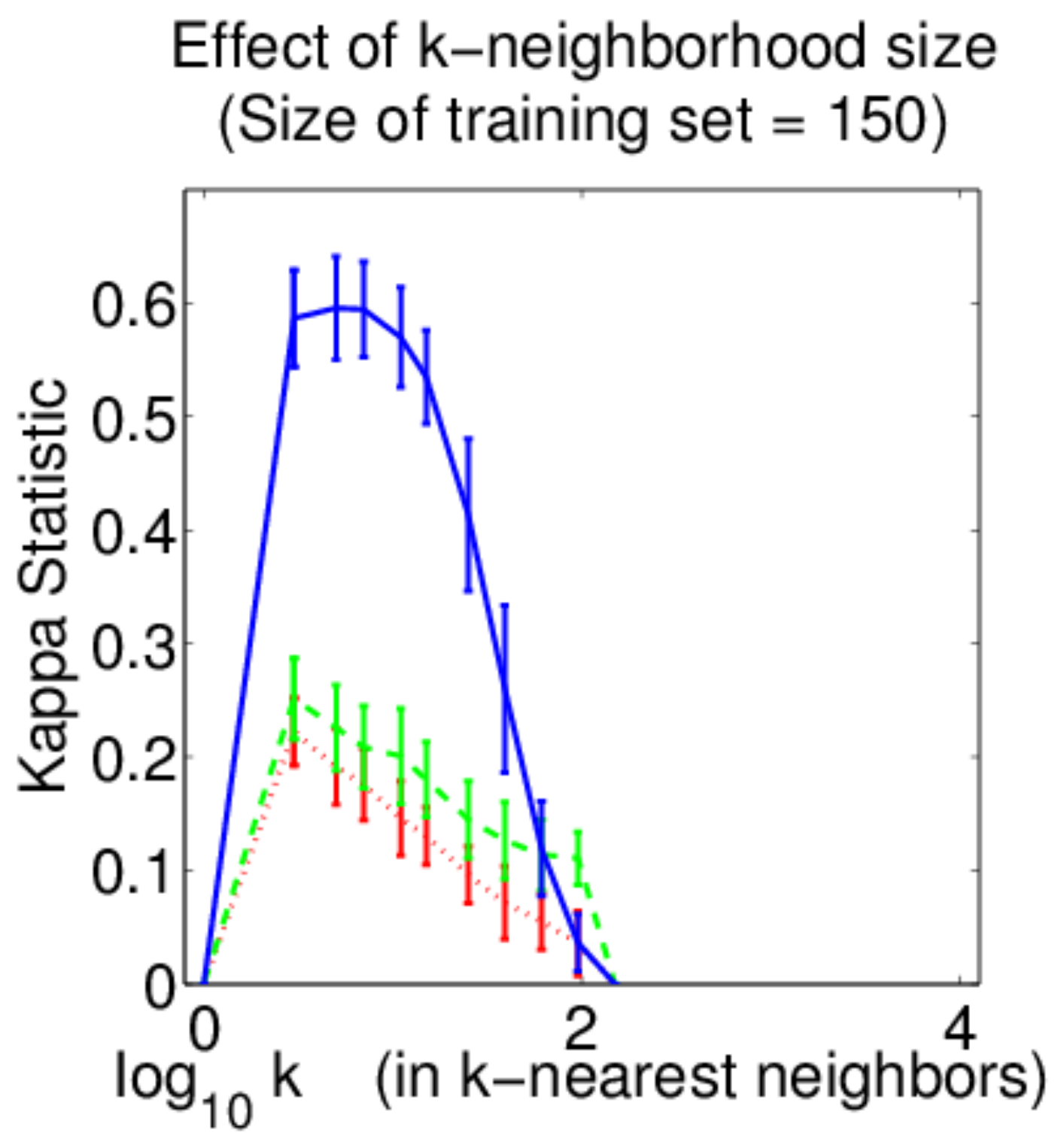} &
\includegraphics[width=3.7cm]{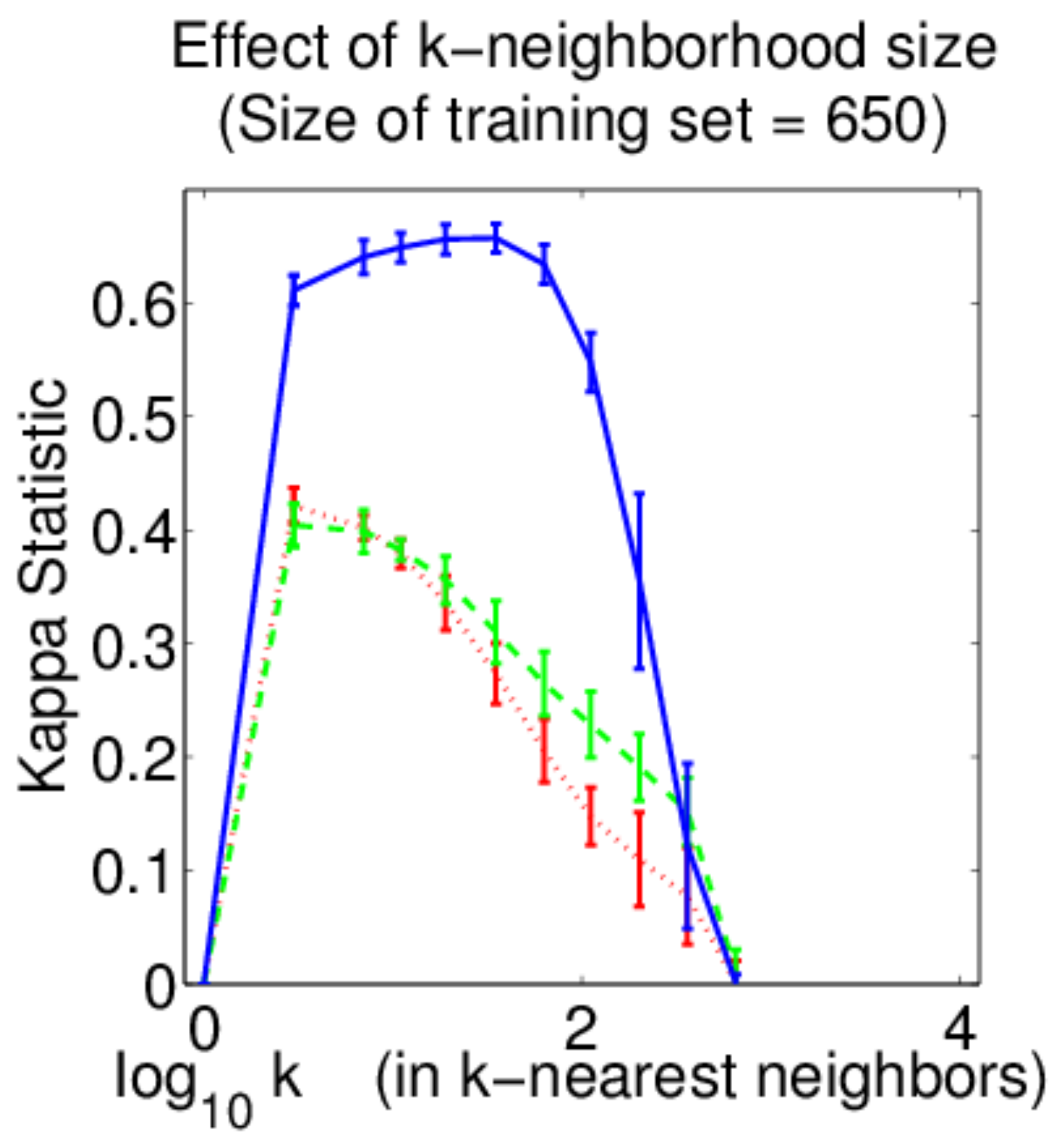} &
\includegraphics[width=3.7cm]{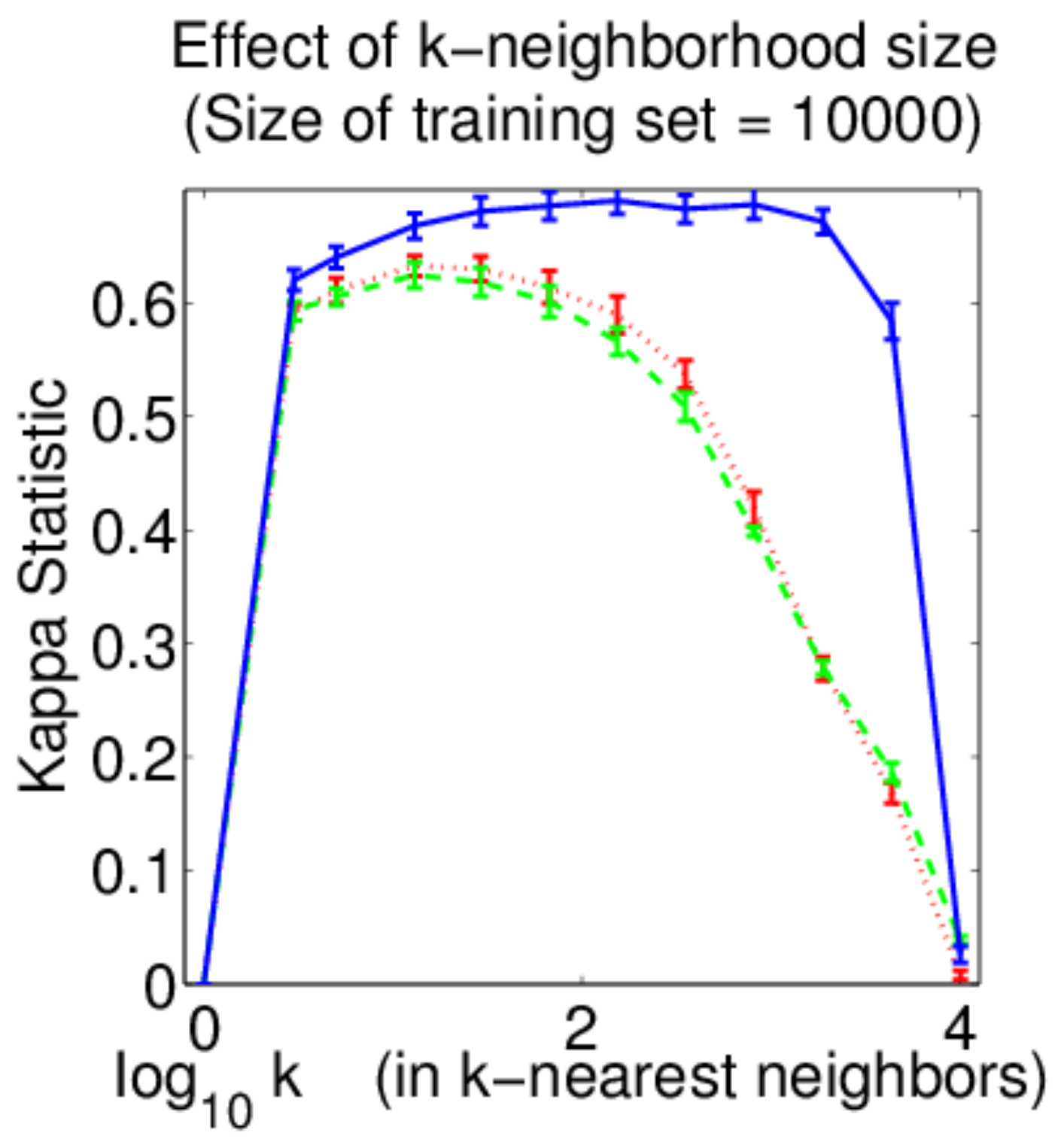} \\
\end{tabular}
\end{center}
\vspace{-0.5cm}
\caption{\small Effect of PPA metric in $k$-nearest neighbors classification for low (top) and high (bottom) curvatures. } \label{metric_helps_classification}
\end{figure*}

\paragraph{Kernel PCA.}
This non-linear generalization of PCA is based on embedding the data into a higher-dimensional Hilbert space.
Linear features in the Hilbert space
correspond to nonlinear features in the input domain~\cite{Scholkopf98}.
{Inverting the Hilbert space representation is not straightforward but a number
of pre-imaging techniques have been developed~\cite{Honeine11}.
However, there is a more important complication. While it is possible to
obtain reduced-dimensionality representations
in the Hilbert space for supervised learning~\cite{Braun08},
the KPCA formulation does not guarantee that these representations
are accurate in MSE terms in the input domain (no matter the pre-imaging technique).
This is a fundamental difference with PCA (and with PPA).
For this reason, using KPCA in experiments where reconstruction is necessary
(as those in Section~\ref{dim_red}) would not be fair to KPCA.}\\

Similarly to~\cite{Garcia11}, the main motivation of PPA is finding the input data manifold that best represents that data structure in a multivariate regression problem.
{The above discussion suggests that the proposed nonlinear extension of PCA
opens new possibilities in recent applications of linear PCA such as~\cite{Arenas13spm,Jimenez13,Meraoumia13,AlNaser12,GhoshDastidar08}, and in cases where it is necessary to take higher order
relations into account due to the nonlinear nature of the data~\cite{Martis13}.}


\section{Experiments} \label{experiments}

This section illustrates the properties of PPA through a set of four experiments. The first one illustrates the advantage of using the manifold-induced PPA metric for classification. The second one shows how to use the analytic nature of PPA to extract geometrical properties of the manifold. The third experiment analyzes the performance of PPA for dimensionality reduction on different standard databases. Finally, we show the benefits of the PPA volume-preserving property to compute the multi-information reduction. {For the interested reader, and for the sake of reproducibility, an online implementation of the proposed PPA method can be found here:
\centerline{{\url{http://isp.uv.es/ppa.html.}}}}
{The software is written in Matlab and was tested in Windows 7 and Linux 12.4 over several workstations. It contains demos for running examples of forward and inverse PPA transforms. The code is licensed under the FreeBSD license (also known as Simplified BSD license).}

\subsection{\em Benefits of the PPA metric in classification}
\label{exp_metric}

As presented above, the PPA manifold-induced metric provides more meaningful distance measures than the Euclidean distance or its linear Mahalanobis distance counterpart. To illustrate this, we consider $k$-nearest neighbors ($k$-NN) classification, whose success strongly depends on the appropriateness of the distance used~\cite{Duda07}.

\begin{figure*}[t!]
\begin{center}
\small
\setlength{\tabcolsep}{6pt}
\vspace{-0cm}
\begin{tabular}{cccccc}
\hspace{-0.1cm}\includegraphics[height=2.9cm]{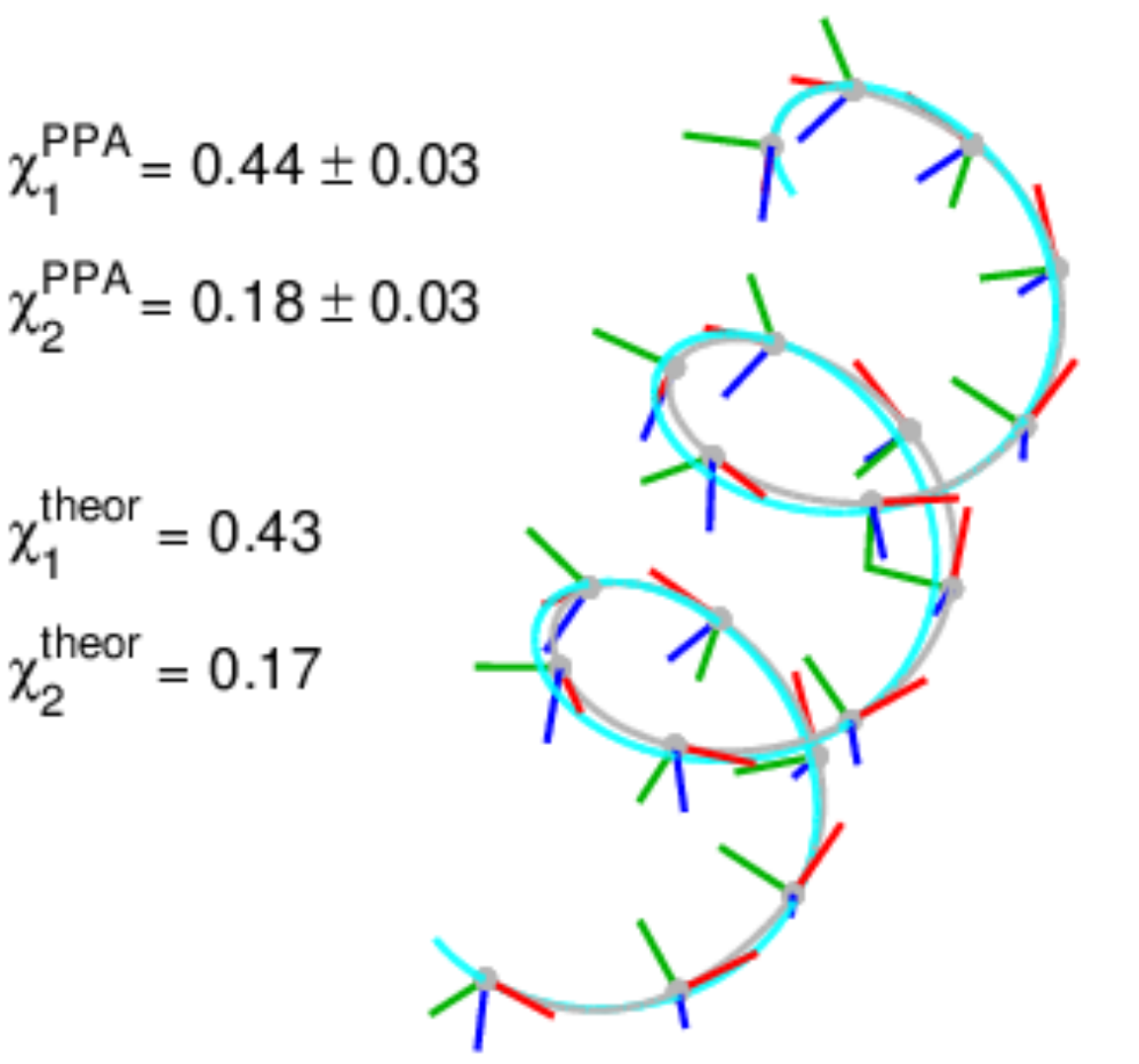} &
\hspace{-0.4cm}\includegraphics[height=2.9cm]{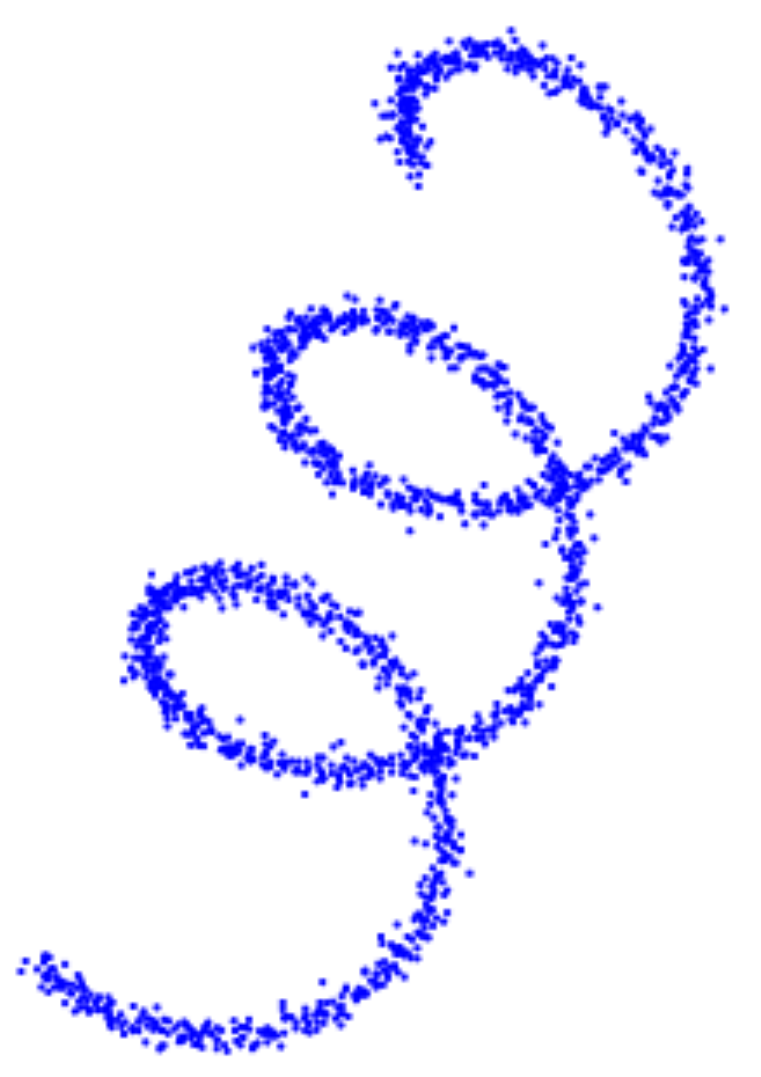} &
\hspace{0.3cm}\includegraphics[height=2.9cm]{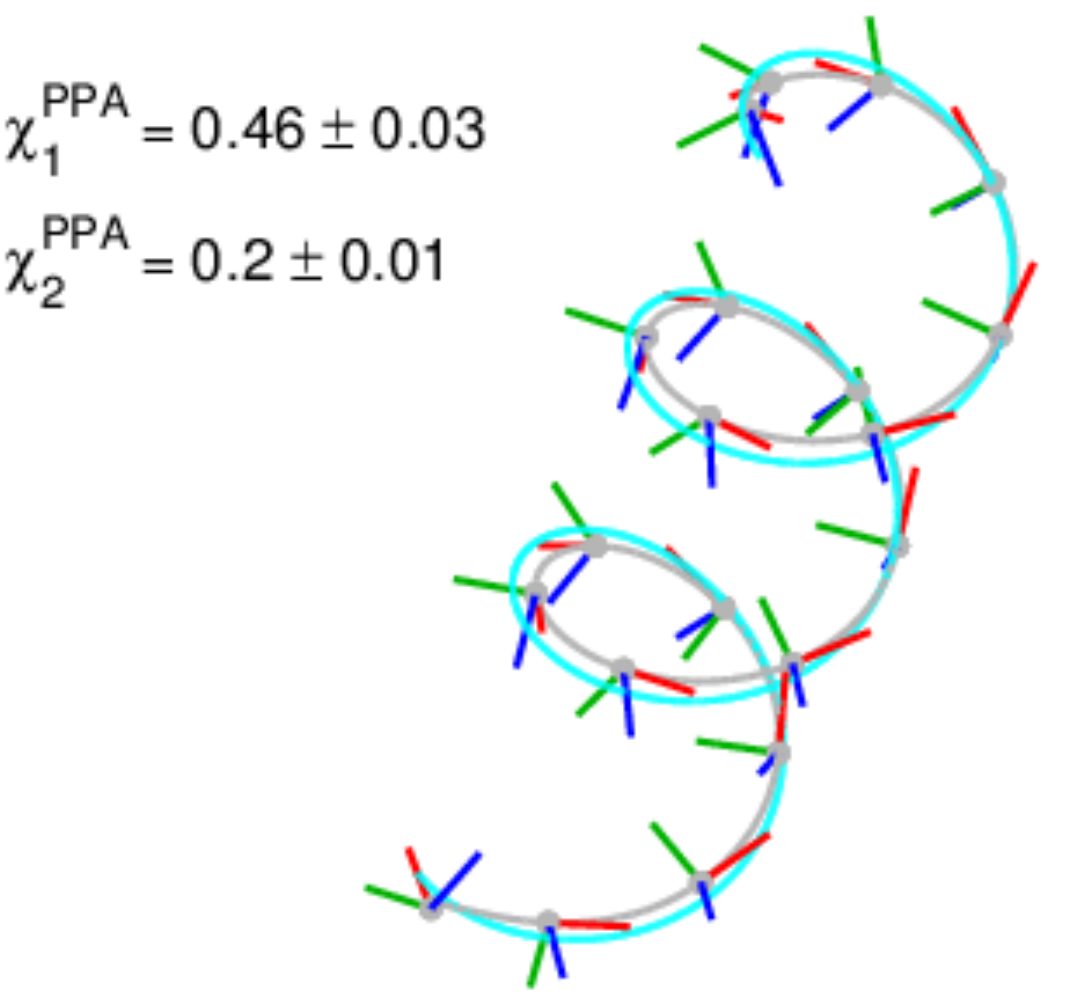} &
\hspace{-0.4cm}\includegraphics[height=2.9cm]{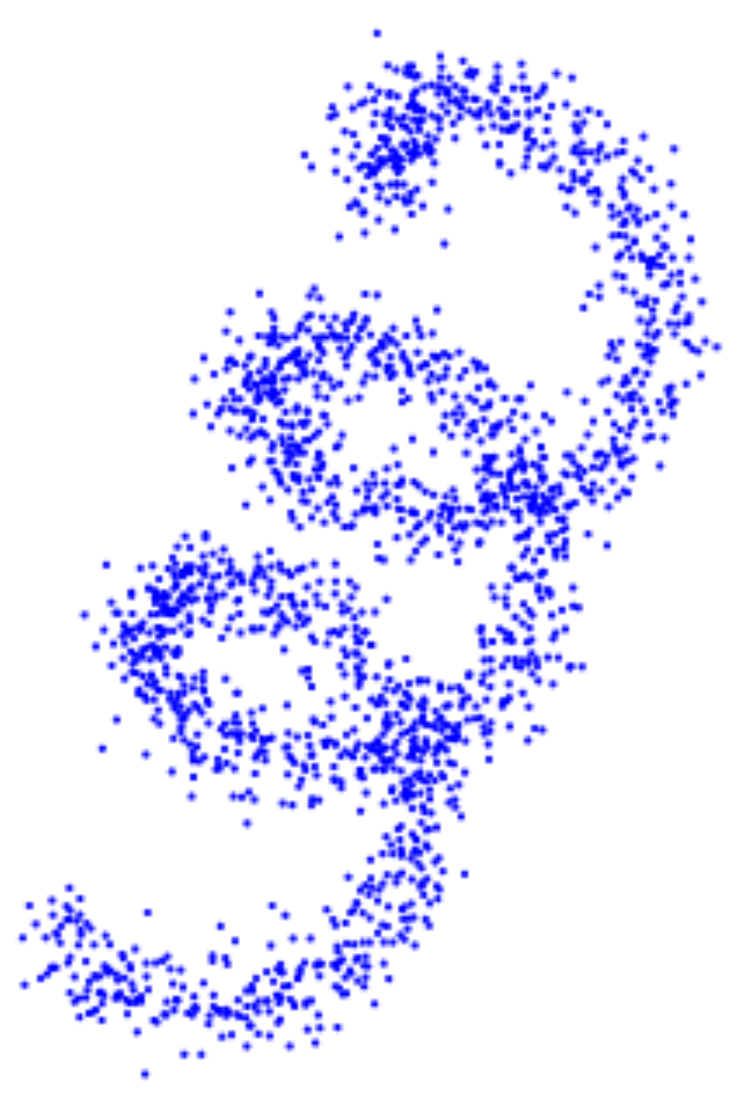} &
\hspace{0.3cm}\includegraphics[height=2.9cm]{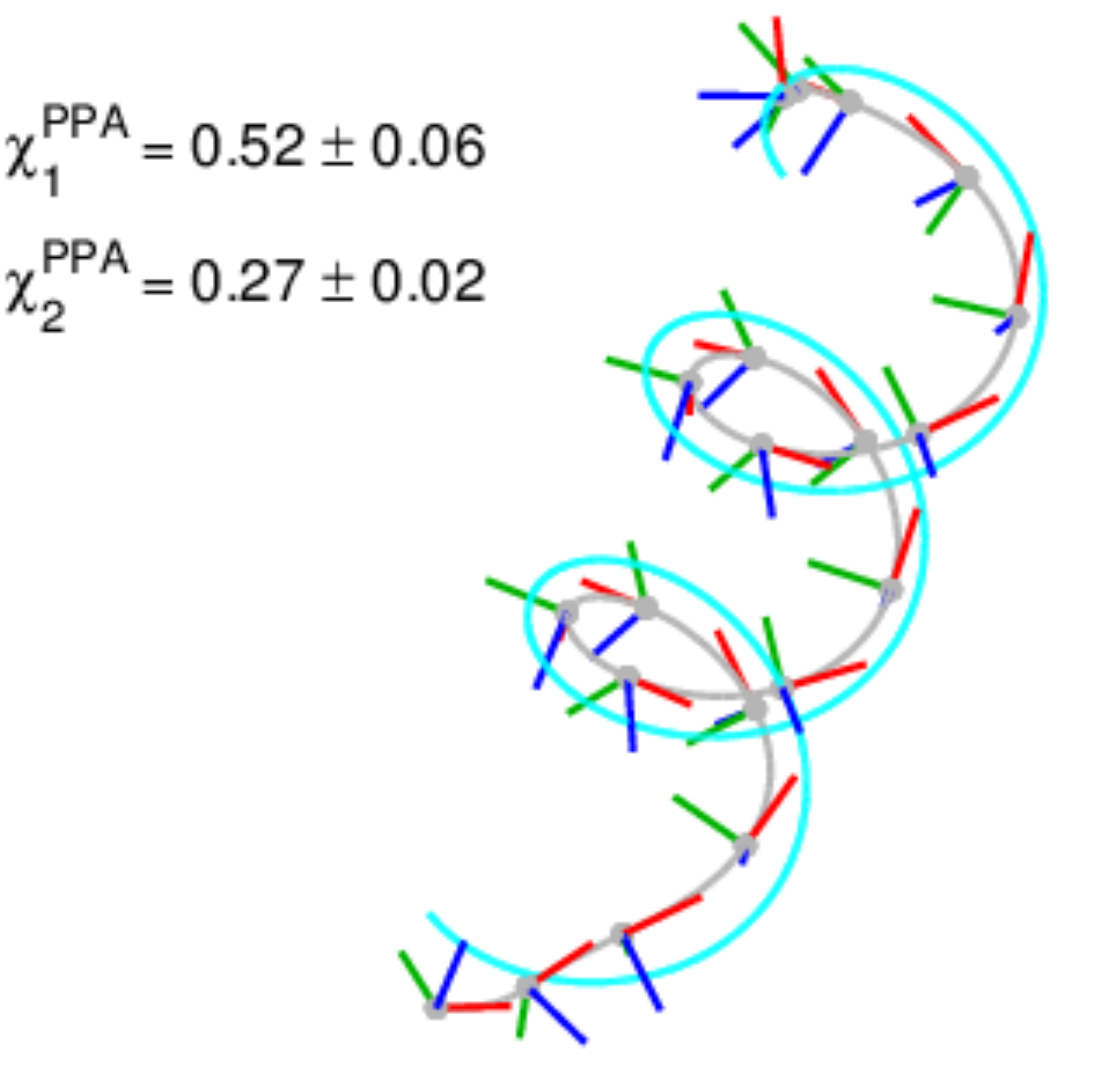} &
\hspace{-0.4cm}\includegraphics[height=2.9cm]{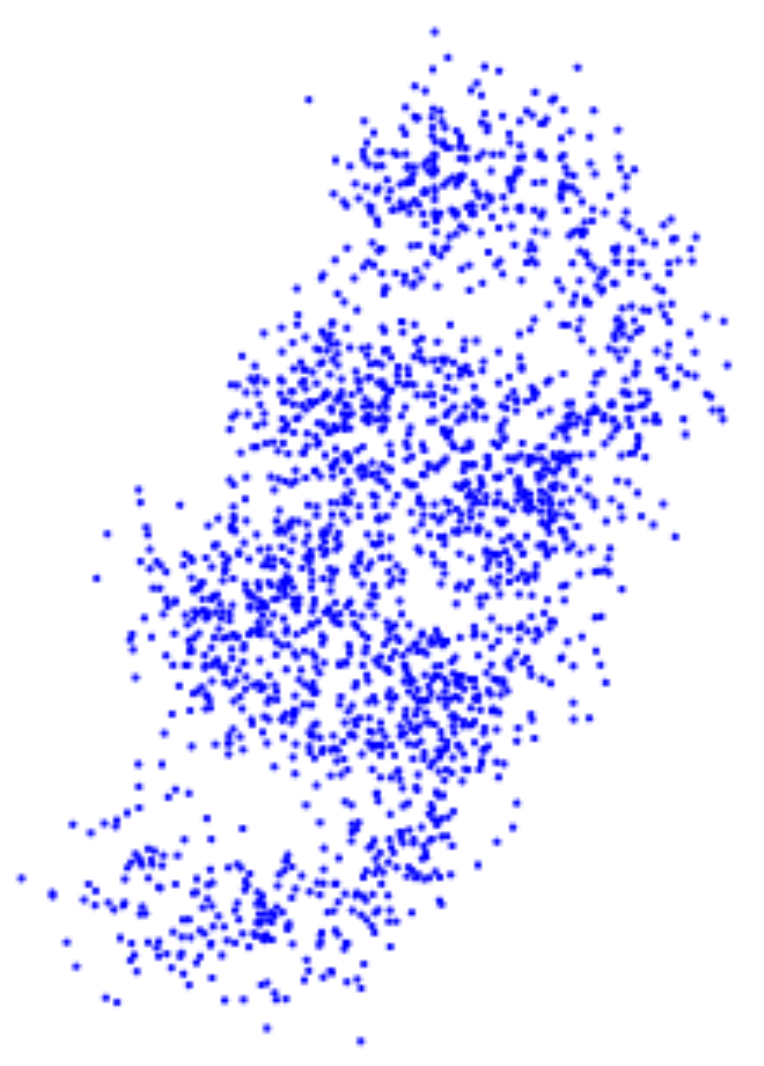} \\
\end{tabular}
\end{center}
\vspace{-0.0cm}
\caption{\small Geometric characterization of curvilinear PPA features in $3d$ helical manifolds. {Scatter plots show data used to train the PPA model (1000 training and 1000 cross-validation samples) under three different noise conditions (see text). Corresponding line plots show the actual first principal curve (in cyan) and the identified first curvilinear PPA feature (in gray). The orders of the first polynomial found by cross validation were $\gamma_1 = [12, 14, 12]$, in the respective noise conditions}. Lines in RGB stand for the \emph{tangent}, \emph{normal} and \emph{binormal} vectors of the Serret-Frenet frame at each point of the PPA polynomial.}
\label{helix}
\end{figure*}

We focus on the synthetic data in Fig.~\ref{ellipsoids}, where two classes are presented. They have both been generated from noisy parabolas.
{A cross-validation procedure on 1000 samples fitted the degree of the polynomials describing the data to $\gamma_p = 2$}.
Figure~\ref{metric_helps_classification} shows the positive effect of considering PPA metric when $\Lambda_{\mathrm{PPA}}$ is a diagonal matrix with the variances of the response coefficients (i.e. generalization of the Mahalanobis distance) for $k$-NN classification~\cite{Duda07}.
Better performance is obtained when considering the PPA metric compared to the Euclidean or the linear Mahalanobis counterparts, especially for few training samples (Fig.~\ref{metric_helps_classification}). Moreover, the accuracy of the classifier built with the PPA metric is fairly insensitive to the number of neighbors $k$ in $k$-NN, no matter the number of samples. The gain observed with the PPA metric increases with the curvature of the data distribution (bottom row of Fig.~\ref{metric_helps_classification}). Note that, with higher curvatures, the Euclidean and the linear Mahalanobis metric perform similarly poor. When a larger number of samples is available the results become roughly independent of the curvature, but even in that situation the PPA metric outperforms the others.

The generalization of the Mahalanobis metric using PPA may also be useful in extending hierarchical SOM models using more general distortion measures~\cite{Ezequiel14}, which are useful
for segmentation~\cite{Ezequiel11}.

\subsection{\em Differential geometry of PPA curvilinear features}\label{helice}

According to standard differential geometry of curves in $d$-dimensional spaces~\cite{Dubrovin82},
characteristic properties of a curve such as generalized curvatures $\chi_p$, with $p~=~[1,\ldots, d-1]$,
and Frenet-Serret frames, are related to the $p$-th derivatives of the vector tangent to the curve.
At a certain point $\x$, the vector tangent to the $p$-th curvilinear dimension corresponds
to the $p$-th column of the inverse of the Jacobian.

We now use the analytical nature of PPA to obtain a complete geometric characterization
of the curvilinear features identified by the algorithm. In each step of the PPA sequence,
the algorithm obtains a curve (polynomial) in $\mathbb{R}^d$.
Below we compute such characterization for data coming from helical
manifolds where the comparison with ideal results is
straightforward\footnote{In $3d$ spaces, the two generalized curvatures
that fully characterize a curve are known simply as \emph{curvature} and \emph{torsion}.
In the case of an helix with radius, $a$, and pitch, $2\pi b$, the
curvature and torsion are given by $\chi_1=|a|/(a^2+b^2)$ and $\chi_2=b/(a^2+b^2) $~\cite{Dubrovin82}.}.
Note that this is not just an illustrative exercise, because this manifold arises
in real communication problems, and due to its interesting structure, it served as test case
for Principal Curves Methods \cite{Ozertem11}.

The first example considers a $3d$ helix where the Frenet-Serret frames are
easy to visualize as orthonormal vectors.
 Figure~\ref{helix} shows the first curvilinear feature identified by PPA (in gray) compared to the actual helix used to generate the $3d$ data (in cyan), for different noise levels. We used $a=2$, $b=0.8$, and Gaussian noise of standard deviations $0.1$, $0.3$, and $0.6$, respectively. Note that in the high noise situation, the noise scale is comparable to the scale of the helix parameters.

The tangent vectors of this first curvilinear feature (in red) are computed from the first column of the inverse of the Jacobian (using Eqs.~\eqref {det_sequence} and~\eqref{individual_jacobian}). The other components of the Frenet-Serret frames (in $3d$, the \emph{normal} and \emph{binormal} vectors, here in green and blue), are computed from the derivatives of the tangent vector, and the generalized curvatures are given by the Frenet-Serret formulas~\cite{Dubrovin82}. For each of the three examples, we report the curvature values obtained by the PPA curves, as well as the theoretical values for the generating helix. Even though curvature and torsion are constant in an helix, $\chi_1^{\mathrm{PPA}}$ and $\chi_2^{\mathrm {PPA}}$ are slightly point-dependent. That is the reason for the standard deviation in the $\chi_i^{\mathrm{PPA}}$ values. In this particular illustration, the effect of noise leads to a more curly helix, hence overestimating the curvatures.

\begin{figure*}[t!]
\begin{center}
\small
\setlength{\tabcolsep}{6pt}
\vspace{-0cm}
\begin{tabular}{cc}
\includegraphics[height=5.2cm]{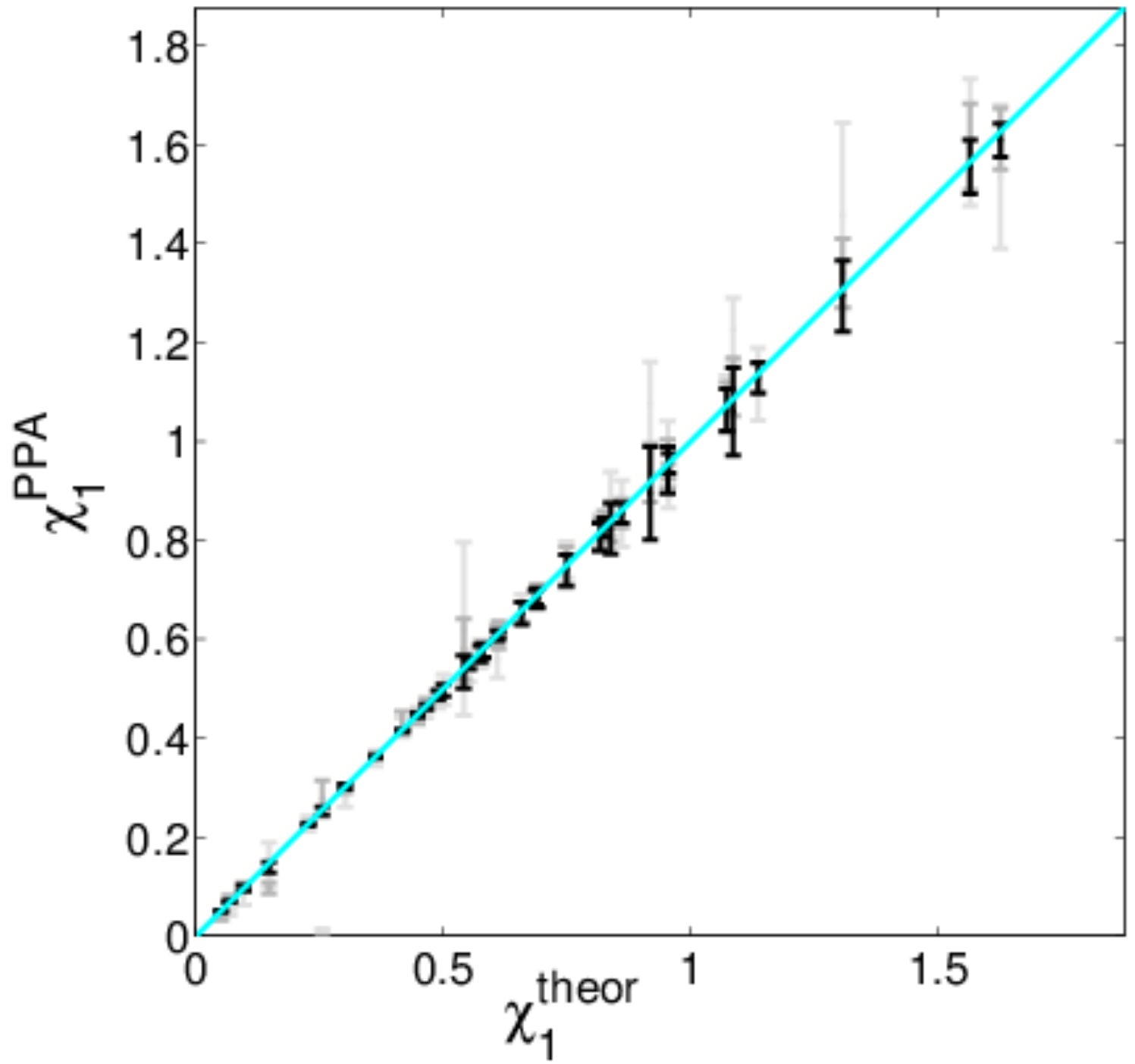} \hspace{0.5cm}& \hspace{0.5cm}
\includegraphics[height=5.2cm]{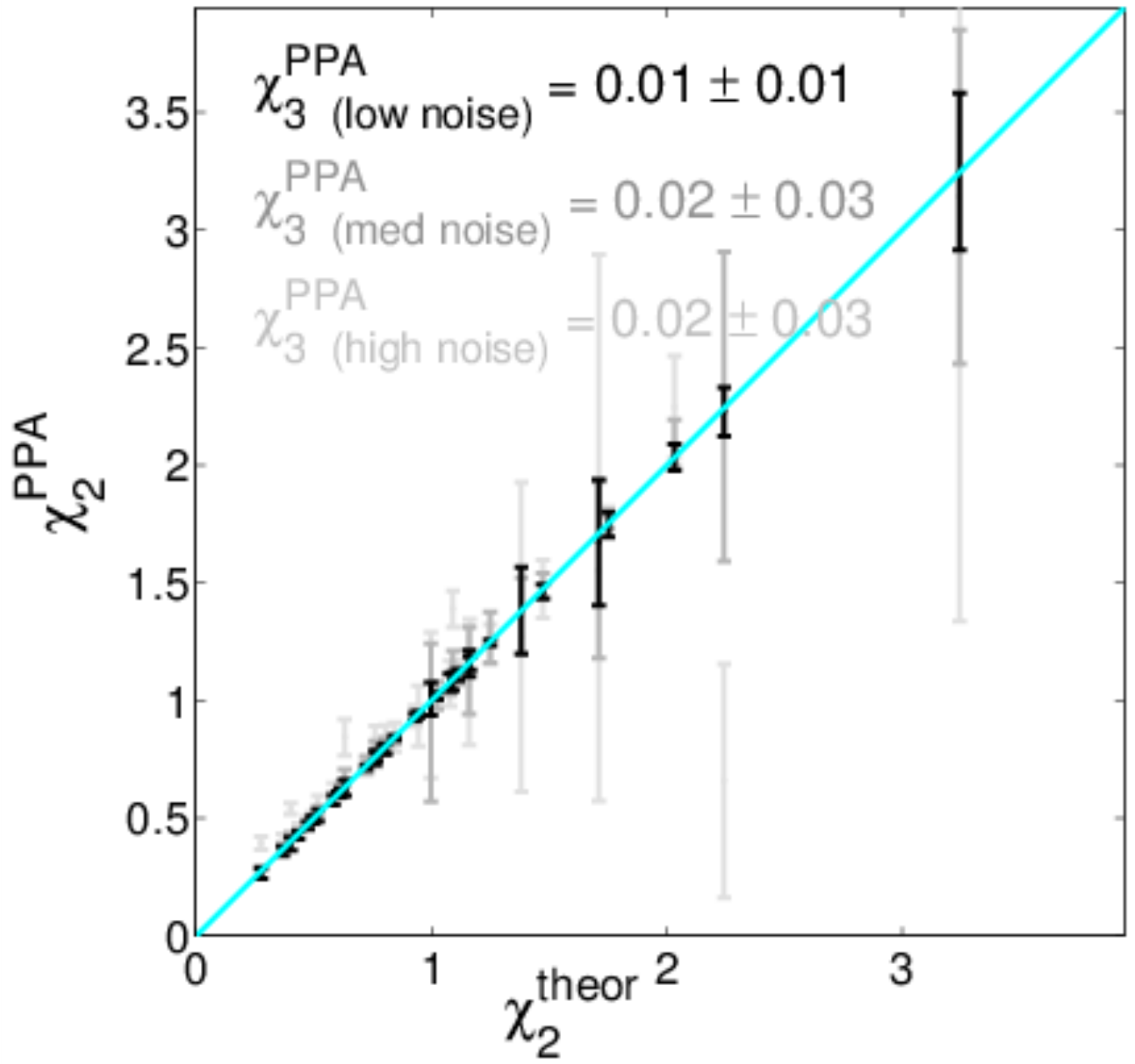}\\
\end{tabular}
\end{center}
\vspace{-0cm}
\caption{\small Geometric characterization of $4d$ helical manifolds using PPA. 
Prediction of generalized curvatures $\chi_1$ (left), and $\chi_2$, $\chi_3$ (right) for a wide family of $4d$ helical datasets (see text for details).
Darker gray stands for lower noise levels.
According to the way data were generated, the theoretical value of the third generalized curvature is $\chi_3^{\mathrm{theor}}=0$.}
\label{helix4d}
\end{figure*}

In the second example, we consider a higher dimensional setting and embed $3d$ helices with arbitrary radius and pitch (in the [0,1] range) into the $4d$ space by first adding zeros in the $4th$ dimension, and then applying a random rotation in $4d$. Since the rotation does not change the curvatures, $\chi_1^{\mathrm{theor}}$ and $\chi_2^{\mathrm{theor}}$ can be computed as in the $3d$ case, and $\chi_3^{\mathrm{theor}}=0$. Fig.~\ref{helix4d} shows the alignment between $\chi_i^{\mathrm{theor}}$ and $\chi_i^{\mathrm{PPA}}$ for different noise levels. We also report the $\chi_3^{\mathrm{PPA}}$ values (that should be zero).
Noise implies different curvature estimations along the manifold (larger variance), and, for particular combinations of $a$ and $b$,
noise also implies bias in the estimations: divergence form the (ideal agreement).

{Also remind that the} PPA formulation allows to obtain Frenet-Serret frames in more than three dimensions. However, visualization in those cases is not straightforward.
For illustration purposes here we focus on the \emph{first} PPA curvilinear dimension. Nevertheless, the same geometric descriptors ($\chi_i$ and Frenet-Serret frames) can be obtained along the
curvilinear features. Estimation of curvatures from the PPA model may be interesting in applications where geometry determines resource allocation~\cite{Ronan11}.

\subsection{\em Dimensionality reduction}
\label{dim_red}
{In this section, we first illustrate the ability of PPA to visualize high dimensional data in a similar way to Principal Volumes and Surfaces. Then, we compared the
performance of PCA, PPA and nonlinear PCA (NLPCA) of \cite{Scholz07} in terms of reconstruction error {obtained} after
truncating a number of features.}

\vspace{-0.7cm}
{\paragraph{Data.} We use six databases extracted from the UCI repository\footnote{The databases are available at \url{http://archive.ics.uci.edu/ml/datasets.html}}.
The selected databases deal with challenging real problems and were chosen according to these criteria:
they are defined in the real domain, they {are} high-dimensional ($d \geq 9$),
the ratio between the number of samples and the number of dimensions is large ($n/d \geq 40$),
and they display nonlinear relations between components (which was evaluated by pre-visualizing the data). See data summary below and in table I:
\begin{itemize}
\item \emph{MagicGamma.}
The data{set} represent traces of high energy gamma particles in a ground-based atmospheric Cherenkov gamma telescope.
The available information consists of pulses left by the incoming Cherenkov photons on the photomultiplier tubes, arranged in an image plane.
The input features are descriptors of the clustered image of gamma rays in an hadronic shower background.
\item \emph{Japanese Vowels.}
This dataset deals with vowel identification in japanese, and contains cepstrum coefficients estimated from speech.
Nine speakers uttered two Japanese vowels {\sf /ae/} successively.
Linear analysis was applied to obtain a discrete-time series with 12 {linear prediction cepstrum coefficients},
which constitute the input features.
\item {\em Pageblocks.}
The database describes the blocks of the page layout of documents that have been detected by a segmentation process.
The feature vectors come from 54 distinct documents and characterize each block with 10 numerical attributes such as height, width, area, eccentricity, etc.
\item {\em Sat.}
This dataset considers a Landsat MSS image consisting of 82$\times$100 pixels with a spatial resolution of 80m$\times$80m,
and 4 wavelength bands.
Contextual information was included by stacking neighboring pixels in 3$\times$3 windows.
Therefore, 36-dimensional input samples were generated, with a high degree of redundancy.
\item {\em Segmentation.}
This dataset contains a collection of images described by 16 high-level numeric-valued attributes, such as average intensity,
rows and columns of the center pixel, local density descriptors, etc.
The images were hand-segmented to create a classification label for every pixel.
\item {\em Vehicles.}
The database describes vehicles through the application of an ensemble of 18 shape feature extractors to the 2D silhouettes of the vehicles.
The original silhouettes come from views from many different distances and angles.
This is a suitable dataset to assess manifold learning algorithms that can adapt to specific data invariances of interest.
\end{itemize}}
{For} every dataset we normalized the values in each dimension between zero and one.
We {use a maximum of} $20$ dimensions which is the limit in the available
implementation of NLPCA (http://www.nlpca.org/) \cite{Scholz12}.
Note that our implementation of PPA does not have this problem.

\begin{table}[h!]
\begin{center}
\caption{Summary of the data-sets.}
 \begin{tabular}{|c|l|c|c|c|c|}
  \hline
                   & Database & $n$ ($\sharp$ samples) & $d$ (dimension) & $n/d$ \\
  \hline
  \hline
 	1 & MagicGamma &  19020 & 10 & 1902\\
  \hline
        2 & Japanese Vowels &  9961 & 12 & 830\\
  \hline
 	3 & Pageblocks &  5473 & 10 & 547\\
  \hline
 	4 & Sat &  6435 & 36 & 179\\
  \hline
 	5 & Segmentation &  2310 & 16 & 144\\
  \hline
 	6 & Vehicles &  846 & 18 & 47\\
 \hline
  \end{tabular}
\label{UCI}
\end{center}
\end{table}
\vspace{0.0 cm}

\begin{figure*}[t!]
\begin{center}
\begin{tabular}{ccccc}
\hspace{-0.2cm} [2,5] &
\hspace{-0.2cm} [3,9] &
\hspace{-0.2cm} [4,3] &
\hspace{-0.2cm} [5,7] &
\hspace{-0.2cm} [9,1] \\
\hspace{-0.2cm} \includegraphics[width=2.7cm]{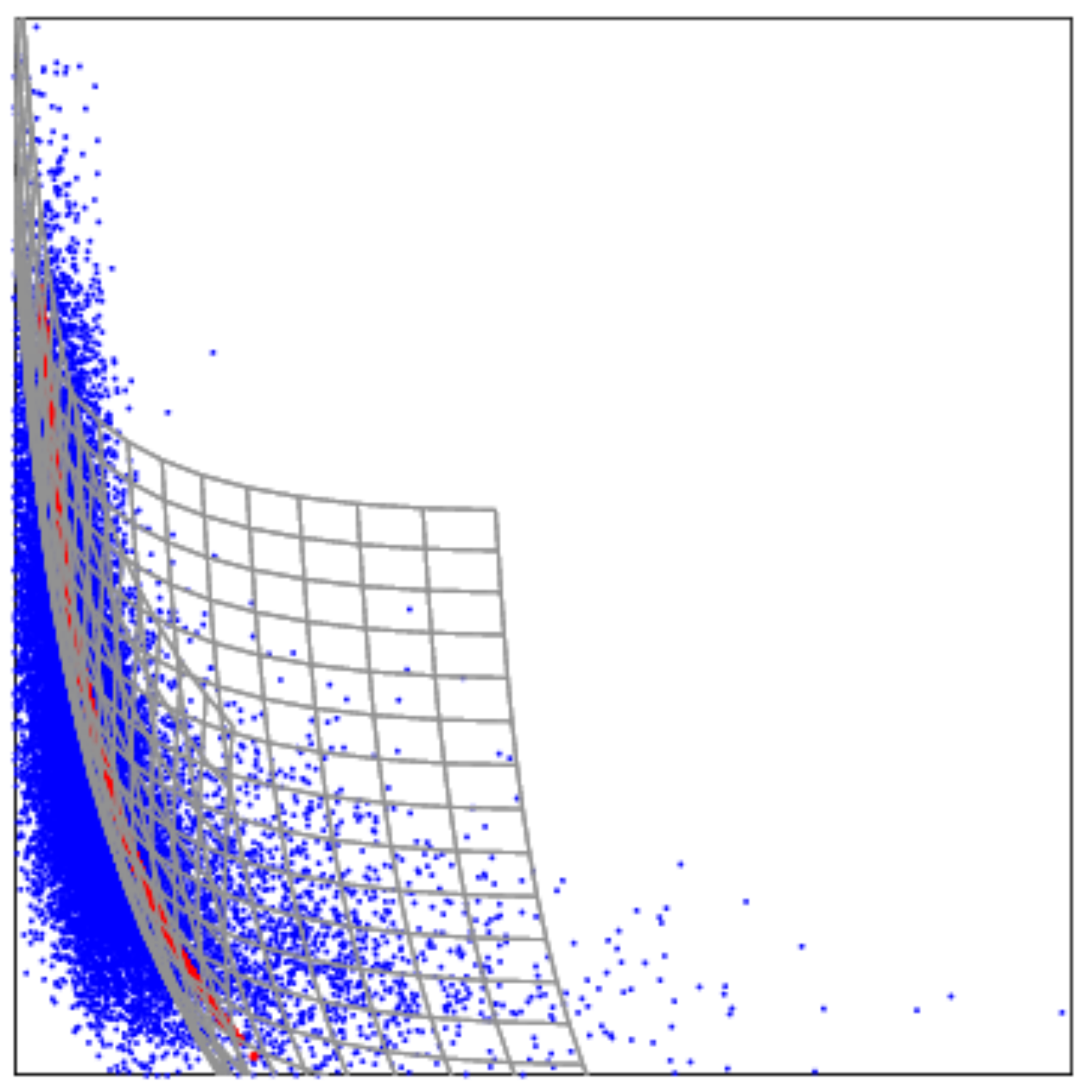} &
\hspace{-0.2cm} \includegraphics[width=2.7cm]{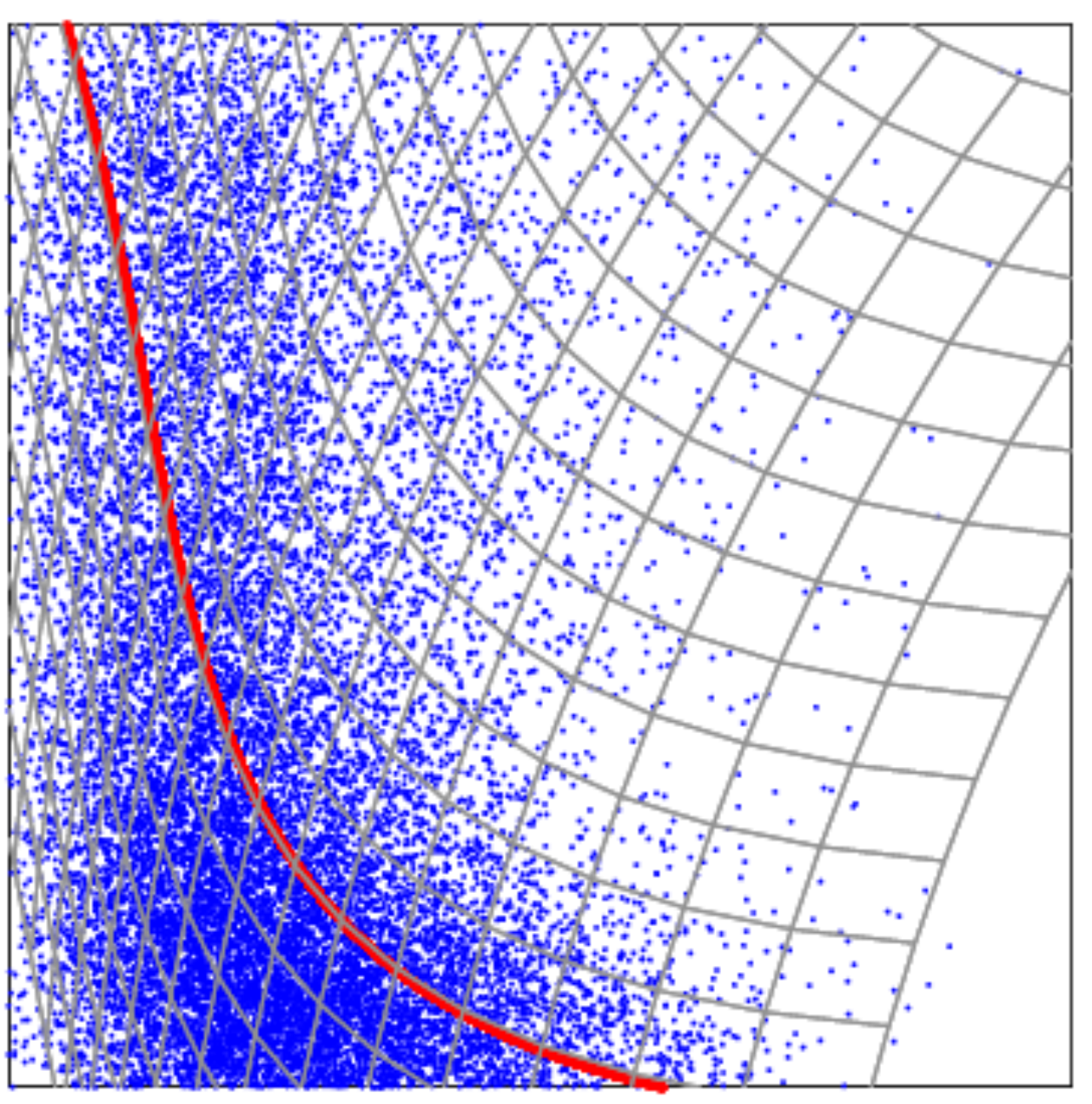} &
\hspace{-0.2cm} \includegraphics[width=2.7cm]{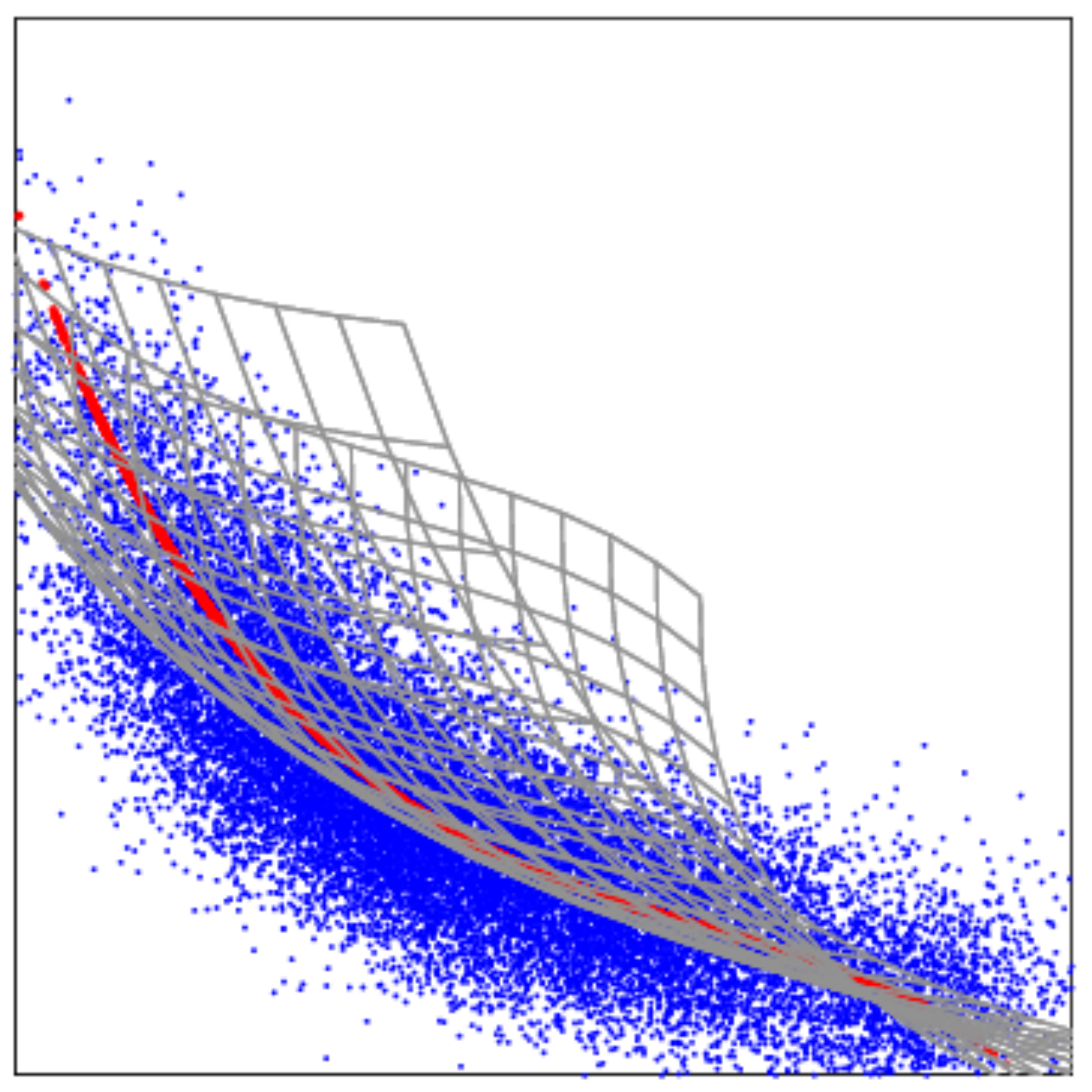} &
\hspace{-0.2cm} \includegraphics[width=2.7cm]{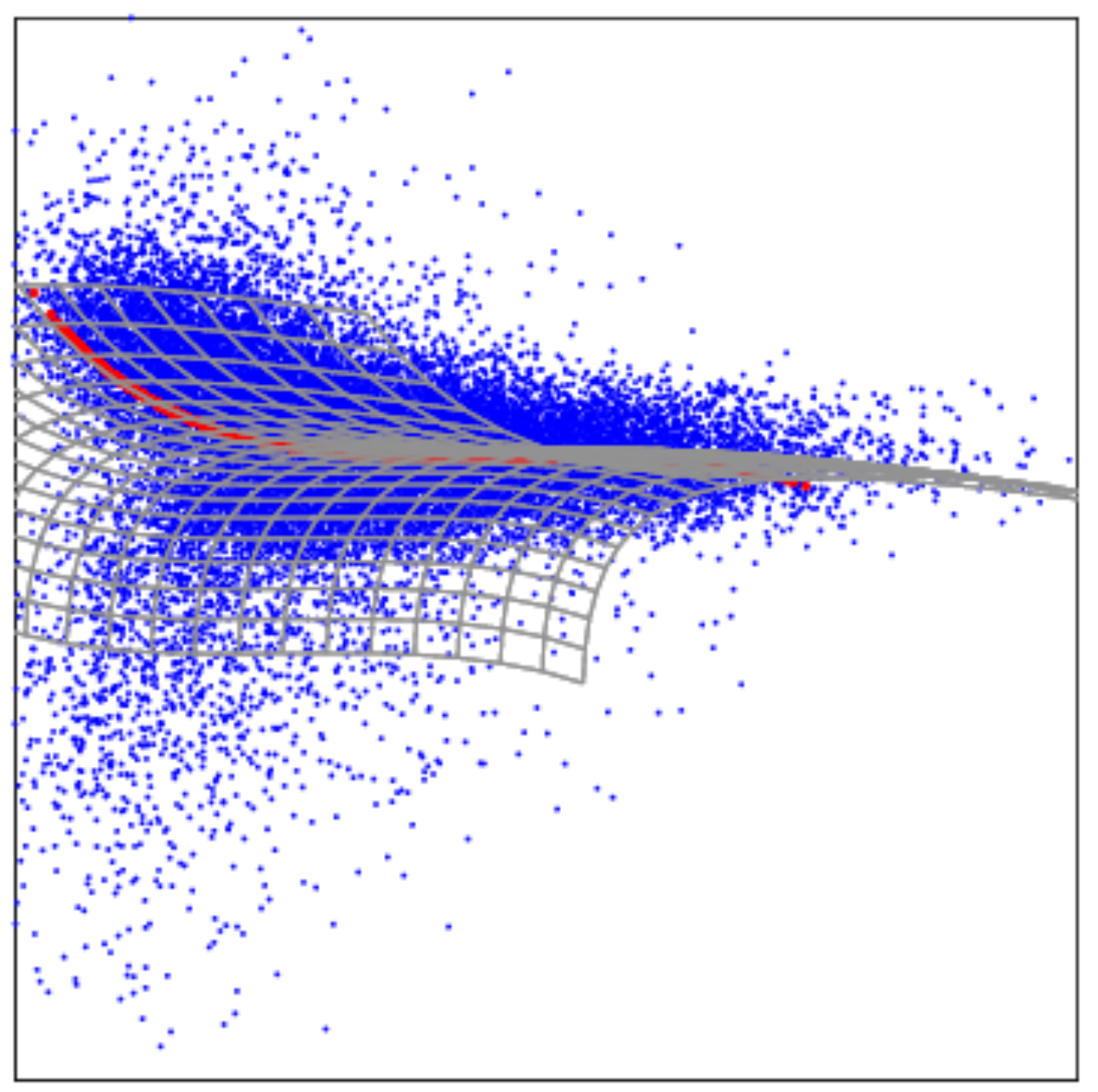} &
\hspace{-0.2cm} \includegraphics[width=2.7cm]{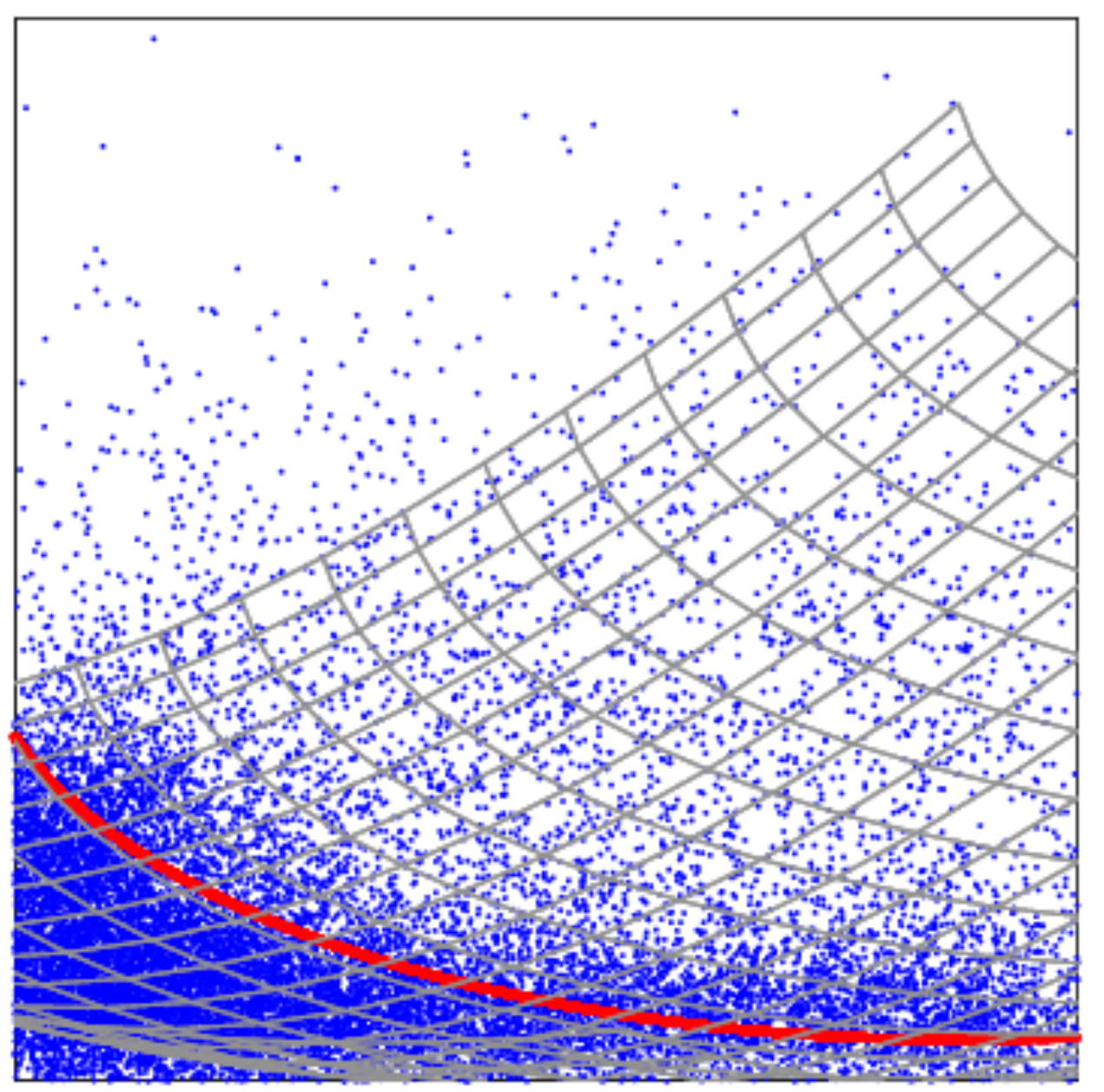}
\end{tabular}
\begin{tabular}{ccc}
\hspace{-0cm} \includegraphics[width=4.3cm]{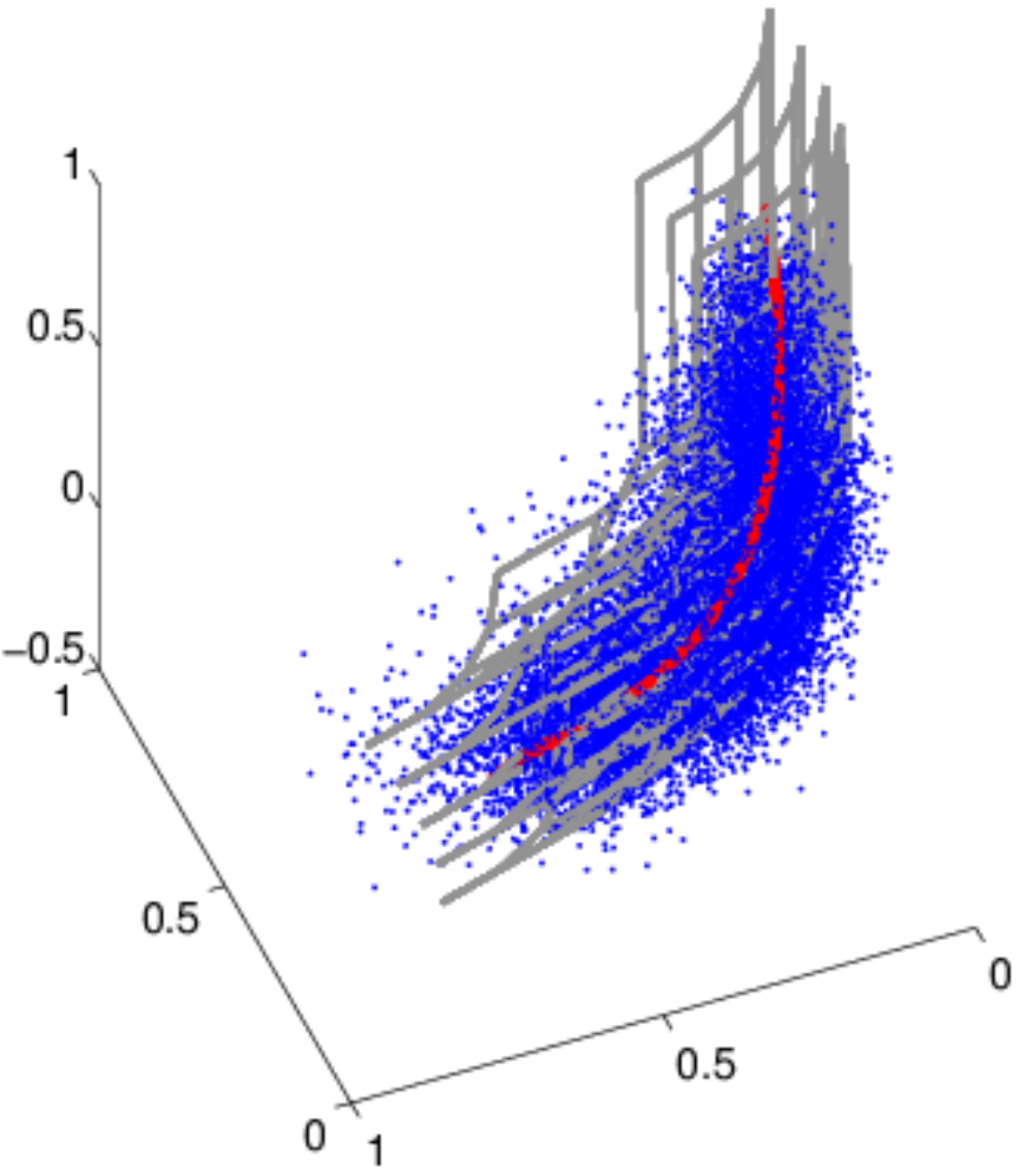}
\hspace{-0cm} \includegraphics[width=4cm]{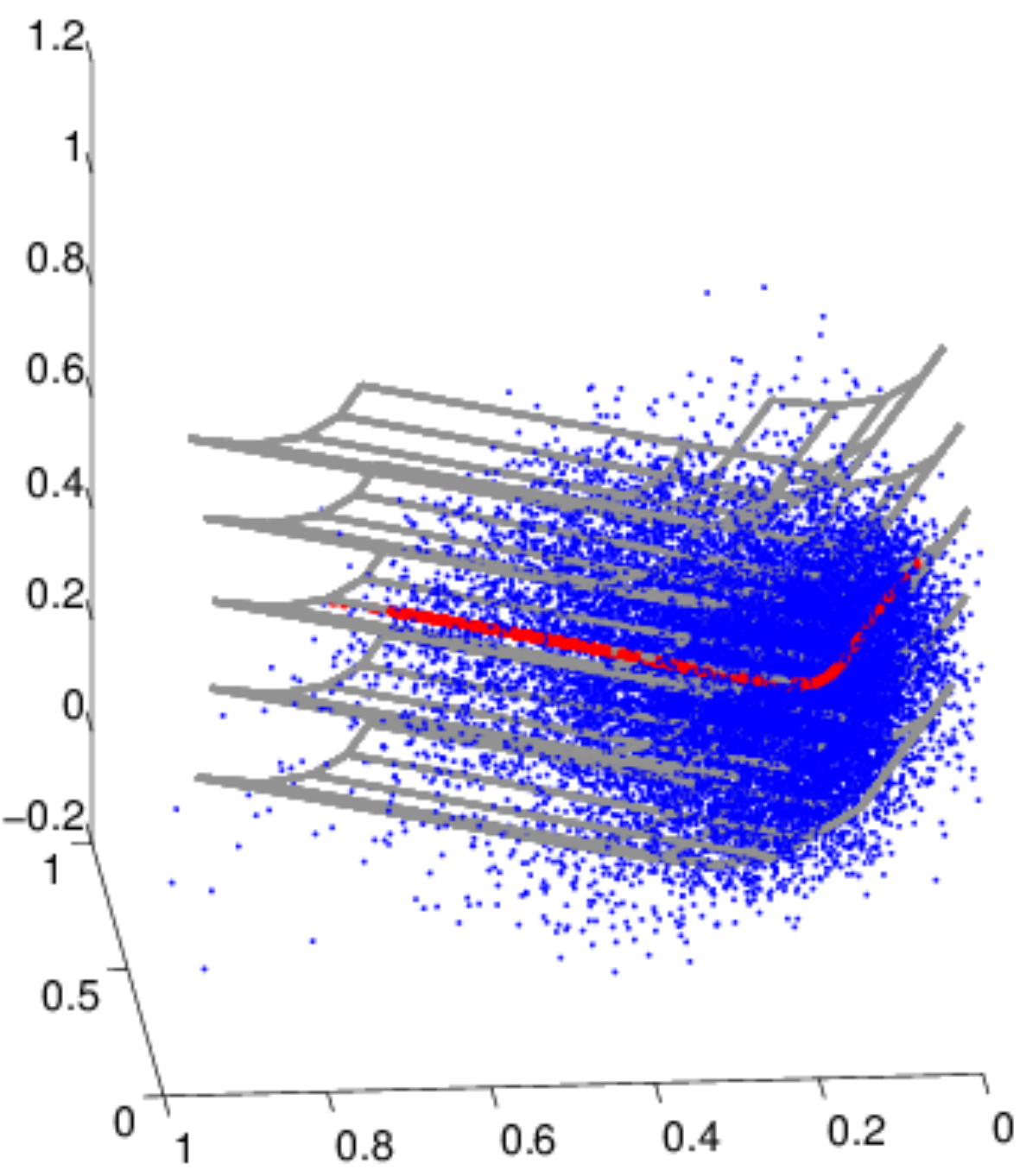}
\hspace{-0cm} \includegraphics[width=4.7cm]{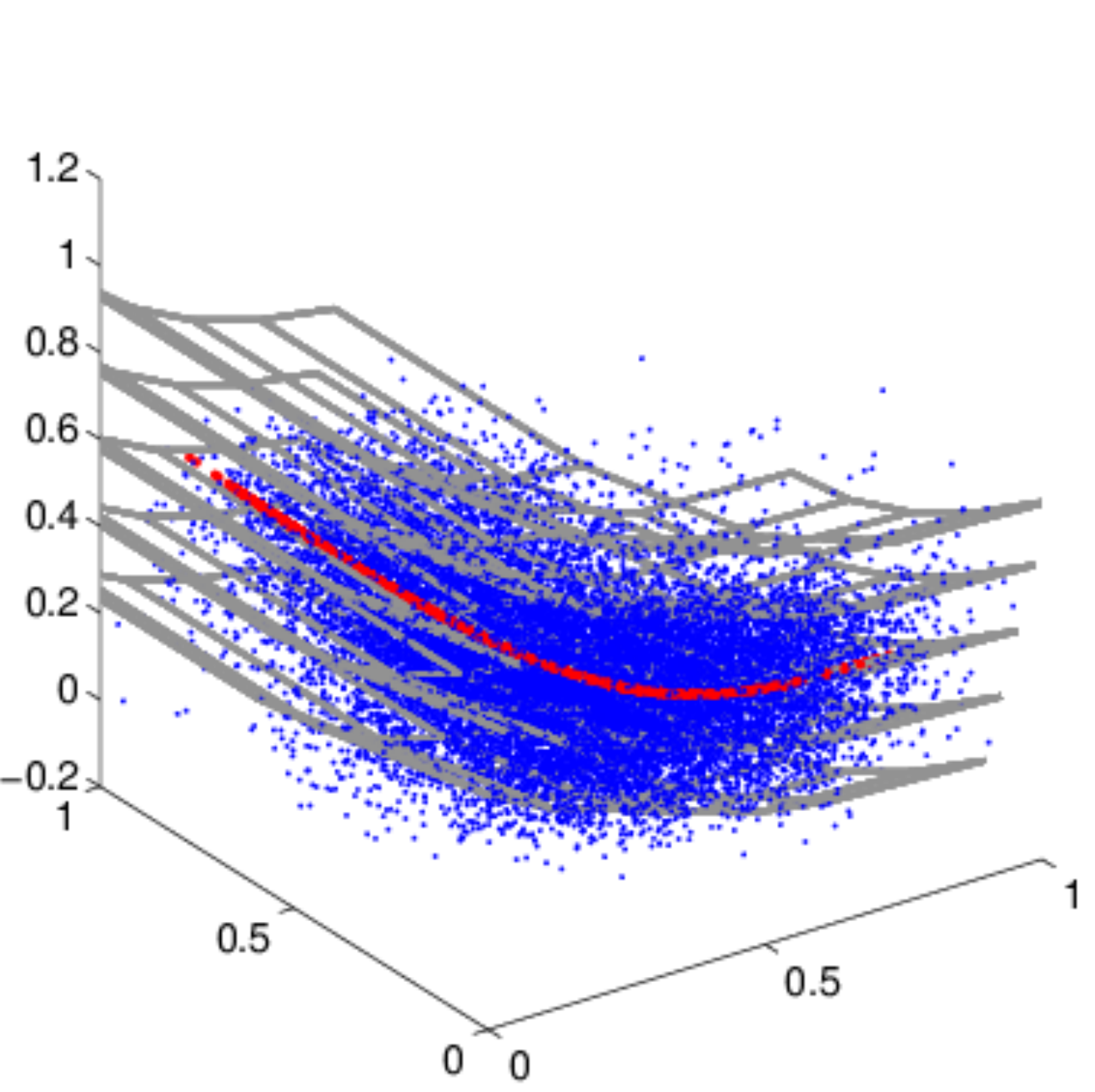}
\end{tabular}
\caption{Principal Curves, Surfaces and Volumes using PPA. {First row shows a 2$d$ visualization. Titles of the panels indicate the dimensions being visualized. In each panel, are the original data (blue dots), the curve (red) is the reconstructed data (when using only one dimension) and the gray lines correspond to a grid representing the two first PPA dimensions. The second row shows 3$d$ visualizations of dimensions $[3,5,10]$ from different camera positions. } In this case, the inverted uniform grid has been constructed in the three first dimensions of the transformed domain. See text for details.}
\label{DR_results_ex_1_2D}
\end{center}
\end{figure*}

{
\paragraph{PPA learning strategies.}
In the experiments, the alternative strategies described in Sections~\ref{the_extension} and~\ref{leading_vector} will be referred to as: (1) {\bf PPA}, which is the \emph{PCA-based solution} that inherits the leading vectors $\e_p$ from PCA; and (2) {\bf PPA GD}, which is the \emph{gradient-descent solution} that obtains $\e_p$ via minimization of Eq.~\eqref{PPA_cost}.
In both cases, the transforms are obtained using 50\% of the data, and the polynomial degree is selected automatically
(in the range $\gamma\in[1,5]$) by cross-validation using 50\% of the training data.
}

\paragraph{PPA Principal Curves, Surfaces and Volumes.}
First we illustrate the use of PPA to visualize the "MagicGamma" data using a small number of dimensions.
Figure~\ref{DR_results_ex_1_2D} shows how the model obtained by PPA (red line and grey grids) adapts to the samples (in blue).
All plots represent the same data from different points of view.
Note that the relation between data dimensions cannot be explained with linear correlation.

The curve (red) in the plots corresponds to the first identified polynomial or to the
data reconstructed using just one PPA dimension.
The grids in the first row of Fig.~\ref{DR_results_ex_1_2D} were computed by defining a uniform grid
in the first two dimensions of the transformed PPA domain, and transforming it back into the original domain.
Second row in Fig.~\ref{DR_results_ex_1_2D} represents visualizations in three dimensions, together with
grids computed inverting uniform samples in a $5 \times 5 \times 5$ cube (or 5 stacked surfaces) in the PPA domain.

The qualitative conclusion is that despite the differences in the cost function (see discussions in Sections~\ref{the_extension} and~\ref{related}),
the first PPA polynomial (red curve) also passes \emph{through the middle of the samples}, so it can be seen as an alternative to
the Principal Curve of the data \cite{Has84}. The gray grids also go \emph{through the middle of the {samples}}, which suggests that not
only alternative Principal Curves can be obtained with PPA, but also Principal Surfaces and Volumes \cite{Has84,Delicado01,OzertemTesis}.
Moreover, these surfaces and volumes help to visualize the structure of the data. This advantage can be seen clearly in the third and fourth plots of the first row,
where the data manifold {seems to be} embedded in more than two dimensions.

\begin{figure*}[p,t!]
\begin{center}
{
\begin{tabular}{p{0.1cm}ccc}
 &[MagicGamma] & [Japanese Vowels] & [Pageblock]\\
\multirow{4}{*}{\rotatebox{90}{Training error}}&
\hspace{-0cm} \includegraphics[width=5.2cm]{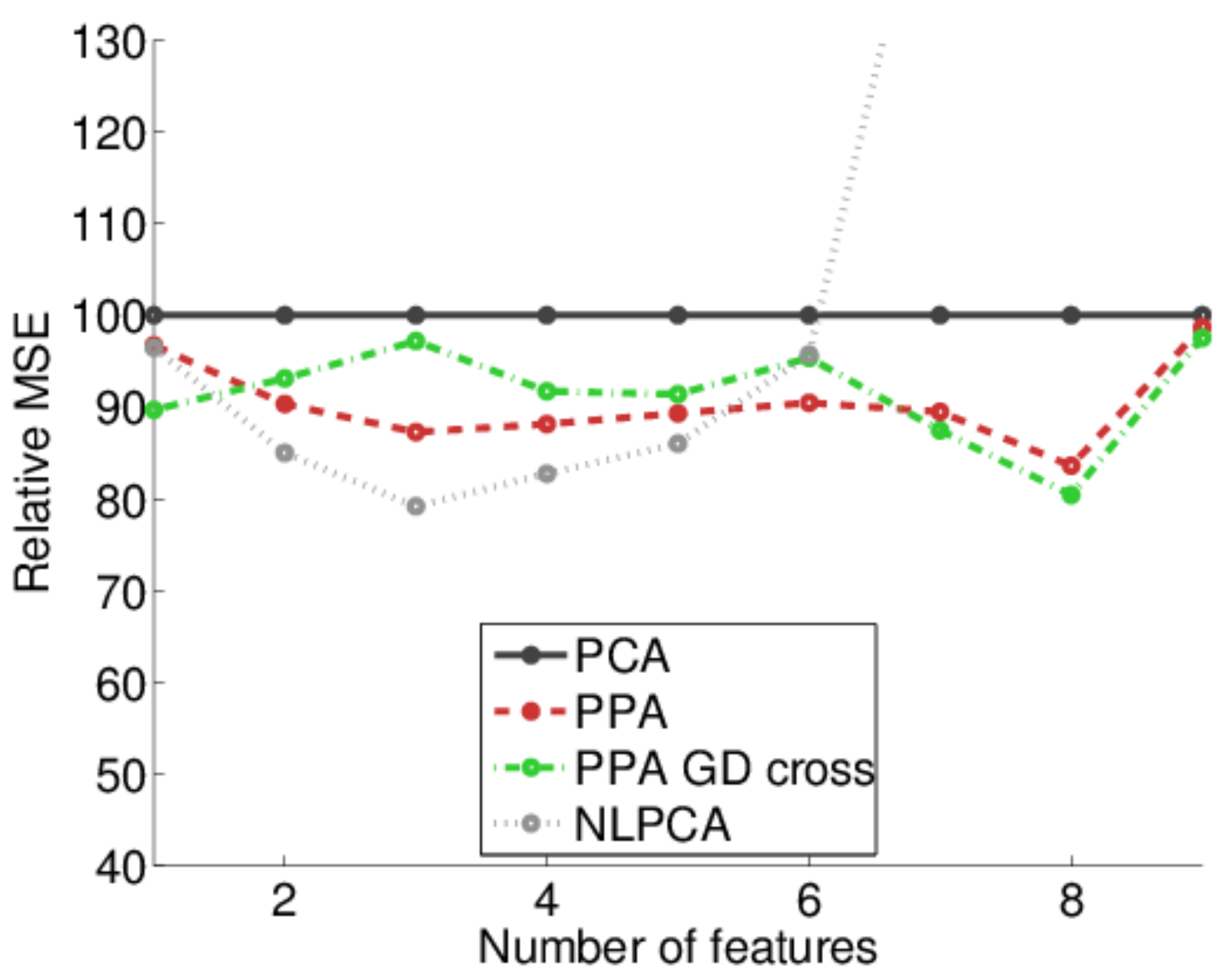}&
\hspace{-0cm} \includegraphics[width=5.2cm]{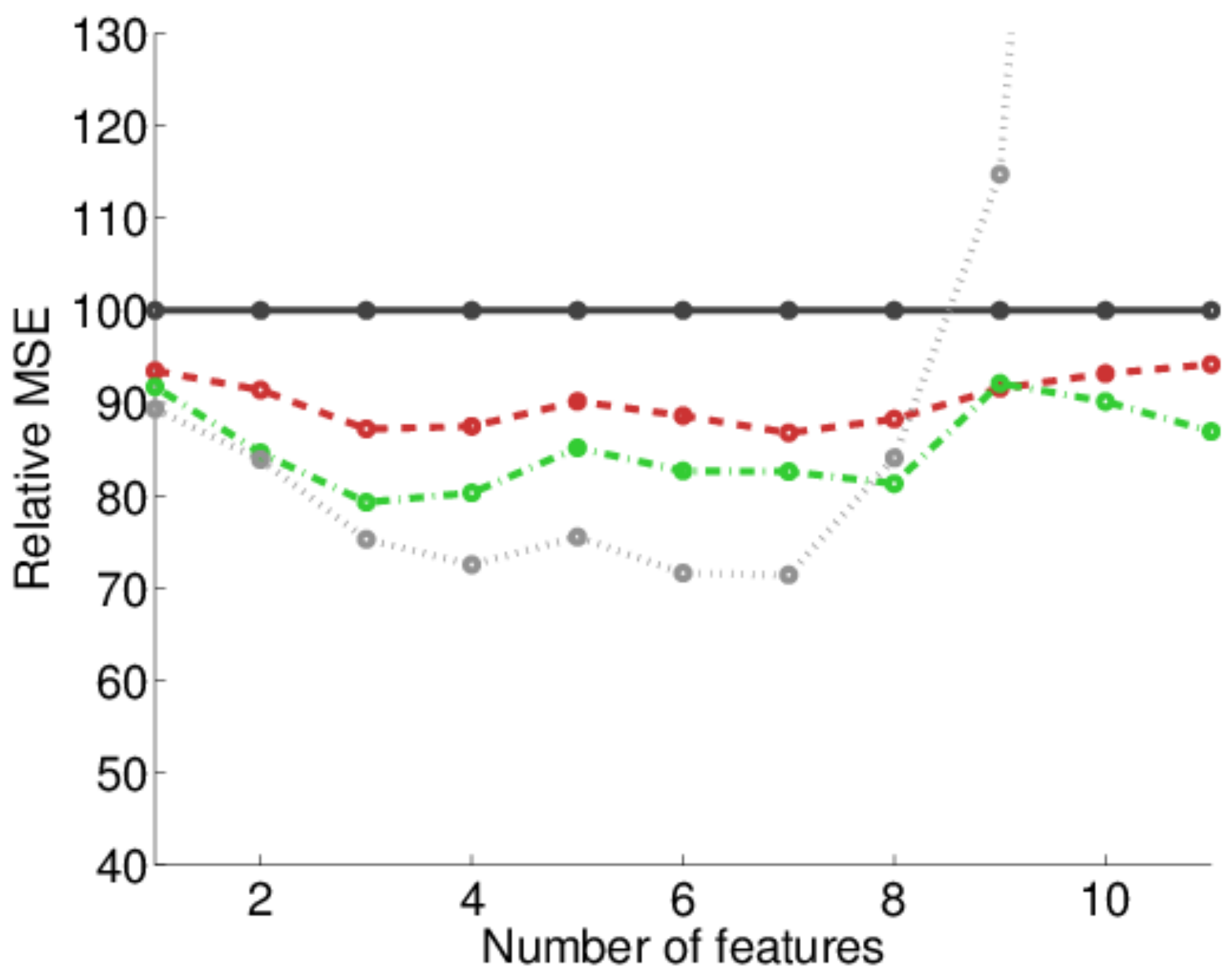}&
\hspace{-0cm} \includegraphics[width=5.2cm]{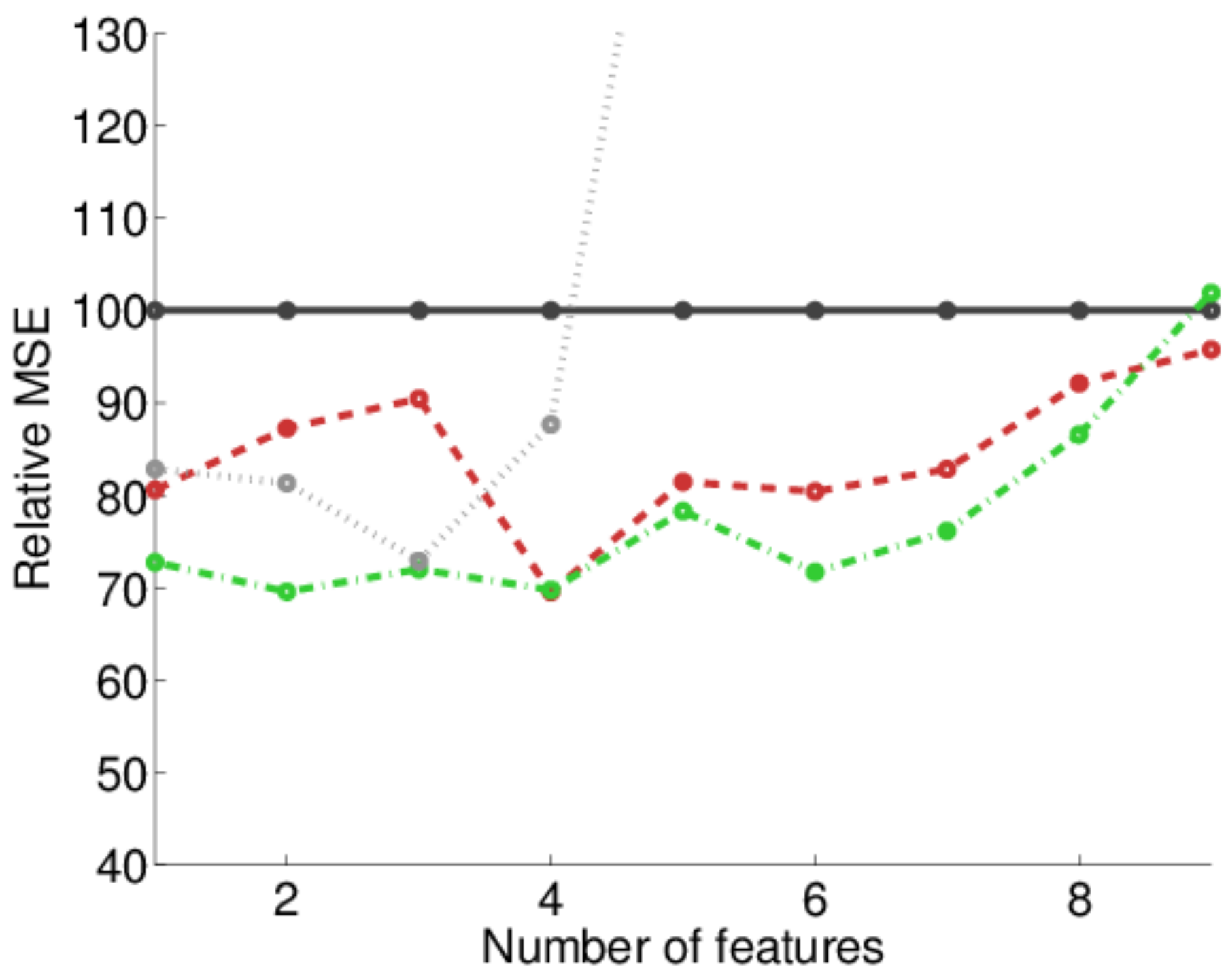}\\
&[Sat] & [Segmentation] & [Vehicles]\\
&\hspace{-0cm} \includegraphics[width=5.2cm]{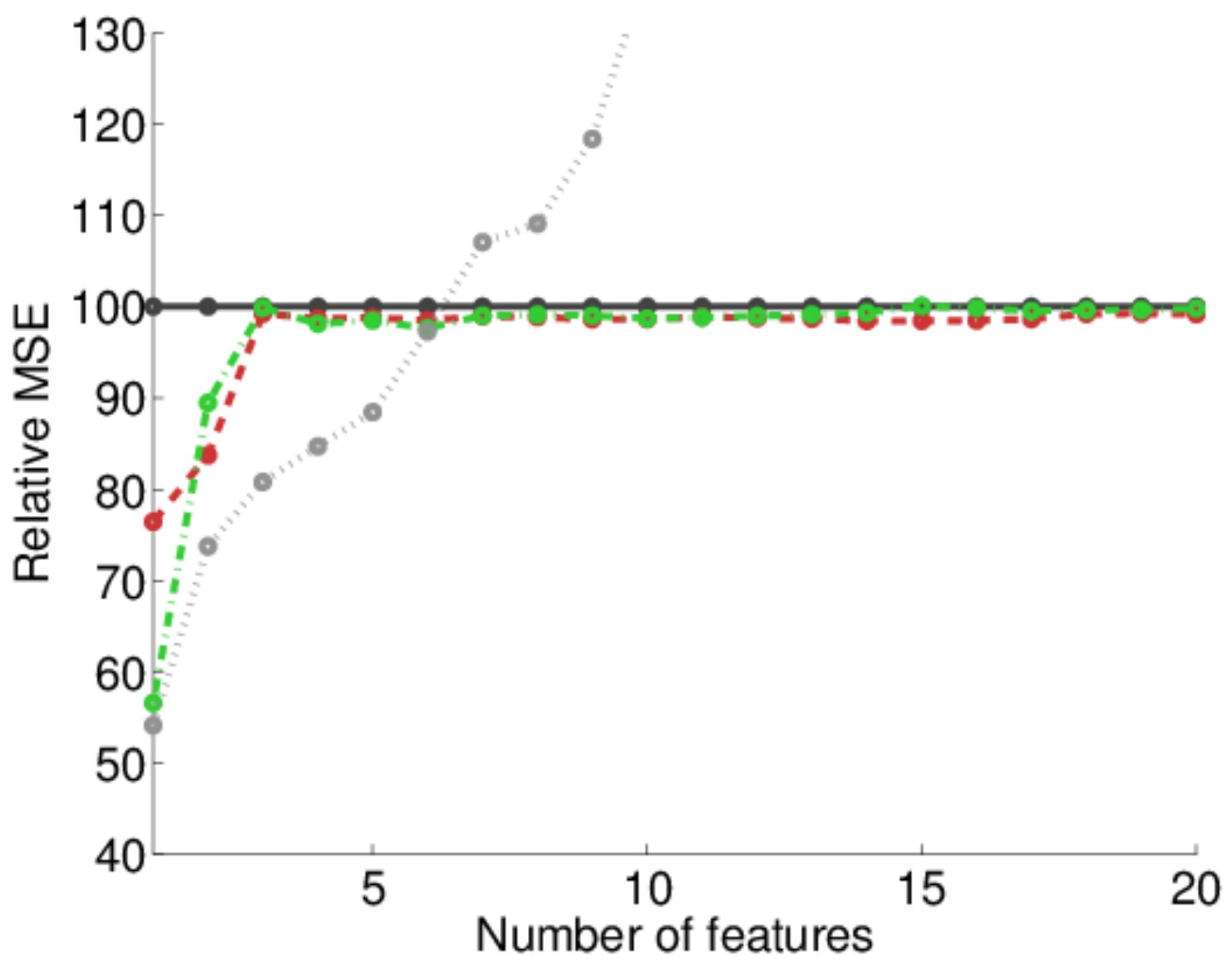}&
\hspace{-0cm} \includegraphics[width=5.2cm]{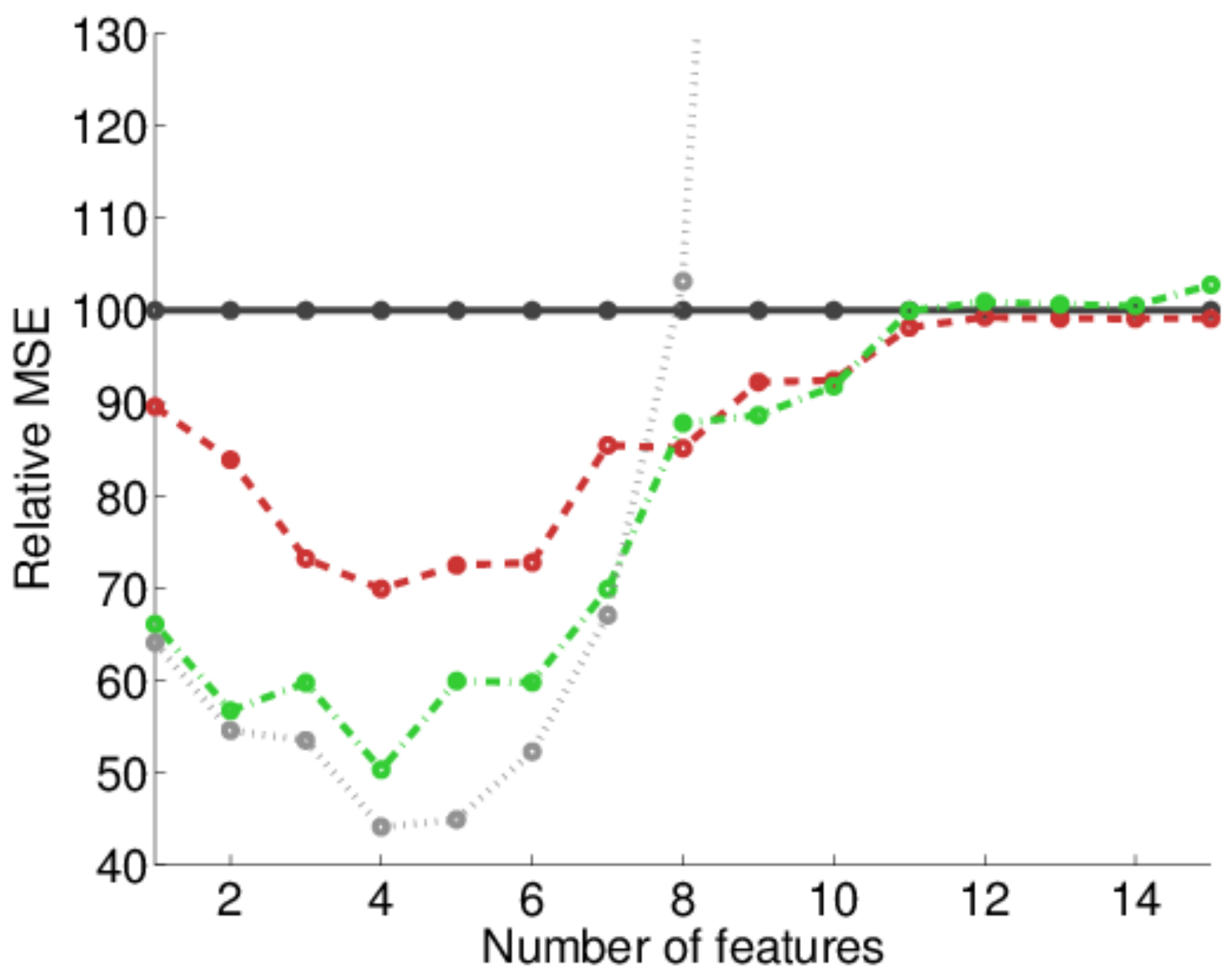}&
\hspace{-0cm} \includegraphics[width=5.2cm]{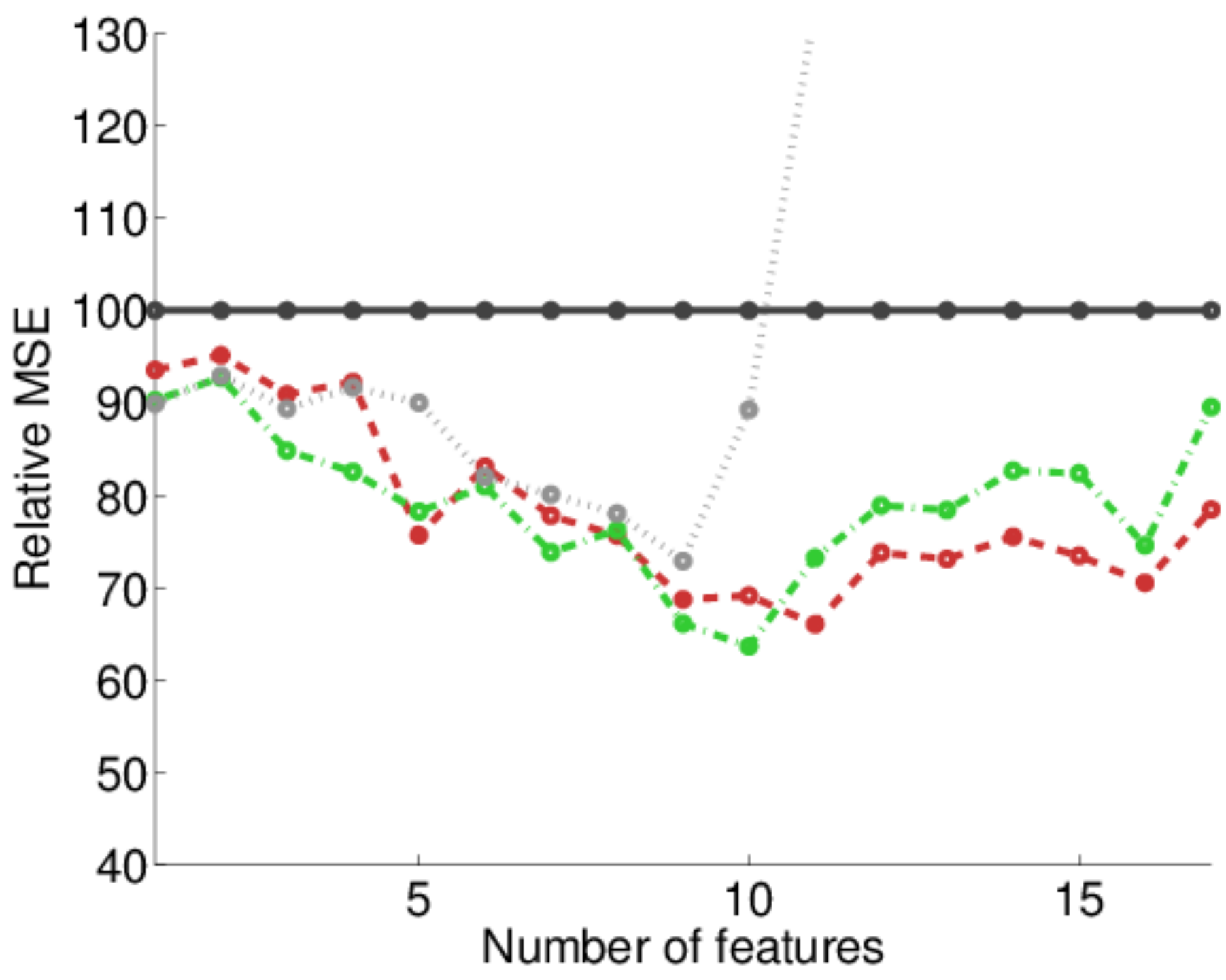}\\
  \hline \\
  &[MagicGamma] & [Japanese Vowels] & [Pageblock]\\
\multirow{4}{*}{\rotatebox{90}{Test error}}&
\hspace{-0cm} \includegraphics[width=5.2cm]{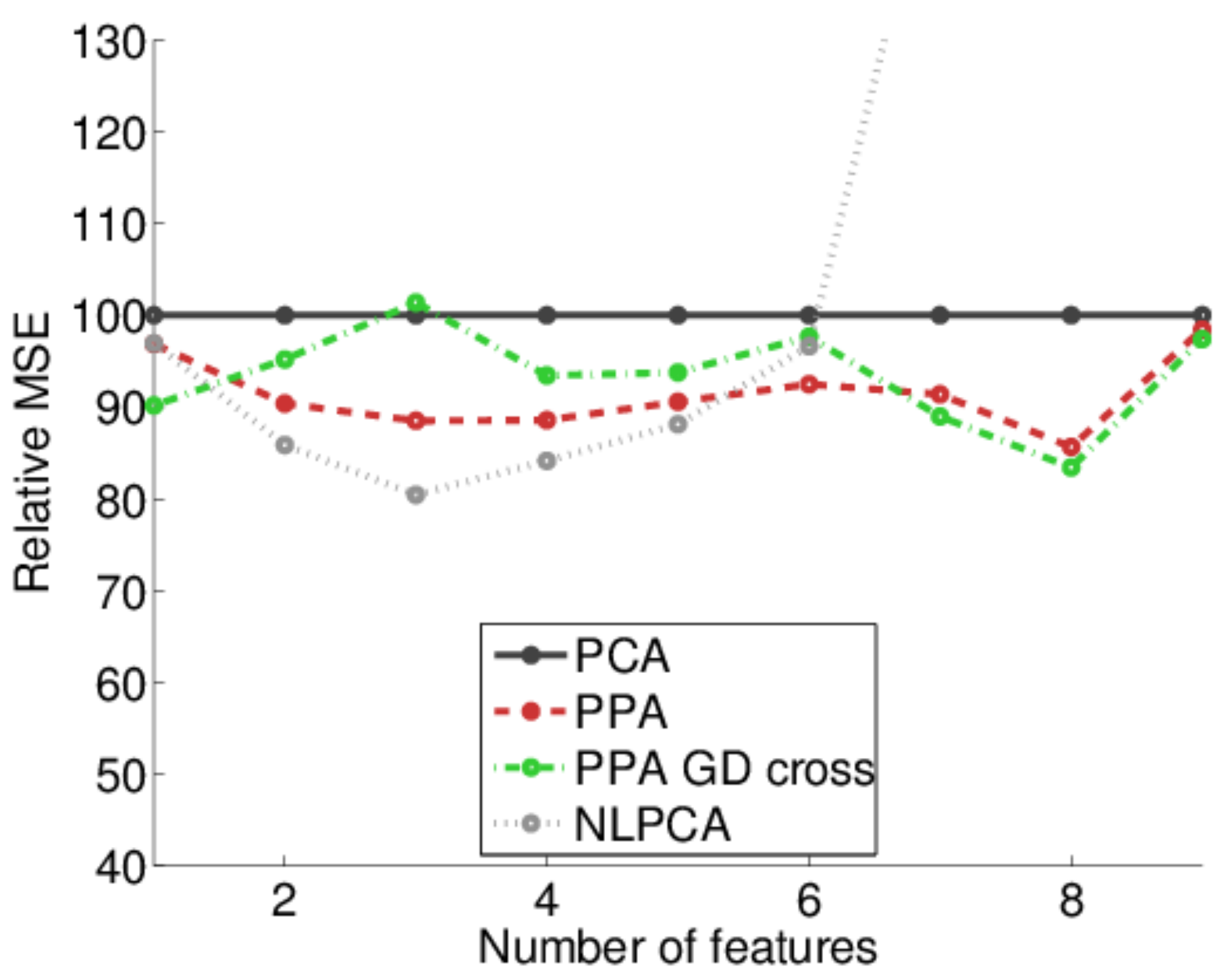}&
\hspace{-0cm} \includegraphics[width=5.2cm]{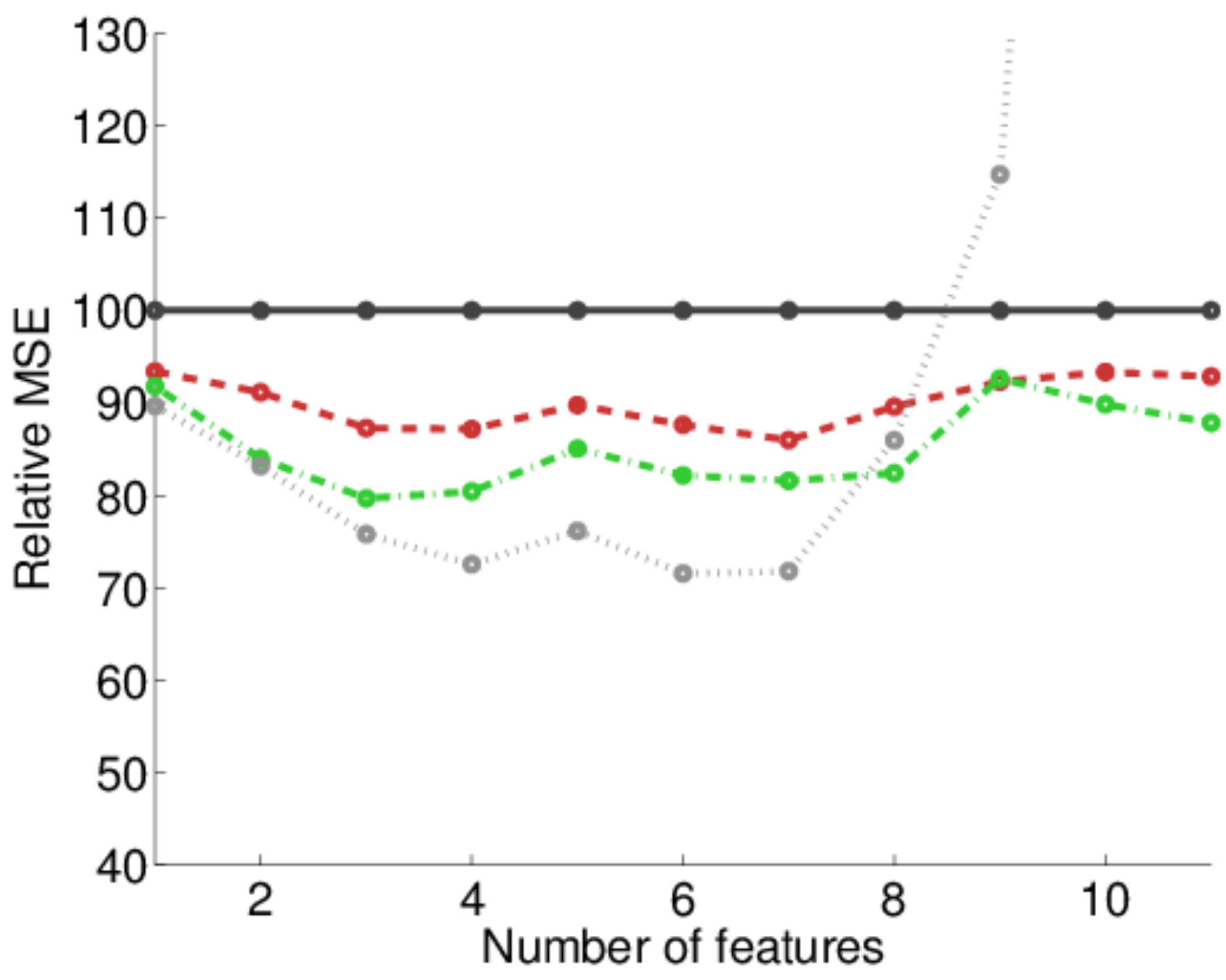}&
\hspace{-0cm} \includegraphics[width=5.2cm]{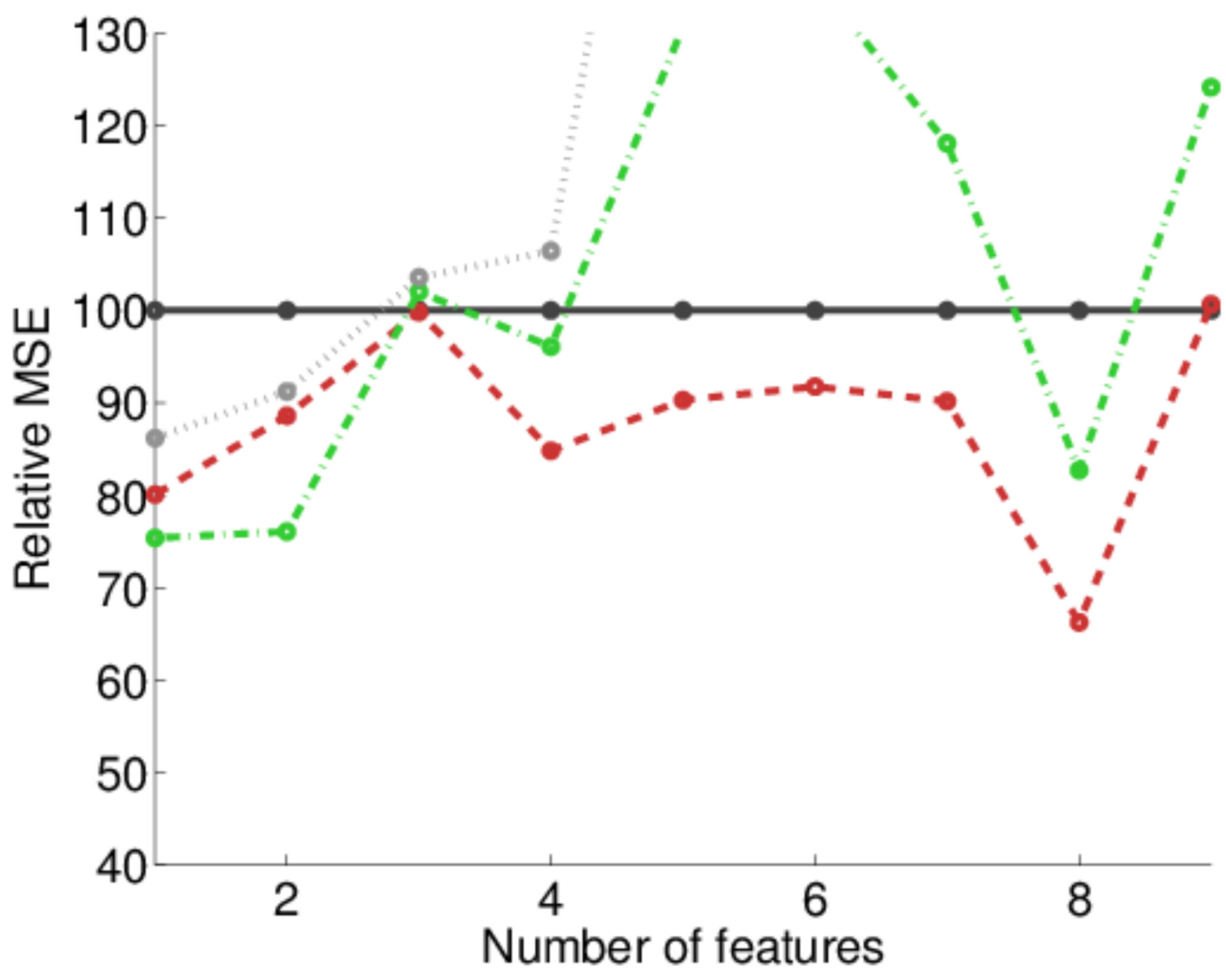}\\
&[Sat] & [Segmentation] & [Vehicles]\\
&\hspace{-0cm} \includegraphics[width=5.2cm]{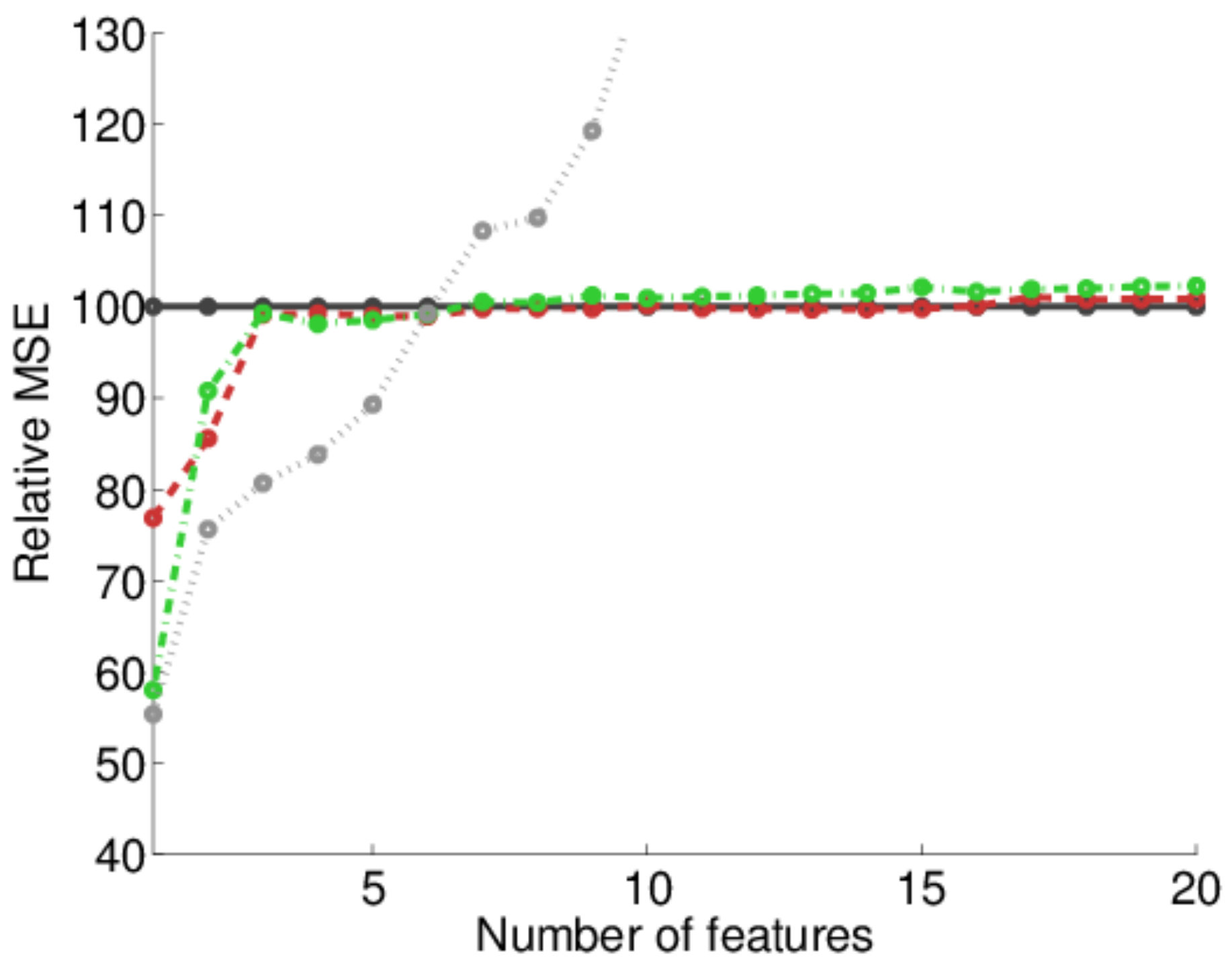}&
\hspace{-0cm} \includegraphics[width=5.2cm]{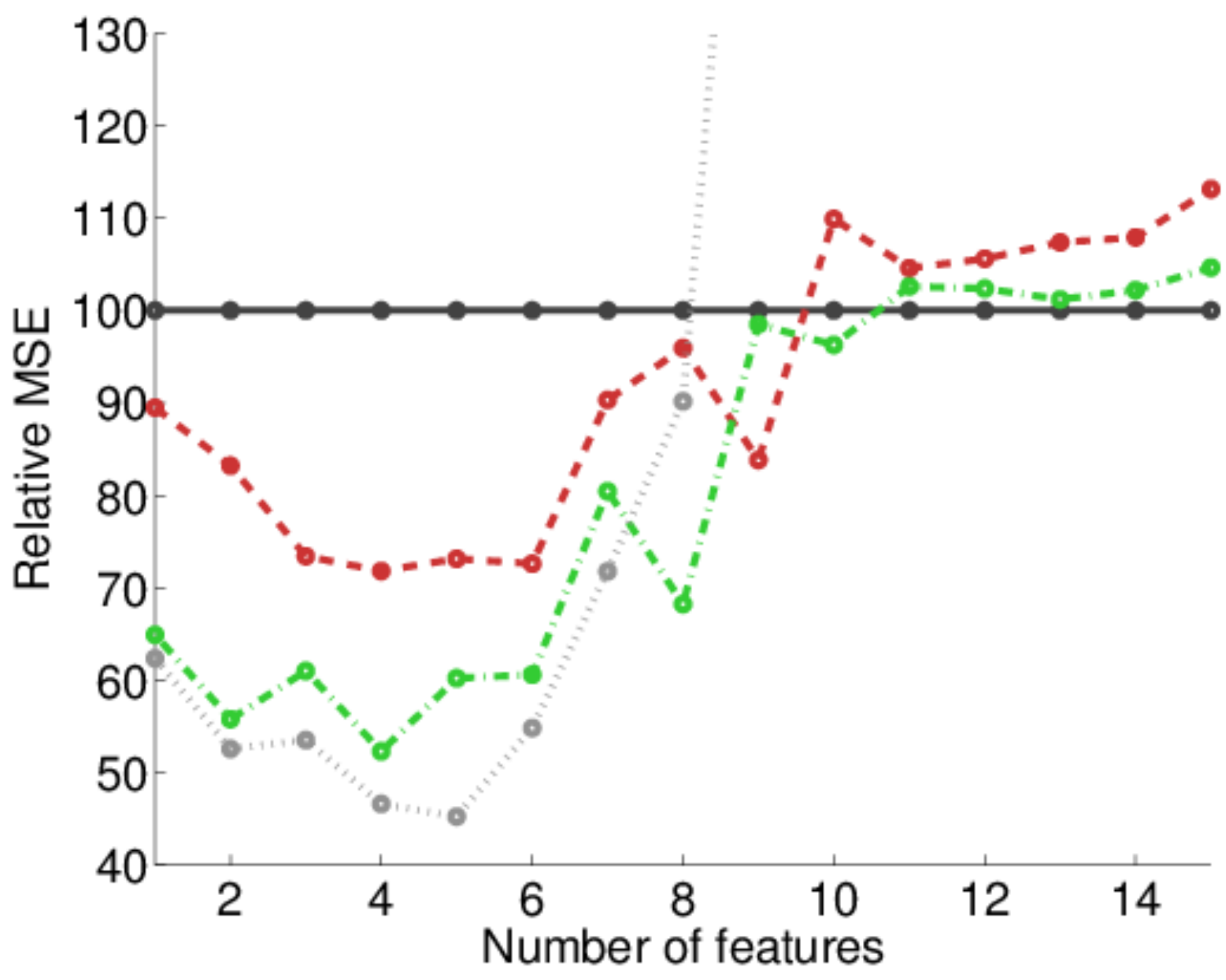}&
\hspace{-0cm} \includegraphics[width=5.2cm]{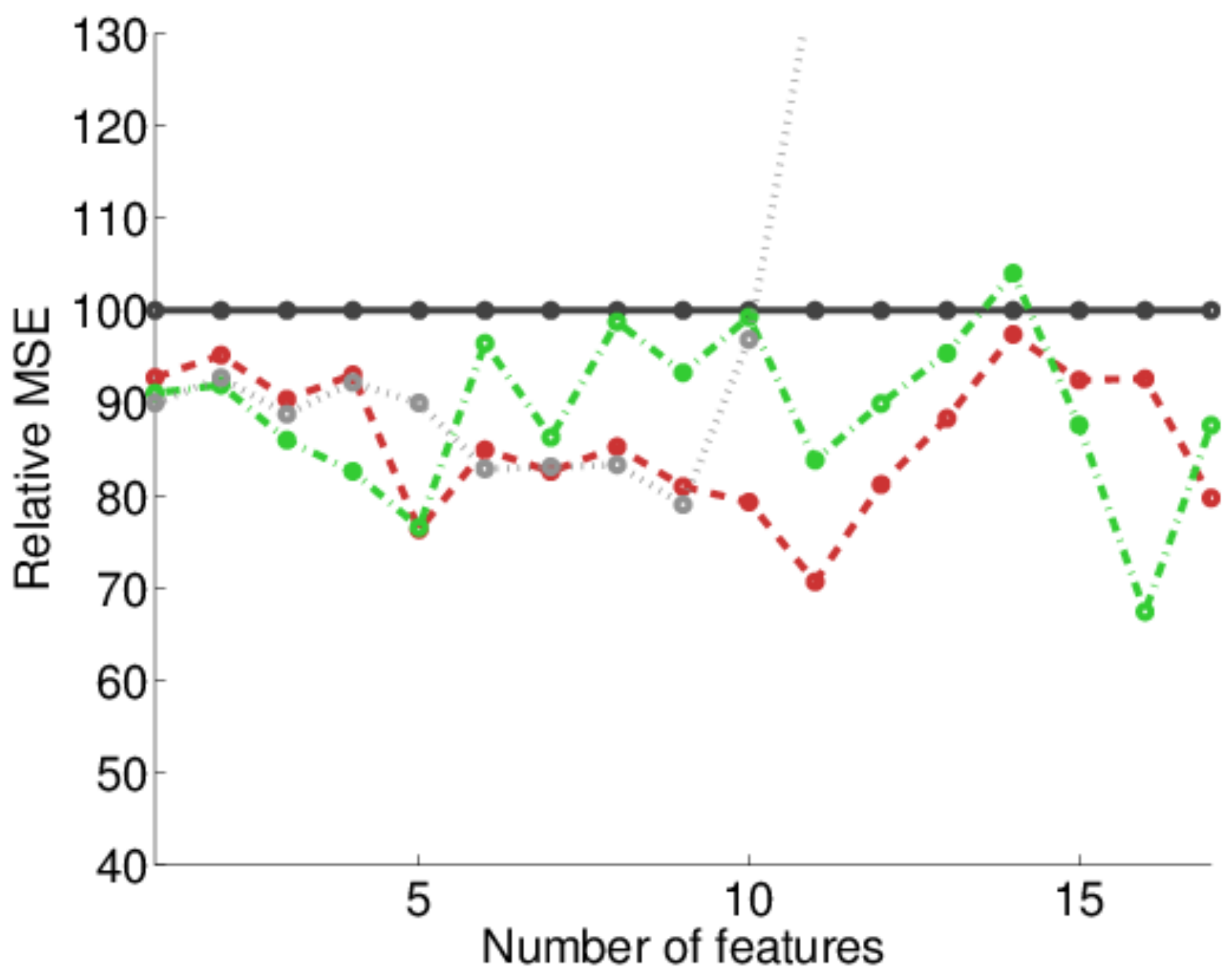}
\end{tabular}}
\vspace{0.3cm}
\caption{Relative reconstruction MSE (with regard to PCA) as a function of the retained dimensions for PCA, PPA, PPA GD and NLPCA.
Top panel: results on the training data. Bottom panel: results on the test data. }
\label{DR_results_1}
\end{center}
\end{figure*}

\paragraph{Reconstruction error}
To evaluate the performance in dimensionality reduction, we employ the reconstruction mean square error (MSE) in the original domain.
For each method, the data are transformed and then inverted retaining a reduced set of dimensions.
This kind of direct evaluation can be used only with invertible methods.
Distortion introduced by method $m$ is shown in terms of the relative MSE (in percentage) with regard to PCA: $\mathrm{Rel.MSE_m} = 100 \times \mathrm{MSE_m}/\mathrm{MSE_{PCA}}$
Results in this section are the average over ten independent realizations of the random selection of training samples.

Figure~\ref{DR_results_1} shows the results in relative MSE as a function
of the number of retained dimensions. Performance on the training and test sets is reported in the top and the
bottom panels respectively.
Note that 100 \% represents the base-line PCA error.

Several conclusions can be extracted from these results.
The most important conclusion is that \emph{PCA-based} PPA performs always better than PCA in the training set, as expected.
This may not be the case with new (unobserved) test data.
On the one hand, PPA is more robust in general than NLPCA for a high number of extracted features.
On the other hand, NLPCA only achieves good performance with a low number of extracted features.
It is worth noting that PPA GD obtains good results for the first component in the training sets,
in particular always better than PPA (as proved theoretically in sec. 2.3).
Generalization ability (i.e. performance in test) depends on the method and the database.
Even though a high samples per dimension ratio may help to obtain better generalization,
it is not always the case (see for instance results for database ``Sat'').
More complex methods (as PPA GD and NLPCA) perform better in training but not necessarily in test,
probably due to over-fitting. More {adapted} schemes for training could be employed (see for instance \cite{Scholz12}).

\paragraph{Computational Cost.}
Table~\ref{time_GD} illustrates the computational load for each method.
The main conclusion is that PCA is the less computationally demanding, and the NLPCA the most costly, as expected. The basic PPA takes around one order of magnitude more than PCA. Although this increases the demanding time to perform an experiment it is still useful for large databases. At this point, it is worth noting that the implementation of PPA has not been optimized, it is just the straight application of the algorithm presented in Section 2. More efficient implementations could be {implemented}, but this is out of the scope of this work. {Searching} the optimal direction by gradient descent makes PPA as costly as NLPCA.

\begin{table}[h!]
\begin{center}
\caption{Computational time (in min.) to learn the transform (per method and database).}
\begin{tabular}{|l|l|c|c|c|c|}
\hline
& & \multicolumn{4}{|c|}{Method} \\
\hline
 & Database & PCA & PPA & PPA GD & NLPCA \\ \hline \hline
1 & MagicGamma & 0.0010 &  0.0092 & 142.7 & 80.8 \\
2 & Japanese Vowels & 0.0006 &  0.0095 & 50.1 & 50.8 \\
3 & Pageblocks & 0.0002 &  0.0025 &  7.4 & 20.0 \\
4 & Sat & 0.0023 &  0.0390 & 68.2 & 122.4 \\
5 & Segmentation & 0.0002 &  0.0065 &  2.5 & 19.8 \\
6 & Vehicles& 0.0002 &  0.0019 &  0.3 &  9.8 \\
\hline
\end{tabular}
\label{time_GD}
\end{center}
\end{table}

\subsection{\em Multi-information reduction}
\label{MI}

Redundancy between the features of a representation is described by the multi-information, $I(\x)$.
Therefore certain transform is suitable for efficient coding if it reduces this redundancy.
Direct estimation of $I(\x)$ is difficult since it involves Kullback-Leibler divergences between multivariate densities.
However, multi-information reduction under a transform $\R$ is given by \cite{Lyu09}:
\begin{eqnarray}
\small
\begin{array}{ll}
	\Delta I &= I(\x) - I(\R(\x)) \\[2mm]
  &= \displaystyle\sum_{j=1}^d h(\x^{j}) - \displaystyle\sum_{j=1}^d h(\R(\x)^{j}) + {\mathbb E}[\log|\nabla \R(\x)|],
\end{array}
\label{eqMI}
\end{eqnarray}
where superscript $j$ in ${\bf z}^{j}$ indicates its $j$-th feature, and $h({\bf z}^{j})$ is the (easy to estimate) zero-order
entropy of the univariate data ${\bf z}^{j}$.

Therefore, multi-information reduction is particularly easy to estimate when $\R$ preserves the volume because in this
case $|\nabla \R| = 1$ so the only multivariate term in Eq.~\eqref{eqMI} vanishes. In that situation redundancy reduction
just depends on comparing marginal entropies before and after the mapping, which only involves univariate densities.

Table~\ref{MI_train} reports the multi-information reduction in bits per dimension for each database and each method. Note that NLPCA is not a volume-preserving map, and therefore its multi-information reduction can not be computed in practice. The main conclusion is that PPA obtains bigger reduction than PCA. This means that PPA obtains a representation where the dimensions of the data are more statistically independent. This is
an important property of PPA when used as a preprocessing method, because one can safely apply classifiers on the projected data that assume independence between dimensions, as for instance the naive Bayes classifier.

\begin{table}[h!]
\begin{center}
\caption{Multi-information reduction (in bits per dimension) achieved by each method (bigger is better).}
\begin{tabular}{|l|l|c|c|c|}
\hline
& & \multicolumn{3}{|c|}{Method} \\
\hline
& Database  & PCA & PPA & PPA GD \\ \hline \hline
1 & {MagicGamma} & 0.35 & 0.42 & 0.47 \\
2 & {Japanese Vowels} & 0.38 & 0.45 & 0.49 \\
3 & {Pageblock} & 0.16 & 0.23 & 0.25 \\
4 & {Sat} & 1.76 & 1.78 & 1.82 \\
5 & {Segmentation} & 1.20 & 1.23 & 1.34 \\
6 & {Vehicles} & 1.32 & 1.49 & 1.38 \\
\hline
\end{tabular}
\label{MI_train}
\end{center}
\end{table}



\section{Conclusions} \label{conclusions}

{Features extracted with linear PCA are optimal for dimensionality reduction {\em only} when data display a very particular symmetry.
The proposed PPA is a nonlinear generalization of PCA that relaxes such constraints. Essentially, PPA describes the data with
a sequence of curves aimed at minimizing the reconstruction error.}

{We analytically proved that PPA outperforms PCA in truncation error and in energy compaction. PPA also inherits all the
appealing properties that make linear PCA successful:
the PPA transform is computationally easy to obtain,
invertible (we presented a closed-form solution for the inverse),
geometrically interpretable (computable metric and curvatures),
allows out-of-sample projections without resorting to approximated methods,
returns a hierarchically layered representation, and
does not depend on the target dimension.
Additionally we showed that PPA is a volume-preserving transform, which is convenient to assess its redundancy reduction performance.}

{We also showed that the PPA functional is not convex. To address this problem we presented (1) a near-optimal closed-form solution
based on PCA that is guaranteed to outperform PCA, and (2) the tools for a gradient descent search of the optimal solution.
We analyzed the computational cost of both approaches. In the gradient descent solution the cost is
very high, similar to representations based on Principal Curves, non-linear PCA, or kernel PCA.
On the contrary, the cost of the PCA-based solution
is only moderately bigger than PCA and clearly inferior to the above methods.
Finally, results on real data showed the practical performance of
PPA on dimensionality and redundancy reduction compared to PCA and non-linear PCA.
In average, PPA roughly reduces a 15\% both the MSE reconstruction error and the redundancy of PCA. }


\begin{figure*}[t!]
\vspace{1cm}
\begin{center}
\small
\setlength{\tabcolsep}{1pt}
\vspace{-1.6cm}
\begin{tabular}{ccc}
\includegraphics[width=4.32cm]{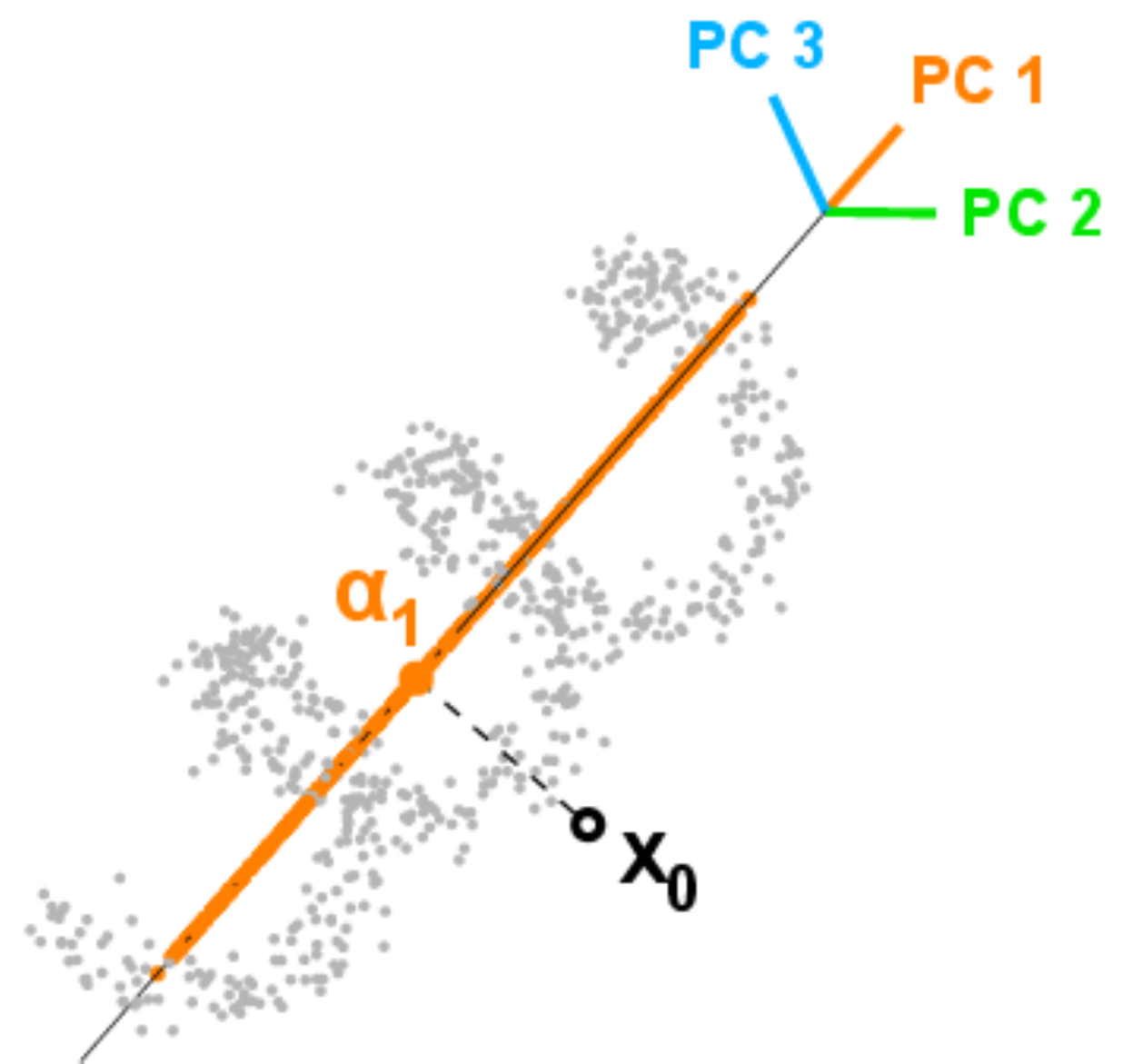} &
\includegraphics[width=6.12cm]{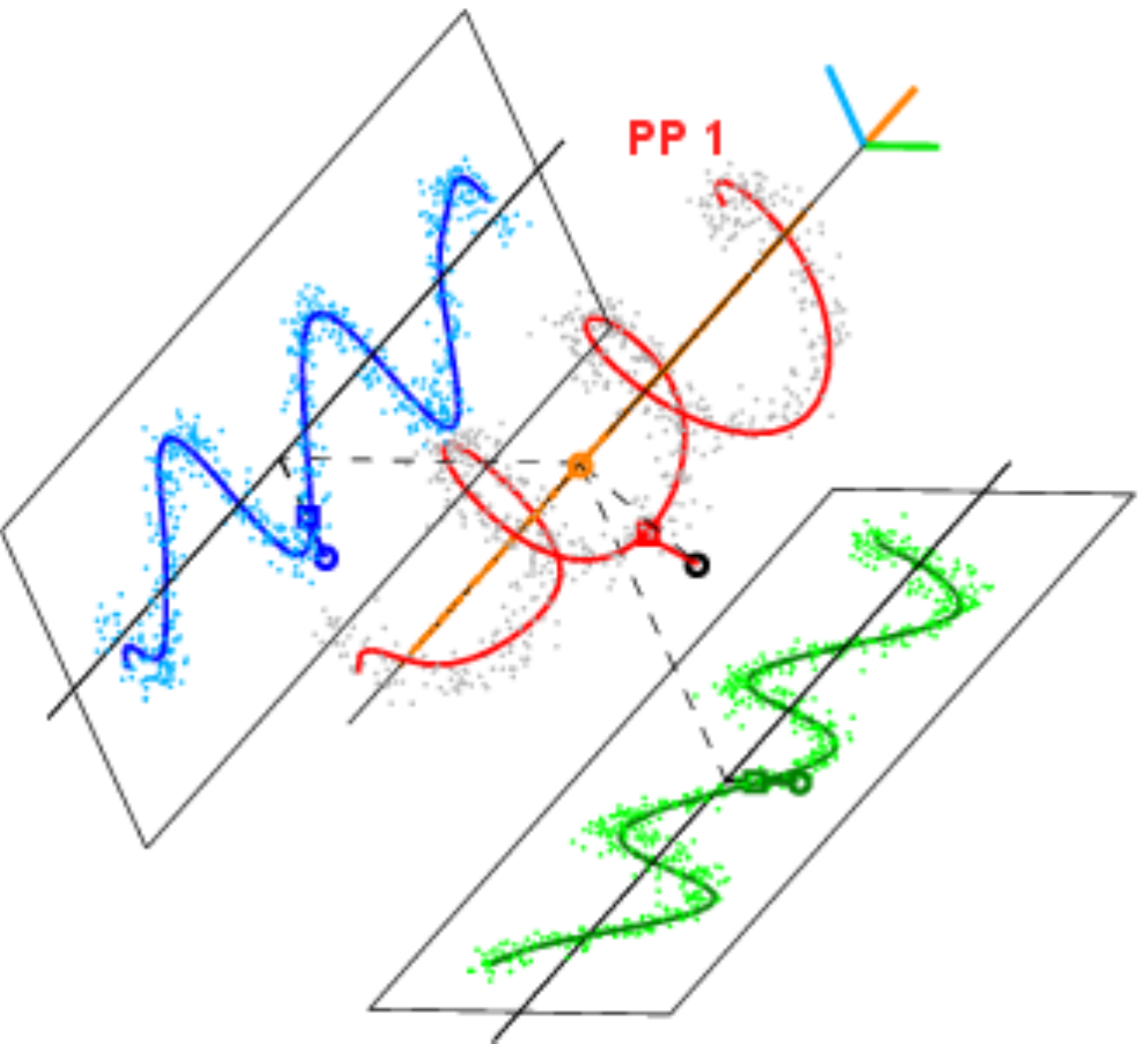} &
\includegraphics[width=4.95cm]{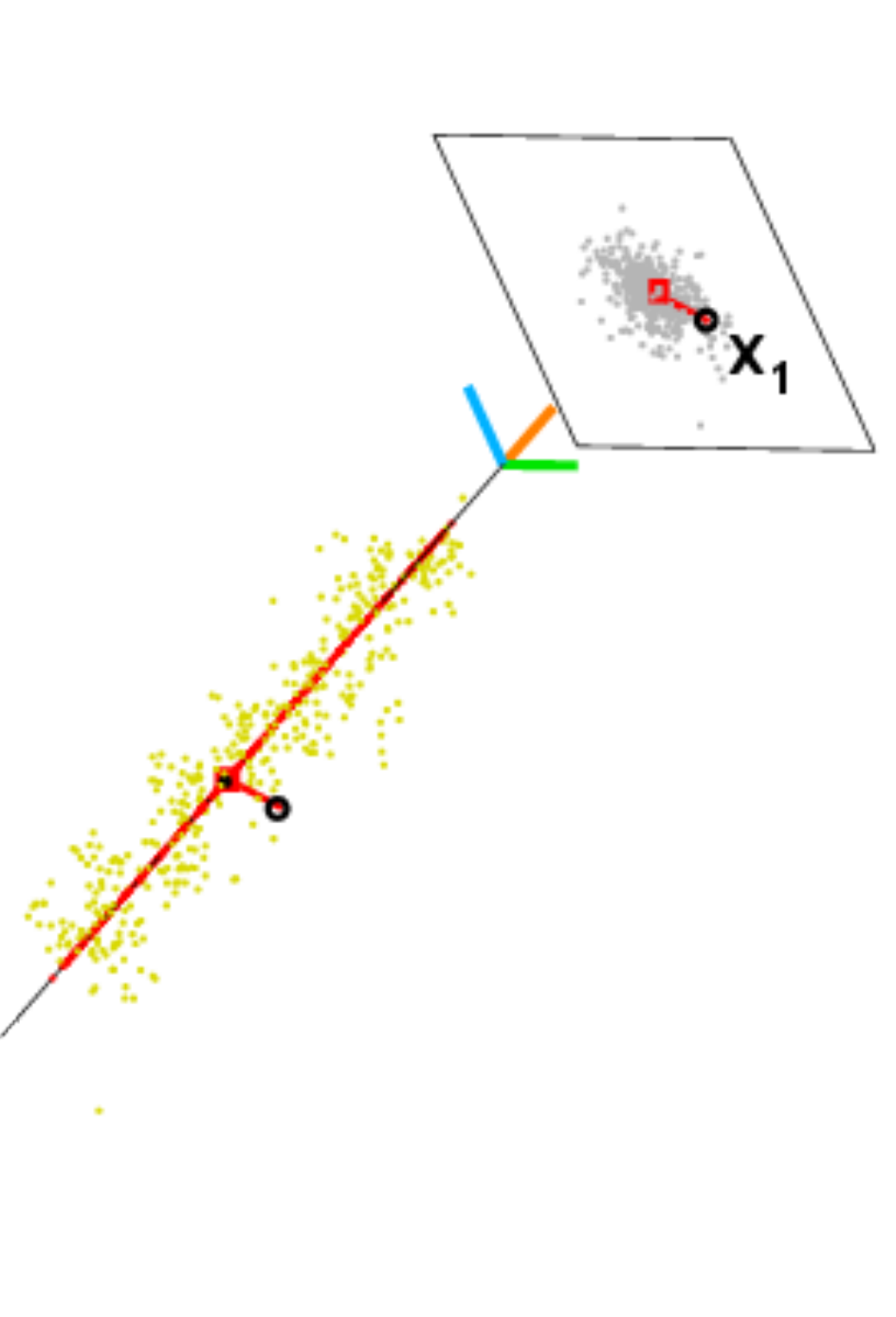} \\
\includegraphics[width=2.52cm]{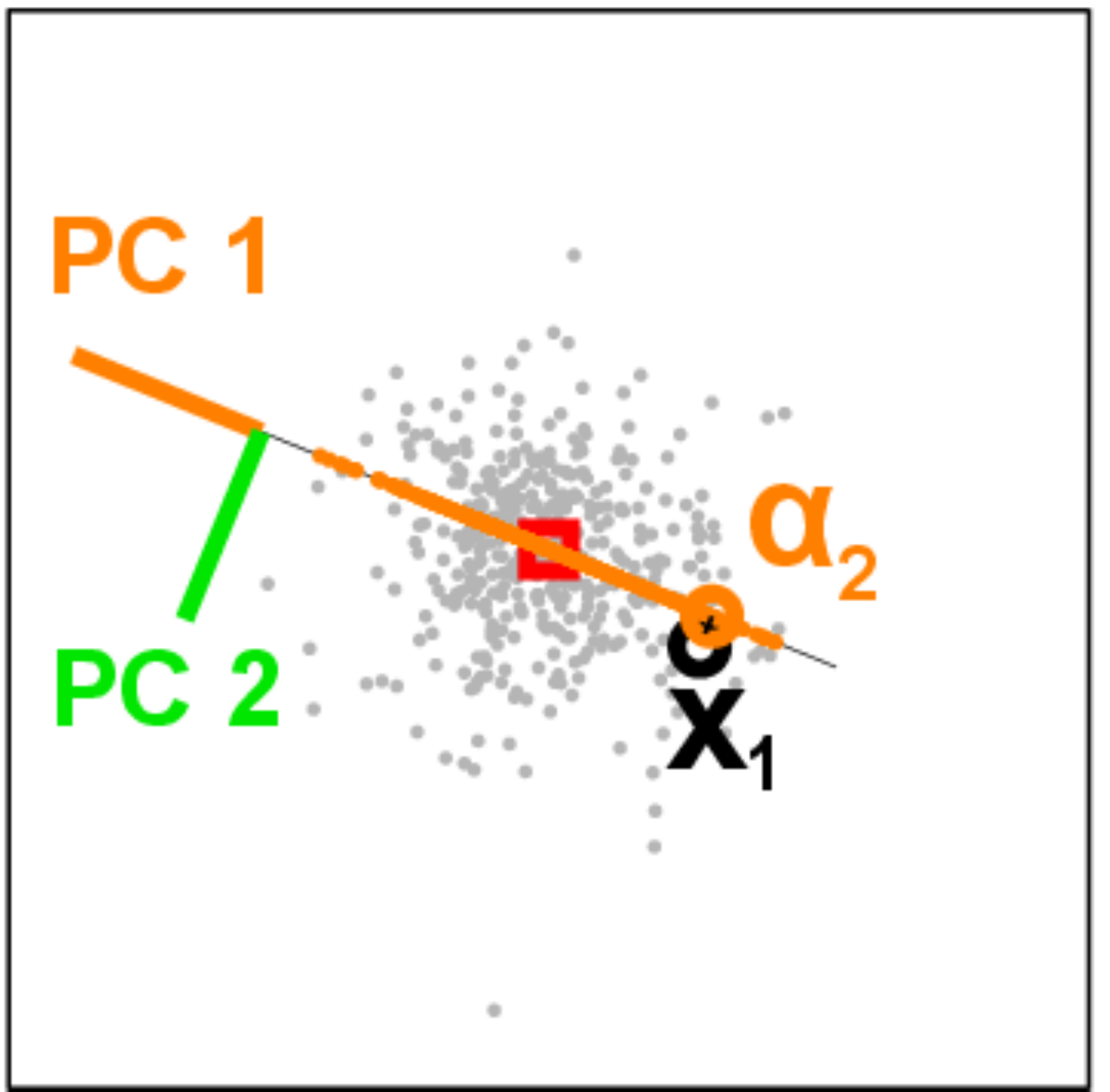} &
\includegraphics[width=2.52cm]{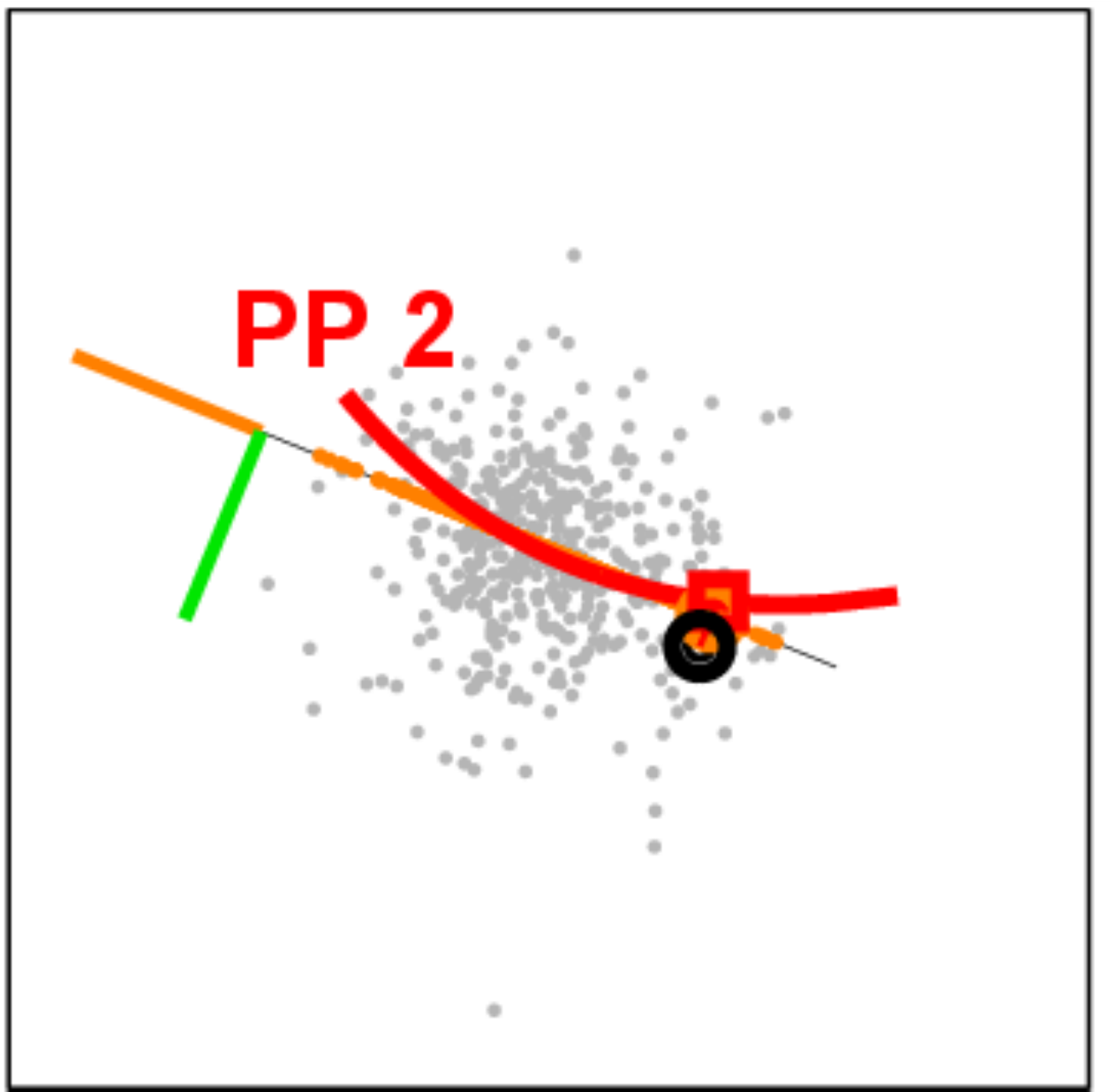} &
\hspace{-0.4cm}\includegraphics[width=2.90cm]{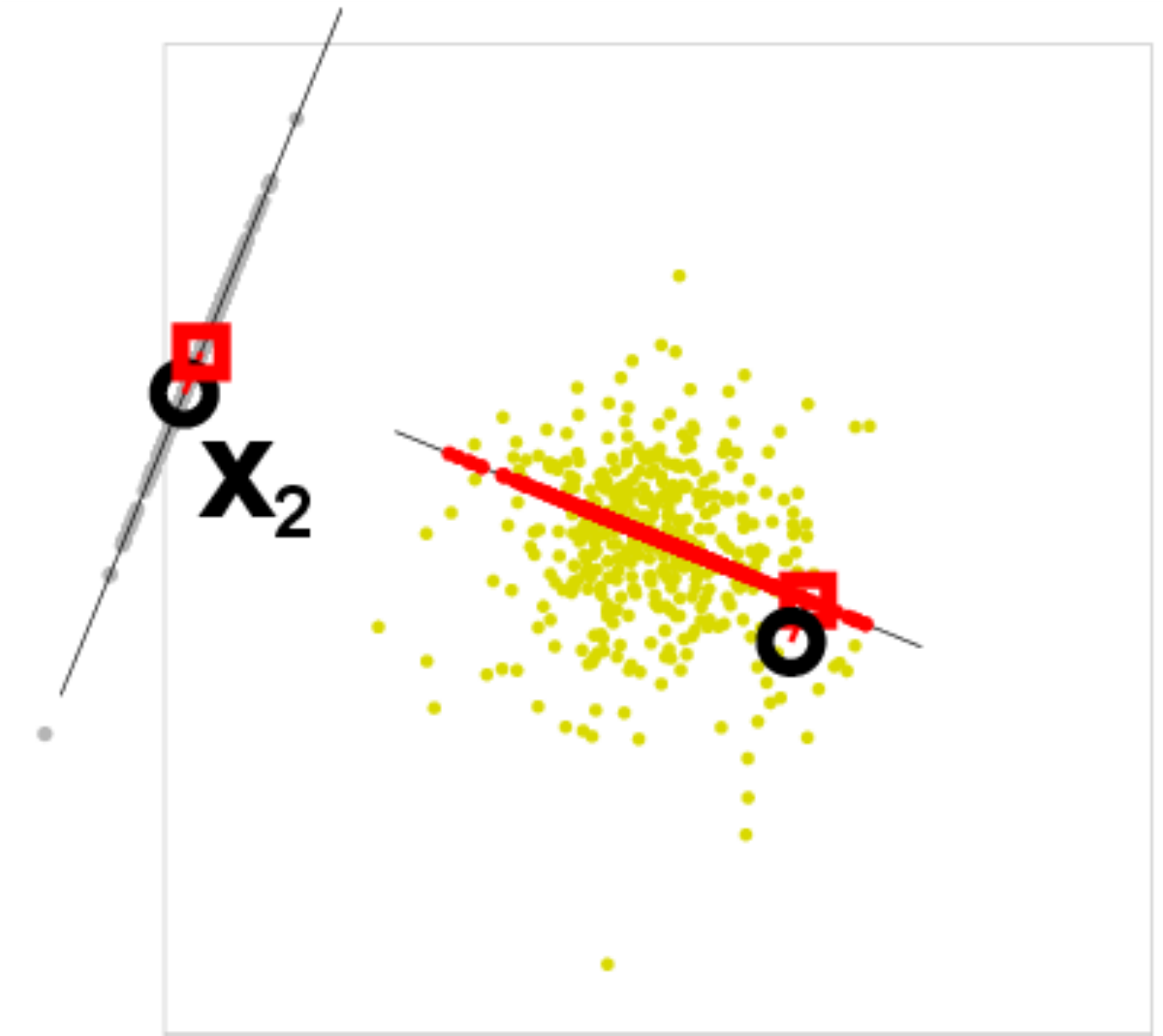} \\
\end{tabular}
\end{center}
\vspace{-0.2cm}
\caption{\small {Forward PPA transform illustrated in a 3d example. Top row summarizes the steps in the first transform of the sequence $\mathbf{R}_1$, which accounts for one curvilinear dimension and leads to a $2d$ residual: projection (left), polynomial fit (center), and conditional mean subtraction (right). See text for details on the symbols. Bottom row shows the equivalent steps in $\mathbf{R}_2$, that leads to the final $1d$ residual.
}
}
\label{con_cucharita}
\end{figure*}

\section{Acknowledgments}

This paper has been partially supported by the Spanish MINECO
under project TIN2012-38102-C03-01 and by the Swiss NSF
under project PZ00P2-136827.

\appendix
\section{Appendix: Forward PPA illustrated} \label{app}

{Figure~\ref{con_cucharita} presents a step-by-step example to illustrate how the sequence
of PPA curvilinear components and projections are computed on a manifold of well-defined
geometry: an helix embedded in a 3d space corrupted with additive Gaussian noise which is
a usual test case in Principal Curves~\cite{Ozertem11}.
Data (in gray) were sampled from the same helix as in section~\ref{helice} and noise with standard deviation 0.3.
Since $d=3$, PPA consists of a sequence of two transforms (see Eq.~\eqref{seq_squeme}):
$\mathbf{R}_1$ (first row in Fig.~\ref{con_cucharita}) and $\mathbf{R}_2$ (second row).
A representative sample is highlighted throughout the transform}.

{In this example we use the PCA-based solution. Therefore, the leading vector $\mathbf{e}_1$
is the first eigenvector (biggest eigenvalue) of the covariance matrix of $\mathbf{x}_0$.
In the example, $\mathbf{e}_1$ (or PC1, in orange), and the vectors PC2 and PC3 (in green and blue respectively)
constitute the basis $\mathbf{E}_1$.
The first PPA component, $\alpha_1$, is the projection of the data onto the first leading vector, $\alpha_1 = \mathbf{e}_1^\top \mathbf{x}_0$, in Eq.~\eqref{PPAapprox} (orange dots and the circle for the highlighted sample).
The conditional mean, $\mathbf{m}_1$, is shown decomposed in two subspaces in the top center panel.
We will call $\mathbf{m}_{1a}$ the conditional mean in the subspace spanned by $\mathbf{e}_1$ and PC2 (green dots),
and let $\mathbf{m}_{1b}$ be the conditional mean in the subspace spanned by $\mathbf{e}_1$ and PC3 (blue dots).
It is obvious the strong non-linear dependence of the conditional mean with $\alpha_1$, i.e. given the value
of $\alpha_1$ ($\mathbf{e}_1$ axis -black line-) it is easy to predict the value of the data in the
orthogonal subspaces (blue and green dots) using a non-linear function.}

 {Fitting the first PPA polynomial in $3$ dimensions with regard to the parameter $\alpha_1$
 is equivalent to fitting the polynomials in the $2d$ subspaces in the center plot (simple univariate regressions).
 The polynomials in the $2d$ subspaces have the coefficients $\mathbf{W}_{1a} = [w_{1a1} \text{ } w_{1a2} \text{ } w_{1a3} \text{ } \ldots w_{1a(\gamma_1+1)}]$, and equivalently, $\mathbf{W}_{1b}$; which are the rows of the matrix $\W_1$.
 Polynomial coefficients are easy to fit by constructing the Vandermonde matrix of degree $\gamma_1$ using $\alpha_1$, $\bold{v}_1 = [1  \text{ } \alpha_1  \text{ } \alpha_1^2 \text{ }  ... \text{ }  \alpha_1^{\gamma_1+1}]^\top$ and applying Eq.~\eqref{matrix_W_ls}.
 This ensures the best fitting in least squares terms.
 Then, we estimate $\mathbf{m}_{1a}$ (and correspondingly $\mathbf{m}_{1b}$) using $\alpha_1$ and the weights, Eq.~\eqref{mean_estimation}:
 }
 \begin{equation}
   \mathbf{\hat{m}}_{1a} = w_{1a1} + w_{1a2}\alpha_1 + w_{1a3}\alpha_1^2 \ldots w_{1a\gamma_1+1}\alpha_1^{\gamma_1}
   \label{Cond_m_Poly}
\end{equation}
{In the top center panel, the estimated conditional mean, $\mathbf{\hat{m}}_1 = [\mathbf{\hat{m}}_{1a} \text{ } \mathbf{\hat{m}}_{1b}]^\top$, is represented by the curve (red), while the curve projected in the bottom plane (green) and the curve projected in the vertical plane (blue) represent the conditional means in the respective subspaces ($\mathbf{\hat{m}}_{1a}$ and $\mathbf{\hat{m}}_{1b}$). Once the polynomial has been fitted, we can remove $\mathbf{\hat{m}}_1$ from each sample (second line in Eq.~\eqref{PPAapprox}) obtaining the residuals (departures from the conditional mean) represented in the top right plot (yellow dots).
}

{Summarizing the process in the top row, the transform $\mathbf{R}_1$, the first Principal Polynomial (red curve) accounts for the first curvilinear dimension of the data. After $\mathbf{R}_1$, we have $(d-1) = 2$ dimensions yet to be explained: $\mathbf{x}_1$,
at the top right and bottom left plots.
 The second row of Fig.~\ref{con_cucharita} reproduces the same steps in the reduced dimension residual:
 projection onto the first PC in the bottom left plot (orange dots), fitting the polynomial (in this case, the best cross-validation solution was a second order polynomial, represented by the curve (red) in the bottom center plot, and removing the conditional mean
 so that the residuals (yellow dots) are aligned, and projected in the orthogonal subspace.}


{\small
\bibliographystyle{unsrt}}

\end{document}